\documentclass[twoside,11pt]{article}

\usepackage[abbrvbib]{jmlr2e}

\usepackage{amsmath,amsfonts}
\usepackage{mathtools}
\usepackage{enumerate,color,xcolor}
\mathtoolsset{showonlyrefs}
\usepackage{subfigure}
\usepackage{url}
\usepackage[textsize=footnotesize]{todonotes}
\usepackage{todonotes}
\usepackage{bbm}

\usepackage{comment}

\newcommand{\beq}{\begin{eqnarray}} 
\newcommand{\eeq}{\end{eqnarray}}
\newcommand{\beqs}{\begin{eqnarray*}} 
\newcommand{\eeqs}{\end{eqnarray*}}
\newcommand{\R}{\mathbb{R}}

\newcommand{\bigO}{\mathcal{O}}

\makeatletter
\def\munderbar#1{\underline{\sbox\tw@{$#1$}\dp\tw@\z@\box\tw@}}
\makeatother

\usepackage{booktabs}

\newtheorem{theo}{Theorem}[section]  

\newtheorem{prop}[theo]{Proposition}
\newtheorem{lemm}[theo]{Lemma}
\newtheorem{rema}[theo]{Remark}
\newtheorem{coro}[theo]{Corollary}

\newtheorem{assum}[theo]{Assumption}

\usepackage{tabularx}

\usepackage{multirow}

\usepackage{xcolor}
\usepackage{comment}
\usepackage[ruled]{algorithm2e}
\usepackage[normalem]{ulem} 
\newcommand\str{\bgroup\markoverwith
{\textcolor{red}{\rule[0.5ex]{2pt}{1.5pt}}}\ULon} 
\usepackage{pifont}
\ShortHeadings{Penalized Langevin Monte Carlo for Constrained Sampling}{G\"{u}rb\"{u}zbalaban, Hu and Zhu}
\firstpageno{1}

\begin{document}

\title{Penalized Overdamped and Underdamped Langevin\\ 
Monte Carlo Algorithms for Constrained Sampling}

\author{\name Mert G\"{u}rb\"{u}zbalaban \email mg1366@rutgers.edu \\
       \addr Department of Management Science and Information Systems \\
       Rutgers Business School\\
       Piscataway, NJ 08854, United States of America
       \AND
       \name Yuanhan Hu \email yh586@scarletmail.rutgers.edu \\
       \addr 
       Department of Management Science and Information Systems \\
       Rutgers Business School\\
       Piscataway, NJ 08854, United States of America
       \AND
       \name Lingjiong Zhu \email zhu@math.fsu.edu \\
       \addr Department of Mathematics \\
       Florida State University \\
       Tallahassee, FL 32306, United States of America\\
       }

\editor{Maxim Raginsky}

\maketitle

\begin{abstract}
We consider the constrained sampling problem where the goal is to sample from a target distribution $\pi(x)\propto e^{-f(x)}$ when $x$ is constrained to lie on a convex body $\mathcal{C}\subset \mathbb{R}^d$. Motivated by penalty methods from continuous optimization, we propose and study penalized Langevin Dynamics (PLD) and penalized underdamped Langevin 
Monte Carlo (PULMC) methods for constrained sampling that convert the constrained sampling problem into an unconstrained sampling problem by introducing a penalty function for constraint violations. When $f$ is smooth and gradients of $f$ are available, we show $\tilde{\mathcal{O}}(d/\varepsilon^{10})$ iteration complexity for PLD to sample the target up to an $\varepsilon$-error where the error is measured in terms of the total variation distance and $\tilde{\mathcal{O}}(\cdot)$ hides some logarithmic factors. For PULMC, we improve this result to $\tilde{\mathcal{O}}(\sqrt{d}/\varepsilon^{7})$ when the Hessian of $f$ is Lipschitz and the boundary of $\mathcal{C}$ is sufficiently smooth. To our knowledge, these are the first convergence rate results for underdamped Langevin Monte Carlo methods in the constrained sampling setting that can handle non-convex choices of $f$ and can provide guarantees with the best dimension dependency among existing methods for constrained sampling when the gradients are deterministically available. We then consider the setting where only unbiased stochastic estimates of the gradients of $f$ are available, motivated by applications to large-scale Bayesian learning problems. We propose PSGLD and PSGULMC methods that are variants of PLD and PULMC that can handle stochastic gradients and that are scaleable to large datasets without requiring Metropolis-Hasting correction steps. 
For PSGLD and PSGULMC, when $f$ is strongly convex and smooth, we obtain an iteration complexity of $\tilde{\bigO}(d/\varepsilon^{18})$ and $\tilde{\bigO}(d\sqrt{d}/\varepsilon^{39})$ respectively in the 2-Wasserstein distance. For the more general case, when $f$ is smooth and $f$ can be non-convex, we also provide finite-time performance bounds and iteration complexity results. Finally, we illustrate the performance of our algorithms on Bayesian LASSO regression and Bayesian constrained deep learning problems.
\end{abstract}


\section{Introduction}

We consider the problem of sampling a distribution $\pi$
on a convex constrained domain $\mathcal{C}\subsetneq\mathbb{R}^{d}$
with probability density function 
\begin{equation} 
\pi(x)\propto\exp(-f(x)),\ x\in\mathcal{C},
    \label{eq-target}
\end{equation}
for a function $f:\mathbb{R}^d \to \mathbb{R}$. This is a fundamental problem arising in many applications, including Bayesian statistical inference \citep{gelman1995bayesian}, Bayesian formulations of inverse problems \citep{stuart2010inverse}, as well as Bayesian classification and regression tasks in machine learning \citep{andrieu2003introduction,teh2016consistency,gurbuzbalaban2021decentralized}.

In the absence of constraints, i.e., when $\mathcal{C}=\mathbb{R}^{d}$ in \eqref{eq-target}, many algorithms in the literature are applicable \citep{geyer1992practical,brooks2011handbook} including the class of Langevin Monte Carlo algorithms. 
One popular algorithm for this setting is the \emph{unadjusted Langevin algorithm}:
\begin{equation}\label{discrete:overdamped}
x_{k+1}=x_{k}-\eta\nabla f(x_{k})+\sqrt{2\eta}\xi_{k+1},
\end{equation}
where $\xi_{k}$ are independent and identically distributed (i.i.d.) $\mathcal{N}(0,I_{d})$ Gaussian vectors in $\mathbb{R}^{d}$. The classical Langevin algorithm \eqref{discrete:overdamped} is the Euler discretization of the
\emph{overdamped (or first-order) Langevin diffusion}:
\begin{equation}\label{eq:overdamped-2}
dX(t)=-\nabla f(X(t))dt+\sqrt{2}dW_{t},
\end{equation}
where 
$W_{t}$ is a standard $d$-dimensional Brownian motion that starts at zero
at time zero. Under some mild assumptions on $f$, the stochastic differential equation (SDE) \eqref{eq:overdamped-2} admits a unique stationary distribution with the density $\pi(x) \propto e^{-f(x)}$,
known as the \emph{Gibbs distribution} \citep{chiang1987diffusion,stroock-langevin-spectrum}. In computing practice, this diffusion is simulated by considering its discretization as in \eqref{discrete:overdamped} whose stationary distribution may contain a bias that a Metropolis-Hasting step can correct. However, for many applications, including those in data science and machine learning, employing this 
correction step can be computationally expensive \citep{bardenet2017markov,teh2016consistency}; therefore, our focus will be on unadjusted algorithms that avoid it. 

Unadjusted Langevin algorithms have a long history and admit various asymptotic convergence guarantees \citep{talay1990expansion,mattingly2002ergodicity,gelfand1991simulated}; however non-asymptotic performance bounds for them are relatively more recent \citep{Dalalyan,DM2017,DM2016,DMP2016,CB2018}. 
The unadjusted Langevin algorithm \eqref{discrete:overdamped} assumes availability of the gradient $\nabla f$. On the other hand, in many settings in machine learning, computing the full gradient $\nabla f$ is either infeasible or impractical. For example, in Bayesian regression or classification problems, $f$ can have a finite-sum form as the sum of many component functions, i.e., $f(x) = \sum_{i=1}^n f_i(x)$ where $f_i(x)$ represents the loss of a predictive model with parameters $x$ for the $i$-th data point and the number of data points $n$ can be large (see, e.g., \citet{gurbuzbalaban2021decentralized,pan-gu-langevin}). In such settings, algorithms that rely on \emph{stochastic gradients}, i.e., unbiased stochastic estimates of the gradient obtained by a randomized sampling of the data points, is often more efficient \citep{bottou2010large}. This fact motivated the development of Langevin algorithms that can support stochastic gradients. 
In particular, if one replaces the full gradient $\nabla f$ in \eqref{discrete:overdamped} by a stochastic gradient, 
the resulting algorithm
is known as the stochastic gradient Langevin dynamics (SGLD) (see, e.g., \citet{welling2011bayesian,chen2015convergence}). 

Unadjusted underdamped Langevin Monte Carlo (ULMC) algorithms based on an alternative diffusion called underdamped (or second-order) Langevin diffusion have also been proposed; see e.g. \citet{dalalyan2018kinetic,Ma2019}. Their versions that support stochastic gradients are also studied (see e.g. \citet{chen2014stochastic,zou2021convergence,GGZ}). 
Although ULMC algorithms can often be faster than unadjusted (overdamped) Langevin algorithms on many practical problems \citep{chen2014stochastic}, this is rigorously proven for particular choices of $f$ \citep{chen2015convergence,GGZ,mangoubi-journal,chen2022optimal} rather than general non-convex choices of $f$ and the convergence of ULMC algorithms remains relatively less studied.

In this paper, we focus on the constrained setting when $\mathcal{C}$ is a convex body, i.e., when $\mathcal{C}$ is a compact convex set with a non-empty interior, and we consider both settings when $f$ can be strongly convex or non-convex. We also consider
both deterministic and stochastic gradients. Among the existing approaches that are the most closely related to our setting, \citet{bubeck2015finite, bubeck2018sampling} studied the projected Langevin Monte Carlo algorithm that projects the iterates back to the constraint set after applying the Langevin step \eqref{discrete:overdamped}
where it is assumed that $f$ is $\beta$-smooth, i.e.
$\Vert\nabla f(x)-\nabla f(y)\Vert\leq\beta\Vert x-y\Vert$ 
for any $x,y\in\mathcal{C}$ and the norm of the gradient of $f$
is bounded, i.e. $\Vert\nabla f(x)\Vert\leq L$. It is shown in \citet{bubeck2018sampling} that $\tilde\bigO(d^{12}/\varepsilon^{12})$ iterations are sufficient for having $\varepsilon$-error in the total variation (TV) metric with respect to the target distribution when the gradients are exact where the notation $\tilde\bigO(\cdot)$ hides some logarithmic factors. 
\citet{pmlr-v134-lamperski21a} considers the projected stochastic gradient Langevin dynamics (P-SGLD) in the setting of non-convex smooth Lipschitz $f$ on a convex body where the gradient noise is assumed to have finite variance with a uniform sub-Gaussian structure. The author shows that $\tilde{\bigO}\left(d^4/\varepsilon^{4}\right)$ iterations suffice in the 1-Wasserstein metric. More recently, \citet{zheng2022constrained} study P-SGLD for constrained sampling for a non-convex potential $f$ that is strongly convex outside a radius of $R$ where data variables are assumed to be $L$-mixing. They obtain an improved complexity of $\tilde{\bigO}\left(d^2/\varepsilon^2\right)$ for P-SGLD in the 1-Wasserstein metric with polyhedral constraints that are not necessarily bounded.
Constrained sampling for convex $f$ and strongly-convex $f$
is also studied in \citet{Brosse}, where a proximal Langevin Monte Carlo
is proposed and a complexity of $\tilde{\bigO}\left(d^5/\varepsilon^6\right)$ is obtained. \citet{SR2020} further studies the proximal stochastic gradient Langevin algorithm
from a primal-dual perspective. For constrained sampling when $f$ is strongly convex, and the constraint set is convex, the proximal step corresponds to a projection step, and they obtain $\tilde{\mathcal{O}}(d/\varepsilon^{2})$
complexity for the proximal stochastic gradient Langevin algorithm in terms
of the 2-Wasserstein distance.

Mirror descent-based Langevin algorithms (see e.g. \citet{hsieh2018mirrored,Chewi2020,Zhang2020,TaoMirror2021,Ahn2021})
can also be used for constrained sampling. Mirrored Langevin dynamics
was proposed in \citet{hsieh2018mirrored}, inspired by the classical mirror descent in optimization. 
For any target distribution with strongly-log-concave density (which corresponds to $f$ being strongly convex), \citet{hsieh2018mirrored} showed that their first-order algorithm requires $\tilde{\bigO}(\epsilon^{-2}d)$ iterations for $\varepsilon$ error with exact gradients and $\tilde{\bigO}(\epsilon^{-2}d^2)$ iterations for stochastic gradients. \citet{Zhang2020} establishes for the first time a non-asymptotic upper bound on the sampling error of the resulting Hessian Riemannian Langevin Monte Carlo algorithm that is closely related to the mirror-descent scheme. This bound is measured according to a Wasserstein distance induced by a Riemannian metric capturing the Hessian structure.  
In contrast to \citet{hsieh2018mirrored},
\citet{Zhang2020} studies a different scheme in which an appropriate diffusion term is used
that entails
a Gaussian noise in the discrete scheme with iteration-dependent covariances that account for
the Hessian Riemannian structure instead of a standard Gaussian noise adopted in \citet{hsieh2018mirrored}. 
Moreover \cite{Zhang2020} relaxes the strong-convexity assumptions to relative versions.
Motivated by \citet{Zhang2020}, \citet{Chewi2020}
propose a class of diffusions called Newton-Langevin diffusions and prove that they
converge exponentially fast with a rate that has no dependence on the condition number of the target density in continuous time. They give an application
of this result to the problem of sampling from the uniform distribution on a convex
body using a strategy inspired by interior-point methods.
In \citet{jiang2021mirror}, the author relaxes the strongly-log-concave density assumption in mirror-descent Langevin dynamics and assumes that the density function satisfies the mirror log-Sobolev inequality.
Further improvements \citet{Zhang2020} have been achieved in \citet{Ahn2021,TaoMirror2021}.
The analysis of \citet{Zhang2020} gives an error bound that contains a bias that does not vanish even if the stepsize goes to zero. The solution to this problem was first attempted by \cite{Ahn2021} who proposed an alternative discretization which achieves a vanishing bias, but requires an exact simulation of the Brownian motion with changing covariance.
Finally, \citet{TaoMirror2021} proved this bias is an artifact of analysis by building upon the mean-square analysis in \citet{Li2019,Li2022}.

\subsection{Our Approach and Contributions}
Recent years have witnessed techniques and concepts from continuous optimization being used for analyzing and developing new Langevin algorithms \citep{Dalalyan2017,balasubramanian2022towards,chen2022improved,gurbuzbalaban2021decentralized}. In this paper, we develop Langevin algorithms for constrained sampling, leveraging penalty functions from continuous optimization. More specifically, penalty methods are frequently used in continuous optimization \citep{nocedal1999numerical}, where one converts the constrained optimization problem of minimizing an objective $f(x)$ subject to $x \in \mathcal{C}$ to an unconstrained optimization problem of minimizing $f_\delta(x):= f(x) + \frac{1}{\delta}S(x)$ on $\mathbb{R}^d$, where $\delta>0$ is called the \emph{penalty parameter}, and the function $S:\mathbb{R}^d\to [0,\infty)$ is called the \emph{penalty function} with the property that $S(x)=0$ for $x \in \mathcal{C}$ and $S(x)$ increases as $x$ gets away from the constraint set $\mathcal{C}$. For $\delta>0$ small enough, it can be seen that the global minimum of $f_\delta$ will approximate the global minimum of $f$ on $\mathcal{C}$. Motivated by this technique, our main approach is to sample from a 
\emph{penalized target distribution in an unconstrained fashion} with the modified target density:
\begin{equation}
\pi_{\delta}(x)\propto\exp\left(-\left( f(x)+\frac{1}{\delta}S(x)\right)\right),\qquad x\in\mathbb{R}^{d},
\end{equation}
for suitably chosen small enough $\delta>0$. Here, a key challenge is to control the error between $\pi_\delta$ and $\pi$ efficiently, leveraging the convex geometry of the constraint set and the properties of the penalty function. We then use the unconstrained SGLD or stochastic gradient underdamped Langevin 
Monte Carlo (SGULMC) algorithm to sample from the
modified target distribution and call the resulting algorithms penalized SGLD (PSGLD) and penalized SGULMC (PSGULMC). If the gradients are deterministic, then we call the algorithms penalized Langevin dynamics (PLD) and penalized underdamped Langevin 
Monte Carlo (PULMC). 
Our detailed contributions are as follows:
\begin{itemize}
\item When $f$ is smooth, meaning that its gradient is Lipschitz, we show $\tilde{\mathcal{O}}(d/\varepsilon^{10})$ iteration complexity in the TV distance for PLD. For PULMC, we improve this result to $\tilde{\mathcal{O}}(\sqrt{d}/\varepsilon^{7})$ when the Hessian of $f$ is Lipschitz and the boundary of $\mathcal{C}$ is sufficiently smooth. To our knowledge, these are the first convergence rate results for underdamped MC methods in the constrained sampling setting that can handle non-convex choices of $f$ and provide guarantees with the best dimension dependency among existing methods for constrained sampling when subject to deterministic gradients. 
To achieve these results, we develop a novel analysis and make a series of technical contributions. 
We first bound the Kullback-Leibler (KL) divergence between $\pi_{\delta}$ and $\pi$ with a careful technical analysis and then  
apply weighted Csisz\'{a}r-Kullback-Pinsker inequality to control the 2-Wasserstein distance between $\pi_{\delta}$ and $\pi$. To obtain the convergence rate to $\pi_{\delta}$, we first regularize the convex domain $\mathcal{C}$ so that the regularized domain $\mathcal{C}^{\alpha}$ is $\alpha$-strongly convex (a notation which will be defined rigorously in \eqref{C:alpha} and in the proof of Lemma \ref{lemm:strongly convex}) and then show that $f+S^{\alpha}/\delta$ is strongly convex outside a compact domain, where $S^{\alpha}$ is the penalty function we construct for the regularized domain that has quadratic growth properties. Moreover, we quantify the differences between $\mathcal{C}^{\alpha}$ and $\mathcal{C}$, and between the regularized target $\pi_{\delta}^{\alpha}$ (defined on the regularized domain $\mathcal{C}^{\alpha}$) and $\pi_{\delta}$ and show their differences are small for the choice of small values of $\alpha$. Finally, we show that $f+S^{\alpha}/\delta$ is uniformly close to a function that is strongly convex everywhere and apply the convergence result for Langevin dynamics in the unconstrained setting to obtain our main result for PLD. The analysis for PULMC is similar but requires an additional technical result showing Hessian Lipschitzness. 
\item We then consider the setting of smooth $f$ that can be non-convex subject to stochastic gradients. For the unconstrained sampling of $\pi_\delta$, when the gradients of $f$ are estimated from a randomly selected subset of data; the variance of the noise is not uniformly bounded over $x$ but instead can grow linearly in $\|x\|^2$ (see e.g. \citet{jain2018accelerating,assran2020convergence}). Therefore, unlike the existing works for constrained sampling, we do not assume the variance of the stochastic gradient to be uniformly bounded but allow the gradient noise variance to grow linearly. For PSGLD and PSGULMC, we show an iteration complexity that scales as $\tilde{\mathcal{O}}(d^{17}/\lambda_{*}^{9})$ and $\tilde{\mathcal{O}}(d^{7}/\mu_{\ast}^{3})$ respectively in dimension $d$, where $\lambda_*$ and $\mu_*$ are constants that relate to overdamped and underdamped Langevin SDEs and will be defined later in \eqref{defn:lambda:ast} and \eqref{def-mu-star-0}. These constants can scale exponentially in the dimension in the worst case (due to hardness of the non-convex setting) but can also be independent of the dimension (see Section 4 in \cite{Raginsky}).
Our iteration complexity bounds for PSGLD and PSGULMC also scale polynomially in $\varepsilon$ (see Table~\ref{tab:methods} for the details).\footnote{In Table~\ref{tab:methods}, we used various metrics TV, $\mathcal{W}_{1}$, $\mathcal{W}_{2}$ and KL to measure the complexity and it is worth noting that they may scale differently. In general, it is always true that $\mathcal{W}_{1}\leq\mathcal{W}_{2}$ and $\text{TV}\leq\mathcal{O}(\sqrt{\text{KL}})$ (Pinsker's inequality). On the other hand, $\mathcal{W}_{2}\leq\mathcal{O}(\sqrt{\text{KL}})$ (Otto and Villani Theorem) if a log-Sobolev inequality is satisfied and more generally $\mathcal{W}_{2}\leq\mathcal{O}(\sqrt{\text{KL}}+(\text{KL})^{1/4})$ (\cite{BV}).} To our best knowledge, these are the first results for ULMC 
algorithms in the constrained setting for general $f$ that can be non-convex. Compared to \citet{pmlr-v134-lamperski21a}, our dimension dependency is worse, but our noise assumption is more general, and we do not require sub-Gaussian noise.  
To achieve these results, in addition to bounding the difference between $\pi_{\delta}$ and $\pi$, we show that $f+S/\delta$ satisfies a dissipativity condition, which is the key technical result, that allows us to apply
the convergence results in the literature for unconstrained Langevin algorithms with stochastic gradients where the target is non-convex and satisfies a dissipativity condition. Here, we also note that the standard penalty function we choose involves computing the distance of a point to the boundary of the constraint set. This is also the case for many algorithms in the literature, such as projected SGLD methods. However, often the set $\mathcal{C}$ is defined with convex constraints, i.e. $\mathcal{C}:=\{ x: h_i(x) \leq 0, i=1,2,\dots,m\}$ where $h_i:\mathbb{R}^d \to \mathbb{R}$ are convex and $m$ is the number of constraints. In this case, we discuss
in Section \ref{subsec-avoiding-projections} that the projections can be avoided when $h_i(x)$ satisfies some growth conditions.
\item When $f$ is strongly convex and smooth, we obtain iteration complexity of $\tilde{\bigO}(d/\varepsilon^{18})$ and $\tilde{\bigO}(d\sqrt{d}/\varepsilon^{39})$ for PSGLD and PSGULMC respectively.  To achieve these results, in addition to bounding the difference between $\pi_{\delta}$ and $\pi$, we also extend the existing result in the unconstrained setting for ULMC with a deterministic gradient
to allow stochastic gradient for strongly convex and smooth $f$, which is of independent interest. 
\end{itemize}
The summary of our main results and their comparison with respect to most closely-related 
approaches are given in Table~\ref{tab:methods}, where in our results it is assumed that the constraint set is compact and convex. We also note that when dealing with target densities where $f$ is smooth but non-convex, the literature
typically assumes growth conditions towards infinity such as dissipativity or isoperimetric inequalities \citep{Raginsky,GGZ,jiang2021mirror}, but in our results we do not require such a condition. This is due to the fact that the constraint set is taken to be a convex body which is a compact set where the growth of $f$ can be controlled.
\begin{small}
\begin{table}[]
\begin{tabular}{|l|l|l|l|l|l|l|}
\hline
Algorithms  & \begin{tabular}{@{}l@{}}Assump. \\ on $f$\end{tabular} & \begin{tabular}{@{}l@{}}Assump. \\ on $\mathcal{C}$\end{tabular} & \begin{tabular}{@{}l@{}}Stoc. \\ grad.\end{tabular} & \begin{tabular}{@{}l@{}}  Bdd. grad. \\ noise var.\textsuperscript{[5]} \end{tabular}  & \begin{tabular}{@{}c@{}}Conv. \\ meas.\end{tabular} & Complexity  \\ \hline

\begin{tabular}{@{}l@{}}Projected  LD \\  \tiny{\citep{bubeck2018sampling}} \end{tabular} & \begin{tabular}{@{}l@{}@{}}Convex,\\ Smooth, \\ Lipschitz\end{tabular}    & \begin{tabular}{@{}l@{}}Convex\\ body\end{tabular}    & No & N/A  & TV  & $\tilde{\bigO}\left(\frac{d^{12}}{\varepsilon^{12}}\right)$   \\ \hline

\begin{tabular}{@{}l@{}} Projected SGLD \\  \tiny{\citep{pmlr-v134-lamperski21a}} \end{tabular} & \begin{tabular}{@{}l@{}} Smooth, \\ Lipschitz\end{tabular}    & \begin{tabular}{@{}l@{}}Convex\\ body \end{tabular}   & Yes & Yes\textsuperscript{[6]}  & $\mathcal{W}_{1}$  & $\tilde{\bigO}\left(\frac{d^4}{\varepsilon^{4}}\right)$   \\ \hline

\begin{tabular}{@{}l@{}}Projected SGLD  \\  \tiny{\citep{zheng2022constrained}} \end{tabular}    & \begin{tabular}{@{}l@{}}Str. cvx.\\ outside\\ a ball,\textsuperscript{[1]} \\ Lipschitz \end{tabular} &\begin{tabular}{@{}l@{}} Polyhedral\\with 0 in \\interior \end{tabular}    & Yes & Yes          & $\mathcal{W}_{1}$   & $\tilde{\bigO}\left(\frac{d^2}{\varepsilon^2}\right)$   \\  \hline  

\begin{tabular}{@{}l@{}}Proximal SGLD \\  \tiny{\citep{SR2020}} \end{tabular} & Str. cvx.   & \begin{tabular}{@{}l@{}} Convex\textsuperscript{[2]} \end{tabular}    & Yes & Yes   & $\mathcal{W}_{2}$  & $\tilde{\bigO}\left(\frac{d}{\varepsilon^{2}}\right)$   \\ \hline

\begin{tabular}{@{}l@{}}Mirrored LD \\  \tiny{\citep{hsieh2018mirrored}} \end{tabular}  & Str. cvx.  & \begin{tabular}{@{}l@{}}Convex,\\ Bounded \end{tabular} & No & N/A    & $\mathcal{W}_{2}$              & $\tilde{\bigO}\left(\frac{d}{\varepsilon^2}\right)$   \\ \hline

\begin{tabular}{@{}l@{}}Mirrored SGLD \\  \tiny{\citep{hsieh2018mirrored}} \end{tabular}  & Str. cvx.  & \begin{tabular}{@{}l@{}}Convex,\\ Bounded \end{tabular} & Yes & Yes    & KL              & $\tilde{\bigO}\left(\frac{d^2}{\varepsilon^2}\right)$   \\ \hline

\begin{tabular}{@{}l@{}}MYULA  \\  \tiny{\citep{Brosse}} \end{tabular}     & \begin{tabular}{@{}l@{}}Convex,\\ Smooth \end{tabular}   & \begin{tabular}{@{}l@{}}Convex\\ body \end{tabular}     & No  &  N/A         & TV   & $\tilde{\bigO}\left(\frac{d^5}{\varepsilon^6}\right)$   \\ \hline

\begin{tabular}{@{}l@{}}PLD  \\  \tiny{Prop.~\ref{cor:LD:convex:outside} in our paper} \end{tabular} & Smooth &  \begin{tabular}{@{}l@{}}Convex\\ body \end{tabular}  & No & N/A & TV & $\tilde{\mathcal{O}}\left(\frac{d}{\varepsilon^{10}}\right)$ \\ \hline

\begin{tabular}{@{}l@{}}PULMC  \\  \tiny{Prop.~\ref{cor:HMC:convex:outside} in our paper} \end{tabular} &  \begin{tabular}{@{}l@{}} Smooth,\\Hessian\\Lipschitz \end{tabular} & \begin{tabular}{@{}l@{}}Convex\\body  $\ddagger$ \end{tabular}  & No & N/A & TV & $\tilde{\mathcal{O}}\left(\frac{\sqrt{d}}{\varepsilon^{7}}\right)$ \\ \hline

\begin{tabular}{@{}l@{}}PSGLD  \\  \tiny{Prop.~\ref{prop:SGLD:convex} in our paper} \end{tabular} & \begin{tabular}{@{}l@{}}Str. cvx.,\\ Smooth \end{tabular} & \begin{tabular}{@{}l@{}}Convex\\body \end{tabular}  & Yes & No & $\mathcal{W}_{2}$ & $\tilde{\bigO}\left(\frac{d}{\varepsilon^{18}}\right)$ \\ \hline

\begin{tabular}{@{}l@{}}PSGULMC  \\  \tiny{Prop.~\ref{prop:SGHMC:convex} in our paper} \end{tabular} & \begin{tabular}{@{}l@{}}Str. cvx.,\\ Smooth \end{tabular} & \begin{tabular}{@{}l@{}}Convex\\body \end{tabular}  & Yes & No & $\mathcal{W}_{2}$ & $\tilde{\bigO}\left(\frac{d\sqrt{d}}{\varepsilon^{39}}\right)$  \\ \hline

\begin{tabular}{@{}l@{}}PSGLD  \\  \tiny{Prop.~\ref{prop:SGLD:nonconvex} in our paper} \end{tabular} &  Smooth  &  \begin{tabular}{@{}l@{}}Convex\\body \end{tabular}  & Yes & No  & $\mathcal{W}_{2}$ & $\tilde{\mathcal{O}}\left(\frac{d^{17}}{\varepsilon^{392}\lambda_{*}^{9}}\right)$\textsuperscript{[3]} \\ \hline

\begin{tabular}{@{}l@{}}PSGULMC  \\  \tiny{Prop.~\ref{prop:SGHMC:nonconvex} in our paper} \end{tabular} & Smooth & \begin{tabular}{@{}l@{}}Convex\\body  \end{tabular}  & Yes & No & $\mathcal{W}_{2}$ & $\tilde{\mathcal{O}}\left(\frac{d^{7}}{\varepsilon^{132}\mu_{\ast}^{3}}\right)$\textsuperscript{[4]} \\ \hline

\end{tabular}
\caption{Comparison of our methods and existing methods.}
\begin{scriptsize}
$\ddagger$: $\mathcal{C}\subseteq \mathbb{R}^d$ is a convex hypersurface of class $C^{3}$ and $\sup_{\xi\in \mathcal{C}} \|D^2 n(\xi) \|$ is bounded, where $n$ is the unit normal vector of $\mathcal{C}$. {\scriptsize
\textsuperscript{[1]} ``Str. cvx.'' stands for ``Strongly convex''. Also, in \citet{zheng2022constrained}, it is assumed that $f$ is $\mu$-strongly convex outsize a Euclidean ball. 
\textsuperscript{[2]} \citet{SR2020} consider the situation, where the target distribution is $\pi \propto e^{-V(x)}$ with $V(x):=f(x)+G(x)$. Function $G$ is assumed to be nonsmooth and convex, and if $G$ is the indicator function of $\mathcal{C}$, then proximal SGLD can sample from the constrained distribution.
\textsuperscript{[3]} $\lambda_{*}$ is the spectral gap of penalized overdamped Langevin SDE \eqref{penalized:overdamped:SDE} which is defined in \eqref{defn:lambda:ast}. 
\textsuperscript{[4]} $\mu_{*}$ is the convergence speed of penalized underdamped Langevin SDE \eqref{eq:VL:penalized}-\eqref{eq:XL:penalized}  defined in \eqref{def-mu-star-0}.
\textsuperscript{[5]} This column specifies whether the methods assume that the gradient noise variance is uniformly bounded or not. 
\textsuperscript{[6]} The gradient noise is assumed to have a uniform sub-Gaussian property. 
}
\end{scriptsize}
\label{tab:methods}
\end{table}
\end{small}
\subsection{Related Work} 

Mirror-descent Langevin algorithms can be viewed as a special case of Riemannian Langevin that can be used to sample from some subset $D\subseteq\mathbb{R}^{d}$
by endowing $D$ with a Riemannian structure \citep{GC2011,patterson-teh}.
Geodesic Langevin algorithm is proposed in \citet{WLP2020} that can sample
a distribution supported on a manifold $M$.
They showed that geodesic Langevin algorithm can sample
a target distribution on a $d$-dimensional compact manifold $M$ without boundary
that satisfies a log-Sobolev inequality with parameter $\alpha$ with $\varepsilon$ accuracy in KL divergence after $\mathcal{O}(\frac{d}{\alpha^{2}\epsilon}\log(1/\varepsilon))$ iterates.
More recently, \citet{GV2022} showed that the Riemannian Langevin algorithm
converges to the target 
that satisfies a log-Sobolev inequality with parameter $\alpha$ with accuracy $\varepsilon$ in KL divergence
after $\mathcal{O}(\frac{d^{5/2}}{\alpha^{2}\epsilon}\log(1/\varepsilon))$ iterates
where $d$ is the dimension for general Hessian manifolds that are second-order self-concordant
where the log-density is gradient and Hessian Lipschitz.
Very recently, \citet{Kook2022} used a Riemannian version of Hamiltonian Monte Carlo to sample ill-conditioned, non-smooth, constrained distributions that maintain sparsity where $f$ is convex. Given a self-concordant barrier function for the constraint set, they empirically demonstrated that they could achieve a mixing rate independent of smoothness and condition numbers.
Moreover, \citet{Chalkis} proposed reflective Hamiltonian Monte Carlo
based on reflected underdamped Langevin diffusion to sample from a strongly-log-concave
distribution restricted to a convex polytope. They showed that from a warm start, it 
mixes in $\tilde{\mathcal{O}}(\kappa d^{2}\ell^{2}\log(1/\varepsilon))$
steps for a well-rounded polytope, where $\kappa$ is the condition number of $f$,
and $\ell$ is an upper bound on the number of reflections. 

It is also worth mentioning that the idea of adding a penalty term
to the Langevin diffusion \eqref{eq:overdamped-2} has appeared in the recent literature 
but in a very different context \citep{KD2020}. By adding a penalty term to the Langevin diffusion
with the log-concave target, the resulting target becomes strongly log-concave, and
as the penalty term vanishes, \citet{KD2020} were able to obtain new convergence results
for sampling a log-concave target.


SGLD algorithms have been studied in the unconstrained setting in a number of papers under various assumptions for $f$. Among these, we discuss closely related works. \citet{DK2017} study the convergence of SGLD for strongly convex smooth $f$. In a seminal work, \citet{Raginsky} show that when $f$ is non-convex and smooth, under a dissipativity condition, SGLD iterates track the overdamped Langevin SDE closely and obtained finite-time performance bounds for SGLD. More recently, \citet{pan-gu-langevin} improve the $\varepsilon$ dependency of the upper bounds of \citet{Raginsky} in the mini-batch setting and obtained several guarantees for the gradient Langevin dynamics and variance-reduced SGLD algorithms. \citet{zou2021faster} improve the existing convergence guarantees of SGLD for unconstrained sampling, showing that $\bigO(d^4\varepsilon^{-2})$ stochastic gradient evaluations suffice for SGLD to achieve $\varepsilon$-sampling accuracy in terms of the TV distance for a class of distributions that can be non-log-concave. They further show that provided an additional Hessian Lipschitz condition on the log-density function, SGLD is guaranteed to achieve $\varepsilon$-sampling error within $\bigO(d^{15/4}\varepsilon^{-3/2})$ stochastic gradient evaluations. There have also been more recent works on SGLD algorithms
that allow dependent data streams \citep{Barkhagen2021,Chau2021} and require
weaker assumptions on the target density \citep{Zhang2019}. \citet{Rolland2020} study a new annealing stepsize schedule for Unadjusted Langevin Algorithm (ULA) and they improve the convergence rate to $\bigO(d^3/T^{\frac{2}{3}})$ for unconstrained log-concave distribution, where $d$ is the dimension
and $T$ is the number of iterates. They also apply the double-loop approach to the constrained sampling algorithm Moreau-Yoshida ULA (MYULA) from \citet{Brosse}. They improve the convergence rate to $\bigO(d^{3.5}/\varepsilon^5)$ in the total variation distance for constrained log-concave distributions. \citet{Lan2016} propose a spherical augmentation method to sample constrained probability distributions by mapping the constrained domain to a sphere in the augmented space. 
 Several other works have also studied SGULMC algorithms in the unconstrained setting. \citet{zou2021convergence} propose a general framework for proving the convergence rate of Hamiltonian Monte Carlo with stochastic gradient estimators for sampling from strongly log-concave and log-smooth target distributions in the unconstrained setting. They show that the convergence to the target distribution in the 2-Wasserstein distance can be guaranteed as long as the stochastic gradient estimator is unbiased and its variance is upper-bounded along the algorithm trajectory.
 
\citet{lehec2021langevin} considers the projected Langevin algorithms and improves upon the work of \cite{bubeck2018sampling}. The author considers the constrained sampling case when the potential $f$ is a convex function that is Lipschitz on a convex constraint set $\mathcal{C}\subseteq \mathbb{R}^d$. 
In this setting, \citet{lehec2021langevin} obtains an upper bound on the discretization error between the iterates $x_k$ of the projected Langevin algorithm and its corresponding points in the Langevin diffusion based on the $\mathcal{W}_2$ distance \cite[Thm 1]{lehec2021langevin}. Using this bound, under the additional assumptions that the target $\pi$ satisfies a log-Sobolev inequality with constant $C_{LS}$ and the initial iterate $x_0$ is a point in the support of $\pi$, a bound on the $\mathcal{W}_2$ distance between the law of the iterates and the target is proven \cite[Thm 2]{lehec2021langevin}. Assuming further that the initial iterate $x_0$ is such that $\sigma_0  := f(x_0) - \min_x f(x) = \mathcal{O}(1)$, 
the latter result implies that $\mathcal{W}_2(\mathcal{L}(x_k),\pi) \leq \varepsilon $ after $k = \Theta^*\left(\frac{C_{LS}^3 d^2}{\varepsilon^4} \max\left(\frac{d}{r_0^2}, \frac{L_f^2}{d}\right)\right)$ iterations, where $\mathcal{L}(x_k)$ denotes the law of the $k$-th iterate $x_{k}$, where $L_f$ is the Lipschitz constant of $f$ on $\mathcal{C}$, $r_0$ is the distance of initial point $x_0$ to the boundary of $\mathcal{C}$, with the convention that $\Theta^*$ hides universal constants and possible $\mbox{polylog}(d)$ dependencies. Here, when $f$ is strongly convex and when the constraint set $\mathcal{C}$ 
is bounded, as discussed in \citet{lehec2021langevin}, we can take $C_{LS} = \frac{1}{\mu}$ where $\mu$ is the strong convexity constant of $f$. 
Also, when the constraint set is a ball of radius $R$ and when the target measure $\pi(x) \propto e^{-f(x)}$ is isotropic in the sense that its covariance matrix is the identity matrix, then we can take $C_{LS}$ to be $R$ up to a universal constant where by the isotropy condition it holds that $R\geq \sqrt{d}$ \citep{lehec2021langevin}. 
Some convex choices of $f$ may not necessarily satisfy the log-Sobolev inequality, but they do satisfy the Poincar\'{e} inequality for some finite constant $C_P$. For convex $f$ (that does not necessarily satisfy the log-Sobolev inequality), \citet{lehec2021langevin} also obtains Wasserstein bounds between the iterates and the target (that depends on the Poincar\'{e} constant $C_P$) when $f$ is globally Lipschitz on the domain $\mathcal{C}$ under a warm-start strategy where the initialization $x_0$ is taken as a random point taking values in $\mathcal{C}$ whose chi-square divergence to $\pi$ is finite \cite[Thm 2]{lehec2021langevin}. This result is applicable to the case when the constraint set $\mathcal{C}$ is unbounded, and when $\sigma_0 = \mathcal{O}(1)$ and all the other parameters are at most polynomial in $d$, it implies in the unconstrained case that $\mathcal{W}_2(\mathcal{L}(x_k),\pi) \leq \varepsilon $ after $k =\Theta^*\left(\frac{C_P^3 L_f^2 d^4}{\varepsilon^4}\right)$ iterations. Compared to \cite{lehec2021langevin}, when $f$ is strongly convex, we can get a better dimension dependency but our dependency on $\varepsilon $ is worse. We can also allow $f$ to be non-convex as long as it is smooth and our analysis can support stochastic gradients for both overdamped and underdamped dynamics; however, we require the constraint set $\mathcal{C}$ to be bounded.

In a recent work, \citet{sato2022convergence} considers the problem of constrained sampling when the potential $f$ is $C^4$ with a Lipschitz gradient on the constraint set and when the constraint set $\mathcal{C}$ has a smooth $(C^4)$ boundary, allowing it to be non-convex. The authors also assume that the projection to set $\mathcal{C}$ is unique and that the projections can be efficiently computed where they study a reflection-based overdamped Langevin algorithm that can be viewed as a discretization of a reflected Langevin diffusion, assuming access to (non-stochastic) exact gradients of $f$. To compute the reflections, their algorithm necessitates to compute projections at every step. The authors show that the optimization error converges to the target distribution and it suffices to have $\tilde{\mathcal{O}}(\frac{d^3}{\lambda_r \varepsilon^3})$ iterations for the suboptimality to be at most $\varepsilon$ in expectation where $\lambda_r$ is the spectral gap of the reflected Langevin diffusion. In our paper, we require $\mathcal{C}$ to be convex but its boundary can be non-smooth. For the overdamped Langevin version of our algorithm which we call PLD, we require $\tilde{\mathcal{O}}(\frac{d}{\varepsilon^{10}})$ iterations which is better dependency to the dimension when $\mathcal{C}$ is convex; furthermore we can avoid projections and therefore we do not necessarily require the projections to be efficiently computable, in addition we do not necessarily require a smooth boundary. Moreover, our results can also handle underdamped dynamics and stochastic gradients which are key to handle machine learning applications, whereas the stochastic gradient setting is not considered in \citet{sato2022convergence}.

Finally, we note that ``hit-and-run walk" achieves a mixing time of $\tilde{\bigO}(d^4)$ iterations \citep{hit-and-run}.
However, they assume a ``zeroth order oracle", i.e., assuming access to function values without access to its gradients. Thus, our setting is different where we work with gradients.

The notations to be used in the rest of the paper are summarized in Appendix~\ref{sec:notations}.

\section{Main Results}\label{sec:main}

Penalty methods in optimization convert a constrained optimization problem to an unconstrained one, where the idea is to add a term to the optimization objective that penalizes for being outside of the constraint set \citep{nocedal1999numerical}. Motivated by such methods, as discussed in the introduction, we propose to add a penalty term $\frac{1}{\delta}S(x)$
to the target distribution, and sample instead from the penalized target distribution in an unconstrained fashion with the modified target density:
\begin{equation}
\pi_{\delta}(x)\propto\exp\left(-f(x)-\frac{1}{\delta}S(x)\right),\qquad x\in\mathbb{R}^{d},
\end{equation}
where $S(x)$ is the penalty function that satisfies the following assumption and $\delta>0$ is an adjustable parameter.

\begin{assum}\label{assump:S:0}
Assume that $S(x)=0$ for any $x\in\mathcal{C}$
and $S(x)>0$ for any $x\notin\mathcal{C}$.
\end{assum}

There are many simple choices of $S(x)$ for which Assumption~\ref{assump:S:0} is satisfied. For instance, if we choose $S(x)=g(\delta_{\mathcal{C}}(x))$, where $\delta_{\mathcal{C}}(x)=\min_{c\in\mathcal{C}}\Vert x-c\Vert$
is the distance of the point $x$ to a closed set $\mathcal{C}$ and $g:\mathbb{R}_{\geq 0}\rightarrow\mathbb{R}_{\geq 0}$ is a strictly increasing function with $g(0)=0$, then Assumption~\ref{assump:S:0} is satisfied. Throughout our paper, we will also discuss other choices of $S(x)$. In many of our results, we will also make the following assumption on the set $\mathcal{C}$.

\begin{assum}\label{assump:C}
Assume that $\mathcal{C}$ is a convex body, i.e., $\mathcal{C}$ is a compact convex set, contains an open ball centered at $0$ with radius $r>0$, and is contained in a Euclidean ball centered at $0$ with radius $R>0$.
\end{assum}

The fact that 0 is in the set $\mathcal{C}$ in Assumption~\ref{assump:C} is made for simplifying the presentation but all our results will hold  even if that is not the case. Assumption~\ref{assump:C} has been commonly made in the literature \citep{bubeck2018sampling,bubeck2015finite,pmlr-v134-lamperski21a,Brosse}. In addition, for many applications including those arise in machine learning, this assumption naturally holds; for instance, when the constraints are polyhedral \citep{Kook2022} or when the constraints are $\ell_p$-norm constraints with $p\geq 1$ or for $p=\infty$ \citep{schmidt2005least,luo2016regression,ma2019transformed,gurbuzbalaban2022stochastic}. 

\subsection{Bounding the Distance Between $\pi_{\delta}$ and $\pi$}

In this section, we aim to bound the 2-Wasserstein distance between the modified target $\pi_{\delta}$
and the target $\pi$ with an explicitly computable upper bound that goes to zero
as $\delta$ tends to zero.
We will first bound the KL divergence between $\pi_{\delta}$ and $\pi$
and then apply weighted Csisz\'{a}r-Kullback-Pinsker inequality (W-CKP) (see Lemma~\ref{lem:BV})
to bound the 2-Wasserstein distance between $\pi_{\delta}$
and $\pi$.
To start with, we first bound the KL divergence between $\pi_{\delta}$
and $\pi$, which relies on a series of technical lemmas. 
The two main ideas are: (i) when the penalty value $S$ is small, 
the Lebesgue measure of the set with small penalty values
is also small so that its contribution is negligible; (ii) for small values of $\delta$, 
the penalty $\frac{S}{\delta}$ is large and the integral
with respect to that is also negligible. 
We start with the following lemma. The proofs of this lemma and our other results can be found in the appendix.
 
\begin{lemm}\label{lem:1}
Suppose Assumption~\ref{assump:S:0} holds
and $e^{-f}$ is integrable over $\mathcal{C}$.
For any $\delta>0$,\footnote{If $e^{-\frac{1}{\delta}S(y)-f(y)}$ is not integrable over $\mathbb{R}^{d}\backslash\mathcal{C}$, 
we take the term $\int_{\mathbb{R}^{d}\backslash\mathcal{C}}e^{-\frac{1}{\delta}S(y)-f(y)}dy$ to be $\infty$ as the convention and the upper bound in equation \eqref{first:upper:bound} becomes trivial.}
\begin{align}\label{first:upper:bound}
D(\pi\Vert\pi_{\delta})
\leq\frac{\int_{\mathbb{R}^{d}\backslash\mathcal{C}}e^{-\frac{1}{\delta}S(y)-f(y)}dy}{\int_{\mathcal{C}}e^{-f(y)}dy}.
\end{align}
\end{lemm}

Next, we provide a technical lemma that provides an upper bound
on the Lebesgue measure of the set with small penalty values $S$.
A special case of the following lemma
can be found in Lemma~10.15 without a proof in \citet{Kallenberg}.\footnote{Note that Lemma~10.15 in \citet{Kallenberg} requires
the set $\mathcal{C}$ to be convex since
it estimates both the outer $\epsilon$-collar of $\mathcal{C}$, 
defined as the set of all points that do not belong to $\mathcal{C}$ but lie
within distance at most $\epsilon$ from it, 
as well as the inner $\epsilon$-collar of $\mathcal{C}$, 
whereas we only need to consider the outer $\epsilon$-collar of $\mathcal{C}$
so that we can remove the convexity assumption on the set $\mathcal{C}$.} 

\begin{lemm}\label{lem:geometry}
Assume the constraint set $\mathcal{C}$ is a bounded closed set containing an open ball with radius $r>0$. 
Let $S(x)=g\left(\delta_{\mathcal{C}}(x)\right)$, 
where $\delta_{\mathcal{C}}(x)=\min_{c\in\mathcal{C}}\Vert x-c\Vert$
is the distance of the point $x$ to the set $\mathcal{C}$ and $g:\mathbb{R}_{\geq 0}\rightarrow\mathbb{R}_{\geq 0}$ is a strictly increasing function with $g(0)=0$ with the property $g(x)\to\infty$ as $x\to\infty$. 
Then, for any $\epsilon>0$, 
\begin{equation}
\left|x\in\mathbb{R}^{d}\backslash\mathcal{C}: S(x)\leq\epsilon\right|
\leq\left(\left(1+\frac{g^{-1}(\epsilon)}{r}\right)^{d}-1\right)|\mathcal{C}|,
\end{equation}
where $\left|\cdot \right|$ denotes the Lebesgue measure and $g^{-1}$ is the inverse function of $g$.
\end{lemm}

We are now ready to provide an upper bound for $D(\pi\Vert\pi_{\delta})$, the KL divergence
between the target distribution $\pi$ and the penalized target distribution $\pi_{\delta}$.

\begin{lemm}\label{lem:D}
In the setting of Lemma~\ref{lem:geometry}, assume $e^{-f}$ is integrable over $\mathcal{C}$,
then for any $\delta,\tilde{\alpha}>0$, we have\footnote{If $\inf_{y\in\mathbb{R}^{d}\backslash\mathcal{C}:S(y)\leq\tilde{\alpha}\delta\log(1/\delta)}f(y)=-\infty$, 
we take the right hand side of equation \eqref{second:upper:bound} to be $\infty$ as the convention
and the upper bound in equation \eqref{second:upper:bound} becomes trivial.}
\begin{align}
&D(\pi\Vert\pi_{\delta})
\leq
\left(\left(1+\frac{g^{-1}(\tilde{\alpha}\delta\log(1/\delta))}{r}\right)^{d}-1\right)\frac{\frac{\pi^{d/2}}{\Gamma(\frac{d}{2}+1)}R^{d}e^{-\inf_{y\in\mathbb{R}^{d}\backslash\mathcal{C}:S(y)\leq\tilde{\alpha}\delta\log(1/\delta)}f(y)}}{\int_{\mathcal{C}}e^{-f(y)}dy}
\nonumber
\\
&\qquad\qquad\qquad
+\delta^{\tilde{\alpha}}\frac{\int_{\mathbb{R}^{d}\backslash\mathcal{C}}e^{-\frac{1}{\delta}S(y)-f(y)}dy}{\int_{\mathcal{C}}e^{-f(y)}dy},\label{second:upper:bound}
\end{align}
where $\Gamma$ denotes the gamma function.
\end{lemm}

In Lemma~\ref{lem:D}, we obtained an upper bound of the KL divergence between $\pi$ and $\pi_{\delta}$. 
In the literature of Langevin Monte Carlo, it is common to use the 2-Wasserstein distance to
measure the convergence to the target distribution \citep{Cheng,DK2017}.
The celebrated W-CKP inequality (see Lemma~\ref{lem:BV})
bounds the 2-Wasserstein distance by the KL divergence of any two probability
distributions where some exponential integrability condition is satisfied (see Lemma~\ref{lem:BV}),
which in our case can be applied to control the 2-Wasserstein distance between $\pi_{\delta}$ and $\pi$.
From Lemma~\ref{lem:geometry}, recall the function
$\delta_\mathcal{C}(x) = \mbox{distance}(x,\mathcal{C}):= \min_{c\in\mathcal{C}}\|x - c\|$,
for $x\in \mathbb{R}^d$.
The convexity of the set $\mathcal{C}$ implies that the function $S(x)=\left(\delta_{\mathcal{C}}(x)\right)^2$ satisfies some differentiability and smoothness properties, which is provided in the following lemma.

\begin{lemm}\label{lem:S}
If $\mathcal{C}$ is convex, then the function $S(x)=\left(\delta_{\mathcal{C}}(x)\right)^2$
is convex, $\ell$-smooth with $\ell = 4$ and continuously differentiable on $\mathbb{R}^d$ with a gradient $\nabla S(x) = 2 (x - \mathcal{P}_{\mathcal{C}}(x))$, where $\mathcal{P}_{\mathcal{C}}(x)$ is the projection of $x$ to the set $\mathcal{C}$, i.e. $P_\mathcal{C}(x) := \arg \min_{c\in\mathcal{C}}\| x- c\|$.
\end{lemm}

In the rest of the paper (except in Section~\ref{subsec-avoiding-projections}), we always take penalty function $S(x)=\left(\delta_{\mathcal{C}}(x)\right)^{2}$ unless otherwise specified.
Building on Lemma~\ref{lem:S}, we have the following result, which quantifies the 2-Wasserstein distance between the target $\pi$ and the modified target $\pi_\delta$ corresponding to the penalized target distribution.

\begin{theo}\label{thm:final}
Suppose Assumptions~\ref{assump:S:0} and \ref{assump:C} hold. 
Moreover, we assume that $f$ is continuous and $e^{-f}$ is integrable over $\mathcal{C}$ and there exist some $\hat{\alpha}>0$ and $\hat{x}\in\mathbb{R}^{d}$
such that and $\int_{\mathbb{R}^{d}}e^{\hat{\alpha}\Vert x-\hat{x}\Vert^{2}}e^{-\frac{S(x)}{\delta}-f(x)}dx<\infty$.
Then, as $\delta\rightarrow 0$,
\begin{equation}\label{final:W:2}
\mathcal{W}_{2}(\pi_{\delta},\pi)\leq\mathcal{O}\left(\delta^{1/8}\left(\log(1/\delta)\right)^{1/8}\right).
\end{equation}
\end{theo}

Theorem~\ref{thm:final} shows that by choosing $\delta$ small enough, we can approximate the compactly supported target distribution $\pi$ with the modified target $\pi_\delta$ which has full support on $\mathbb{R}^d$. This amounts to converting the problem of constrained sampling to the problem of unconstrained sampling with a modified target. In the next remark, we discuss that if we take $\mathcal{C}$ to be the closed ball and $g(x)=x^2$, and apply the W-CKP inequality, we obtain the same bound in \eqref{final:W:2} except the logarithmic factor. This shows that it is not possible to improve our bound with an approach that relies on W-CKP inequality except for logarithmic factors. 
\begin{rema}
In the setting of Theorem~\ref{thm:final}, consider the special case
$\mathcal{C} = \{ x : \|x\| \leq R\}$ to be the closed ball of radius $R$. In this case $S(x)=s(r)$, with $r=\Vert x\Vert$ and $s(r)=(r-R)^{2}1_{r\geq R}$,
where $s(r)=0$ for any $r\leq R$ and $s(r)>0$ for any $r>R$
and moreover $s$ is differentiable and $s(r)$ is strictly increasing in $r>R$.
Moreover, we assume that $f\geq 0$.
Then, by Lemma~\ref{lem:1} and using the spherical symmetry, we can compute that
\begin{equation}\label{bound:symmetric}
D(\pi\Vert\pi_{\delta})
\leq\frac{\int_{\mathbb{R}^{d}\backslash\mathcal{C}}e^{-\frac{1}{\delta}S(y)}dy}{\int_{\mathcal{C}}e^{-f(y)}dy}
=\frac{\int_{\Vert y\Vert\geq R}e^{-\frac{1}{\delta}s(\Vert y\Vert)}dy}{\int_{\Vert y\Vert<R}e^{-f(y)}dy}
=\frac{\int_{r\geq R}e^{-\frac{1}{\delta}s(r)}r^{d-1}dr}{\int_{\Vert y\Vert<R}e^{-f(y)}dy}.
\end{equation}
Since $s(r)$, for $r\geq R$, achieves the unique minimum at $r=R$
and $s'(R)=0$, we can apply Laplace's method (see e.g. \citet{BH2010}), and obtain
\begin{equation}\label{laplace:expansion}
\int_{r\geq R}e^{-\frac{1}{\delta}s(r)}r^{d-1}dr
=\sqrt{\frac{\pi}{2s''(R)}}R^{d-1}\sqrt{\delta}\cdot(1+o(1)),
\qquad
\text{as $\delta\rightarrow 0$}.
\end{equation}
Therefore, it follows from \eqref{bound:symmetric}
and \eqref{laplace:expansion} that for any sufficiently small $\delta>0$, 
\begin{equation}
D(\pi\Vert\pi_{\delta})
\leq\left(\frac{\sqrt{\frac{\pi}{2s''(R)}}R^{d-1}}{\int_{\Vert y\Vert<R}e^{-f(y)}dy}\right)\sqrt{\delta}.
\label{eq:d-example2}
\end{equation}
By applying W-CKP inequality (see Lemma~\ref{lem:BV}) and \eqref{eq:d-example2}, we conclude 
$\mathcal{W}_{2}(\pi_{\delta},\pi)\leq\mathcal{O}\left(\delta^{1/8}\right)$.
\end{rema}

\subsection{Penalized Langevin Algorithms with Deterministic Gradient}\label{sec:deterministic}

In this section, we are interested in penalized Langevin algorithms with deterministic gradient when $f$ is non-convex.  
\citet{Raginsky} and \citet{GGZ} developed non-asymptotic convergence bounds for SGLD and SGULMC, respectively, when $f$ belongs to the class of non-convex smooth functions that are dissipative. This is a relatively general class of non-convex functions that admit critical points on a compact set. 
In our case, since $\mathcal{C}$ is assumed to be a compact convex set, we will not need growth conditions such as the dissipativity of $f$. The only assumption we are going to make about $f$ is that
$f$ is smooth, i.e. the gradient of $f$ is Lipschitz.
We will show that the penalty function $S$ is dissipative and smooth, so that $f+\frac{1}{\delta}S$ is dissipative and smooth for $\delta>0$ small enough.

\begin{assum}\label{assump:f:2}
Assume that $f$ is $L$-smooth, i.e.
$\left\Vert\nabla f(x)-\nabla f(y)\right\Vert\leq L\Vert x-y\Vert,$
for any $x,y\in\mathbb{R}^{d}$.
\end{assum}

If Assumption~\ref{assump:f:2} and Assumption~\ref{assump:C} hold, 
then the conditions in Theorem~\ref{thm:final} are satisfied (see Lemma~\ref{condition:thm:final} in the Appendix for details). Building on this result, next we derive iteration complexity corresponding to the penalized Langevin dynamics.

\subsubsection{Penalized Langevin Dynamics}\label{subsubsec-penal_langevin}
First, we introduce the penalized overdamped Langevin SDE:
\begin{equation}\label{penalized:overdamped:SDE}
dX(t)=-\nabla f(X(t))dt-\frac{1}{\delta}\nabla S(X(t))dt+\sqrt{2}dW_{t},
\end{equation}
where $W_{t}$ is a standard $d$-dimensional Brownian motion,
and under mild conditions, it admits a unique stationary distribution
$\pi_{\delta}(x)\propto\exp\left(-f(x)-\frac{1}{\delta}S(x)\right)$; see e.g. \citet{herau-nier-underdamped,pavliotis2014stochastic}.
Consider the penalized Langevin dynamics (PLD):
\begin{equation}\label{eq:LD:nonconvex}
x_{k+1}=x_{k}-\eta\left(\nabla f(x_{k})+\frac{1}{\delta}\nabla S(x_{k})\right)+\sqrt{2\eta}\xi_{k+1},
\end{equation}
where $\xi_{k}$ are i.i.d. $\mathcal{N}(0,I_{d})$ Gaussian noises in $\mathbb{R}^{d}$.

In many applications, the constrained set $\mathcal{C}$ is defined by functional constraints, i.e. 
$\mathcal{C}:=\{x: h_i(x)\leq 0, i=1,2,\dots,m\},
$ where $h_i$ is a (merely) convex function defined on an open set that contains $\mathcal{C}$ and $m$ is the number of constraints. For example, when $\mathcal{C}$ is an $\ell_p$ ball with radius $R$ with $p\geq 1$ or when $\mathcal{C}$ is an ellipsoid. In this case, we can write 
\begin{equation}\mathcal{C}:=\{x: h(x)\leq 0\}
\label{def-constraint-set-with-ineq},
\end{equation}
where $h(x):=\max_i h_i(x)$ is convex and therefore locally Lipschitz continuous (see e.g. \citet{cvx-locally-lip}). The choice of the $h(x)$ function here is clearly not unique. In fact, such an $h(x)$ can be constructed even if we do not possess an explicit formula for the functions $h_i(x)$.
More specifically, Minkowski functional $\Vert\cdot\Vert_{K}$, also known as the gauge function, is
defined as $\Vert x\Vert_{K}:=\inf\{t\geq 0,x\in t\mathcal{C}\}$ 
such that given that $0$ is in the interior of $\mathcal{C}$, we can write $\mathcal{C}:=\{x:h(x)\leq 0\} \quad
\mbox{where} \quad h(x)=\Vert x\Vert_{K}-1$ \citep{rockafellar1970convex,Minkowski}. 
It is also well-known that the gauge function is merely convex. Thus, we can conclude that any convex body $\mathcal{C}$ admits the representation \eqref{def-constraint-set-with-ineq} where $h(x)$ is convex and finite-valued and therefore Lipschitz continuous on $\mathcal{C}$ \citep{cvx-locally-lip}. 
Equipped with this representation given by \eqref{def-constraint-set-with-ineq}, we now consider a regularized constraint set 

\begin{equation}\label{C:alpha}
\mathcal{C}^{\alpha} = \{ x: h^{\alpha}(x)\leq 0\}, \quad
\quad \mbox{where} \quad h^{\alpha}(x):=h(x)+\frac{\alpha}{2}\Vert x\Vert^{2},
\end{equation}
is $\alpha$-strongly convex for $\alpha>0$ as $h(x)$ is merely convex, and it holds that $\mathcal{C}^{\alpha}\subseteq\mathcal{C}\subseteq\mathbb{R}^{d}$. We define the regularized distribution $\pi^{\alpha}$ supported on $\mathcal{C}^{\alpha}$ with 
probability density function
\begin{equation}
\pi^{\alpha}(x)\propto\exp(-f(x)),\qquad x\in\mathcal{C}^{\alpha}.
\end{equation}
We also consider adding a penalty term $\frac{1}{\delta}S^{\alpha}(x)$
to the regularized target distribution $\pi^{\alpha}$, and sample instead from the ``penalized target distribution" with the regularized target density:
\begin{equation}
\pi_{\delta}^{\alpha}(x)\propto\exp\left(-f(x)-\frac{1}{\delta}S^{\alpha}(x)\right),\qquad x\in\mathbb{R}^{d},
\end{equation}
where $S^\alpha(x)=\left(\delta_{\mathcal{C}^\alpha}(x)\right)^2$ is the penalty function that satisfies
$S^{\alpha}(x)=0$ for any $x\in\mathcal{C}^{\alpha}$ and $S^{\alpha}(x)>0$ otherwise. Our motivation for considering the penalty function 
$S^\alpha(x)=\left(\delta_{\mathcal{C}^\alpha}(x)\right)^2$ 
is that as we show in the Appendix, under some conditions,
$S^{\alpha}$ is strongly convex outside a compact
set (Lemma~\ref{lemm:strongly convex}); it can be seen that the function $S(x)=\left(\delta_{\mathcal{C}}(x)\right)^2$ does not always have this property.\footnote{For example, when $\mathcal{C}$ is the unit $\ell_\infty$ ball in dimension $2$, the function $S$ is not strongly convex at any point $(0,y)$ for $y\in \mathbb{R}$. } 
Consequently, as a corollary, the function $f + \frac{1}{\delta}S^{\alpha}$ becomes strongly convex outside a compact set for $\delta$ small enough (Corollary~\ref{cor:strongly:convex}). Our main result in this section builds on exploiting this structure to develop stronger iteration complexity results for sampling $\pi_{\delta}^{\alpha}$ compared to sampling $\pi$ directly, and then controlling the error between $\pi_{\delta}^{\alpha}$ and $\pi$ by choosing $\delta$ and $\alpha$ small enough appropriately. For this purpose, first  
we estimate the size of the set difference $\mathcal{C}\backslash\mathcal{C}^{\alpha}$. 

\begin{lemm}\label{lemma:O:alpha}
For the constrained set $\mathcal{C}^{\alpha}$ defined in \eqref{C:alpha}, we have
\begin{equation}\label{C:alpha:assump}
\frac{|\mathcal{C}\backslash\mathcal{C}^{\alpha}|}{|\mathcal{C}^{\alpha}|}
\leq\mathcal{O}(\alpha),\qquad\text{as $\alpha\rightarrow 0$}.
\end{equation}
\end{lemm}

Second, we show that there exists
a function $U$ that is strongly convex everywhere
and the difference between $U$
and $f+S^{\alpha}/\delta$ can be uniformly bounded (Lemma~\ref{lem:close-piecewise-quadratic}).
Then by combining all these technical results (Lemma~\ref{lemm:strongly convex} and Corollary~\ref{cor:strongly:convex}, Lemma~\ref{lem:close-piecewise-quadratic}, Lemma \ref{lemma:O:alpha}) discussed above, and estimating the distance of $\pi_{\delta}^{\alpha}$ to $\pi$,
we obtain
the following result.

\begin{prop}\label{cor:LD:convex:outside}
Suppose Assumptions~\ref{assump:S:0}, \ref{assump:C}, and \ref{assump:f:2} hold. Given the constraint set $\mathcal{C}$, consider its representation as $\mathcal{C} = \{ x: h(x)\leq 0\}$ given in \eqref{def-constraint-set-with-ineq} where $h(x)=\max_{1\leq i \leq m} h_i(x)$ for some $m\geq 1$ with $h_i$ convex for $i=1,2,\dots,m$. Let $\nu_{K}$ be the distribution of the $K$-th iterate $x_{K}$
of penalized Langevin dynamics \eqref{eq:LD:nonconvex} with the constrained set $\mathcal{C}^{\alpha}$
that is defined in \eqref{C:alpha}
and the initialization $\nu_{0}=\mathcal{N}(0,\frac{1}{L_{\delta}}I_{d})$,
where we take $\alpha=0$ if $h$ is strongly convex 
and we take $\alpha=\varepsilon^{2}$ is $h$ is merely convex.
Then, we have $\text{TV}(\nu_{K},\pi)\leq\tilde{\mathcal{O}}(\varepsilon)$
provided that $\delta=\varepsilon^{4}$ and
\begin{equation}
K=\tilde{\mathcal{O}}\left(d/\varepsilon^{10}\right),
\qquad
\eta=\mathcal{O}\left(\varepsilon^{10}/d\right),
\end{equation}
where $\tilde{\mathcal{O}}$ ignores the dependence on $\log d$ and $\log(1/\varepsilon)$.
\end{prop}

\begin{rema}
In Proposition~\ref{cor:LD:convex:outside}, when $h$ is $\beta$-strongly convex (with $\beta>0$ and $\alpha=0$)
the leading-order complexity $K=\tilde{\mathcal{O}}\left(\frac{d}{\varepsilon^{10}}\right)$
does not depend on $\beta$. It can be seen from from the proof of Proposition~\ref{cor:LD:convex:outside}
that the complexity $K$ has a second-order dependence on $\beta$, such that
$K=\tilde{\mathcal{O}}\left(\frac{d}{\varepsilon^{10}}\right)+\tilde{\mathcal{O}}\left(\frac{d}{\beta\varepsilon^{6}}\right)$,
where we ignored the dependence on the other constants when
we consider the second-order dependence on $\beta$. When $h$ is merely convex (with $\beta=0$ and $\alpha=\varepsilon^{2}$), we have $K=\tilde{\mathcal{O}}\left(\frac{d}{\varepsilon^{10}}\right)+\tilde{\mathcal{O}}\left(\frac{d}{\varepsilon^{8}}\right)$.
\end{rema}

\subsubsection{Penalized Underdamped Langevin Monte Carlo}

We can also design sampling algorithms based on the underdamped (also known as second-order, inertial, or kinetic) Langevin diffusion given by the following SDE:
\begin{align}
&dV(t)=-\gamma V(t)dt- \nabla f (X(t))dt+\sqrt{2\gamma }dW_{t}, \label{eq:VL}
\\
&dX(t)=V(t)dt, \label{eq:XL}
\end{align}
(see e.g. \citet{Cheng,cheng-nonconvex,dalalyan2018kinetic,GGZ,GGZ2,Ma2019,JianfengLu}) where $\gamma>0$ is the friction coefficient, $X(t),V(t) \in \mathbb{R}^d$ model the position and the momentum of a particle moving in a field of force (described by the gradient of $f$) plus a random (thermal) force described by the Brownian noise $W_{t}$, which is a standard $d$-dimensional Brownian
motion that starts at zero at time zero. It is known that under some mild assumptions on $f$, the Markov process $(X(t), V(t))_{t\geq 0}$ is ergodic and admits a unique stationary distribution $\pi$ with density
$\pi(x,v) \propto \exp\left(- \left(\frac{1}{2} \Vert v\Vert^2 +f(x)\right) \right)$ (see e.g. \citet{herau-nier-underdamped,pavliotis2014stochastic}).
Hence, the $x$-marginal distribution of the stationary distribution with the density $\pi(x,v)$ is exactly the invariant distribution of the overdamped Langevin diffusion. For approximate sampling, 
various discretization schemes of \eqref{eq:VL}-\eqref{eq:XL}
have been used in the literature (see e.g. 
\citet{cheng-nonconvex,teh2016consistency,chen2016bridging,chen2015convergence}).

To design a constrained sampling algorithm based on the underdamped Langevin diffusion, 
we propose the ``penalized underdamped Langevin SDE":
\begin{align}
&dV(t)=-\gamma V(t)dt- \nabla f (X(t))dt-\frac{1}{\delta}\nabla S(X(t))dt+\sqrt{2\gamma }dW_{t}, \label{eq:VL:penalized}
\\
&dX(t)=V(t)dt, \label{eq:XL:penalized}
\end{align}
where $W_{t}$ is a standard $d$-dimensional Brownian motion.
Under mild conditions, it admits a unique stationary distribution
$\pi_{\delta}(x,v)\propto\exp\left(-f(x)-\frac{1}{\delta}S(x)-\frac{1}{2}\Vert v\Vert^{2}\right)$,
whose $x$-marginal distribution is $\pi_{\delta}(x)\propto\exp\left(-f(x)-\frac{1}{\delta}S(x)\right)$,
which coincides with the stationary distribution of the penalized overdamped Langevin SDE \eqref{penalized:overdamped:SDE}.

A natural way to sample the penalized target distribution $\pi_{\delta}(x,v)$
is to consider the Euler discretization of \eqref{eq:VL:penalized}-\eqref{eq:XL:penalized}. 
We adopt a more refined discretization, introduced by \citet{Cheng}.
We propose the penalized underdamped Langevin 
Monte Carlo (PULMC):
\begin{align}
&v_{k+1}=\psi_{0}(\eta)v_{k}-\psi_{1}(\eta)\left(\nabla f(x_{k})+\frac{1}{\delta}\nabla S(x_{k})\right)+\sqrt{2\gamma}\xi_{k+1},\label{HMC:nonconvex:1}
\\
&x_{k+1}=x_{k}+\psi_{1}(\eta)v_{k}-\psi_{2}(\eta)\left(\nabla f(x_{k})+\frac{1}{\delta}\nabla S(x_{k})\right)+\sqrt{2\gamma}\xi'_{k+1},\label{HMC:nonconvex:2}
\end{align}
(see e.g. \citet{dalalyan2018kinetic}) where 
$(\xi_{k},\xi'_{k})$ are i.i.d. $2d$-dimensional Gaussian noises and independent of the initial condition $v_{0},x_{0}$,
and for any fixed $k$, the random vectors $((\xi_{k})_2,(\xi'_{k})_2),((\xi_{k})_2,(\xi'_{k})_2),\dots ((\xi_{k})_d,(\xi'_{k})_d)$ are i.i.d. with the covariance matrix
\begin{equation}\label{defn:cov}
C(\eta):=\int_{0}^{\eta}[\psi_{0}(t),\psi_{1}(t)]^{\top}[\psi_{0}(t),\psi_{1}(t)]dt,
\end{equation}
where 
\begin{equation}\label{defn:psi}
\psi_{0}(t):=e^{-\gamma t}\quad\text{and}\quad \psi_{k+1}(t):=\int_{0}^{t}\psi_{k}(s)ds \quad\text{for every $k\geq 0$}.
\end{equation}

\citet{dalalyan2018kinetic} studied the unconstrained kinetic (underdamped) Langevin Monte Carlo algorithms (subject to deterministic gradients) for strongly log-concave and smooth densities, and \citet{Ma2019} investigated the case when $f$ is strongly convex outside a compact domain. 
When $f$ is non-convex, 
\citet{GGZ} studied the unconstrained underdamped Langevin Monte Carlo algorithms (which allows stochastic gradients)
under a dissipativity assumption.
Since the $x$-marginal distribution of the Gibbs
distribution of the penalized underdamped Langevin SDE \eqref{eq:VL:penalized}-\eqref{eq:XL:penalized}
coincides that with the penalized overdamped Langevin SDE \eqref{penalized:overdamped:SDE},
we can bound $\mathcal{W}_{2}(\pi,\pi_{\delta})$
using Theorem~\ref{thm:final}
with the same bounds as in the overdamped case.

Under some additional assumptions, 
we showed in Lemma~\ref{lemm:strongly convex}
that $f+S/\delta$ is strongly convex outside a compact domain, and thus one can leverage
the non-asymptotic guarantees in \citet{Ma2019} for unconstrained underdamped 
Monte Carlo
to obtain better performance guarantees for the penalized underdamped Langevin Monte Carlo. 
Before we proceed, we first
provide a technical lemma
that shows that under some additional
assumptions on $\mathcal{C}$, $S$ is Hessian Lipschitz.

\begin{lemm}\label{lemma:Frobenius-lipschitz}
Suppose $\mathcal{C}\subseteq \mathbb{R}^d$ is a convex hypersurface of class $C^{3}$ and $\sup_{\xi\in \mathcal{C}} \|D^2 n(\xi) \|$ is bounded, where $n$ is unit normal vector of $\mathcal{C}$.
Then $S$ is $M_S$-Hessian Lipschitz for some $M_{S}>0$.
\end{lemm}

As a corollary, if $f$ is Hessian Lipschitz, 
then $f+S/\delta$ is Hessian Lipschitz and we 
immediately have the following result.

\begin{coro}\label{cor:Frobenius-lipschitz}
Under assumptions of Lemma~\ref{lemma:Frobenius-lipschitz} 
and assume that $f$ is $M_{f}$-Hessian Lipschitz
for some $M_{f}>0$. Then $f+S/\delta$ is
$M_{\delta}$-Hessian Lipschitz, 
where $M_{\delta}:=M_{f}+\frac{M_{S}}{\delta}$.
\end{coro}

Now, we are ready to state
the following proposition
that provides performance guarantees for the penalized underdamped Langevin 
Monte Carlo. 



\begin{prop}\label{cor:HMC:convex:outside}
Suppose Assumptions~\ref{assump:S:0}, \ref{assump:C}, and \ref{assump:f:2} hold,
and also assume
the conditions in 
Corollary~\ref{cor:Frobenius-lipschitz}
are satisfied.
Given the constraint set $\mathcal{C}$, consider its representation as $\mathcal{C} = \{ x: h(x)\leq 0\}$ given in \eqref{def-constraint-set-with-ineq} where $h(x)=\max_{1\leq i \leq m} h_i(x)$ for some $m\geq 1$ with $h_i$ convex for $i=1,2,\dots,m$.
Let $\nu_{K}$ be the distribution
of the $K$-th iterate $x_{K}$ of penalized underdamped Langevin 
Monte Carlo \eqref{HMC:nonconvex:1}-\eqref{HMC:nonconvex:2} for the constrained set $\mathcal{C}^{\alpha}$ defined in \eqref{C:alpha}
and the distribution of $(v_{0},x_{0})$ follows $\mathcal{N}(0,\frac{1}{L_{\delta}}I_{d})\otimes\mathcal{N}(0,\frac{1}{L_{\delta}}I_{d})$, 
where we take $\alpha=0$ if $h$ is strongly convex
and we take $\alpha=\varepsilon^{2}$ if $h$ is merely convex.
Then, we have $\text{TV}(\nu_{K},\pi)\leq\tilde{\mathcal{O}}(\varepsilon)$
provided that $\delta=\varepsilon^{4}$, $\alpha=\varepsilon^{2}$ and  
\begin{equation}
K=\tilde{\mathcal{O}}\left(\sqrt{d}/\varepsilon^{7}\right),
\end{equation}
where $\tilde{\mathcal{O}}$ ignores the dependence on $\log d$ and $\log(1/\varepsilon)$.
\end{prop}


\begin{rema}
When we compare the algorithmic complexity
in Proposition~\ref{cor:HMC:convex:outside}
with Proposition~\ref{cor:LD:convex:outside}, 
we see that for the underdamped-Langevin-based penalized underdamped Langevin Monte Carlo
has complexity $K=\tilde{\mathcal{O}}(\sqrt{d}/\varepsilon^{7})$,
which improves the dependence on both the dimension $d$ and the accuracy level $\varepsilon$
compared to the overdamped-Langevin-based penalized Langevin dynamics
where the complexity is $K=\tilde{\mathcal{O}}(d/\varepsilon^{10})$. This is obtained under additional assumptions on the smoothness of the boundary of $\mathcal{C}$ and Hessian Lipschitzness of $f$. To the best of our knowledge, 
$\tilde{\bigO}(\sqrt{d})$ is the best dependency on dimension for the constrained sampling.
\end{rema}

\begin{rema}
In Proposition~\ref{cor:HMC:convex:outside}, when $h$ is $\beta$-strongly convex (with $\beta>0$ and $\alpha=0$)
the leading-order complexity $K=\tilde{\mathcal{O}}\left(\sqrt{d}/\varepsilon^{7}\right)$
does not depend on $\beta$. It can be seen from the proof of Proposition~\ref{cor:HMC:convex:outside}
that the complexity $K$ has a second-order dependence on $\beta$, such that
$K=\tilde{\mathcal{O}}\left(\frac{\sqrt{d}}{\varepsilon^{7}}\right)+\tilde{\mathcal{O}}\left(\frac{\sqrt{d}}{\beta\varepsilon^{3}}\right)$,
where we ignored the dependence on the other constants when
we consider the second-order dependence on $\beta$. When $h$ is merely convex (with $\beta=0$ and $\alpha=\varepsilon^{2}$), we have $K=\tilde{\mathcal{O}}\left(\frac{\sqrt{d}}{\varepsilon^{7}}\right)+\tilde{\mathcal{O}}\left(\frac{\sqrt{d}}{\varepsilon^{5}}\right)$.
\end{rema}

\subsection{Penalized Langevin Algorithms with Stochastic Gradient}\label{sec:stochastic}

In the previous sections, we studied penalized Langevin algorithms with deterministic gradient
when the objective $f$ is non-convex. In this section, we study the extension
to allow stochastic estimates of the gradients in our algorithms. Supporting stochastic gradients becomes especially key in machine learning and data science applications where the exact gradients can be computationally expensive but stochastic estimates can be obtained efficiently from data. We start with the case when $f$ is assumed to be strongly convex and smooth.

\subsubsection{Strongly Convex Case}

In this section, 
we assume that the target $f$ is strongly convex and its gradient is Lipschitz. 
More precisely, we make the following assumption.

\begin{assum}\label{assump:f:1}
Assume that $f$ is $\mu$-strongly convex and $L$-smooth. 
\end{assum}

Assumption~\ref{assump:f:1} is equivalent to assuming that the target density $\pi(x) \propto e^{-f(x)}$ is strongly log-concave and smooth. This assumption has also been made frequently in the literature (see, e.g., \citet{bubeck2018sampling,bubeck2015finite,pmlr-v134-lamperski21a}).
Such densities arise in several applications including but not limited to Bayesian linear regression and Bayesian logistic regression (see, e.g., \citet{castillo2015bayesian,o2004bayesian}).  
Under Assumption~\ref{assump:f:1}, we have the following property for the target function $f$.

\begin{lemm}\label{lem-str-cvx-implies-lower-bound}
Under Assumptions~\ref{assump:f:1}
and Assumption~\ref{assump:C},
the minimizers of $f+\frac{S}{\delta}$ are
uniformly bounded in $\delta$ such that there exists some $c\geq 0$
and the norm of any minimizer of $f+\frac{S}{\delta}$
is bounded by $(1+c)R$. 
\end{lemm}

When $\delta$ is large, 
the minimizers of $f+\frac{S}{\delta}$ are
close to the minimizers of $f$, which are
uniformly bounded, and when $\delta$ is small, 
by the definition of the penalty function $S$,
the minimizers of $f+\frac{S}{\delta}$
will concentrate on the set $\mathcal{C}$.
Moreover, if the minimizers of $f$ are
inside the constrained set $\mathcal{C}$, 
then, the minimizers of $f+\frac{S}{\delta}$
must also lie in the set because $S(x)=(\delta_{\mathcal{C}}(x))^{2} = 0$ for $x\in \mathcal{C}$.
Hence, the above lemma naturally holds.




Moreover, under Assumptions~\ref{assump:f:1} and \ref{assump:C}, 
the conditions in Theorem~\ref{thm:final} are satisfied (see Lemma~\ref{condition:thm:final:2} in the Appendix for details). Building on this result, in the following subsections, we study penalized Langevin algorithms and the number of iterations needed to sample from a distribution within $\varepsilon$ distance to the target.

\textbf{Penalized Stochastic Gradient Langevin Dynamics.}
We now consider the extension to allow stochastic gradients, known as the stochastic gradient Langevin dynamics in the literature (see, e.g., \citet{welling2011bayesian, chen2015convergence,Raginsky}). 
In particular, we propose the penalized stochastic gradient Langevin dynamics (PSGLD):
\begin{equation}\label{SGLD:convex}
x_{k+1}=x_{k}-\eta\left(\nabla\tilde{f}(x_{k})+\frac{1}{\delta}\nabla S(x_{k})\right)+\sqrt{2\eta}\xi_{k+1},
\end{equation}
where $\xi_{k}$ are i.i.d. $\mathcal{N}(0,I_{d})$ Gaussian noises in $\mathbb{R}^{d}$ and we assume that we have access to noisy estimates $\tilde \nabla f(x_k)$ of the
actual gradients satisfying the following assumption:

\begin{assum}\label{assump:grad:noise} 
We assume at iteration $k$, we have access to $\tilde \nabla f(x_k, w_k)$ which is a random estimate of $\nabla f(x_k)$ where $w_k$ is a random variable independent from $\{w_j\}_{j=0}^{k-1}$ and satisfies $\mathbb{E}\left[\nabla\tilde{f}(x_k, w_k)-\nabla f(x_k)| x_{k}\right]=\nabla f(x_k)$ and
\begin{equation}\label{noise:gradient:assump}
\mathbb{E}\left\Vert\nabla\tilde{f}(x_k, w_k)-\nabla f(x_k)\Big| x_k\right\Vert^{2}
\leq
2\sigma^{2}\left(L^{2}\Vert x_k\Vert^{2}+\Vert\nabla f(0)\Vert^{2}\right).
\end{equation}
To simplify the notation, we suppress the $w_k$ dependence and denote $\nabla\tilde{f}(x_k, w_k)$ by $\tilde \nabla f(x_k)$.
\end{assum}

We note that the assumption \eqref{noise:gradient:assump} has been commonly made in data science and machine learning applications (see, e.g., \citet{Raginsky}) and arises when gradients are estimated from randomly sampled subsets of data points in the context of stochastic gradient methods. It is more general
than the assumption
$\mathbb{E}\left\Vert\nabla\tilde{f}(x_k)-\nabla f(x_k)\Big |x\right\Vert^{2}\leq\sigma^{2}d$
that has also been used in the literature \citep{DK2017} and allows handling gradient noise arising in many machine learning applications where the variance is not uniformly bounded \citep{Raginsky,aybat2019universally,gurbuzbalaban2021decentralized}.
In \eqref{noise:gradient:assump}, if $f(x)$ takes the form $f(x)=\sum_{i=1}^{n}f_{i}(x)$, 
and $\nabla\tilde{f}(x)=\frac{1}{b}\sum_{j\in\Omega}\nabla f_{j}(x)$, where $\Omega$ is a random
subset of $\{1,2,\ldots,n\}$ with batch-size $b$, due to the central limit theorem, we can assume that $\sigma^{2}=\mathcal{O}(1/b)$, where $b$ is the batch-size of the mini-batch.
We have the following proposition, which characterizes the number of iterations necessary to sample from the target up to an $\varepsilon$ error using the penalized stochastic gradient Langevin dynamics.

\begin{prop}\label{prop:SGLD:convex}
Suppose Assumptions~\ref{assump:C}, \ref{assump:f:1} and \ref{assump:grad:noise} hold.
Let $\nu_{K}$ denote the distribution 
of the $K$-th iterate $x_{K}$ of penalized stochastic gradient Langevin dynamics \eqref{SGLD:convex}.
We have $\mathcal{W}_{2}(\nu_{K},\pi)\leq\tilde{\mathcal{O}}(\varepsilon)$, where $\tilde{\mathcal{O}}$ ignores the dependence on $\log(1/\varepsilon)$,
provided that $\delta=\varepsilon^{8}$, the batch-size $b$ is of the constant order, and the stochastic gradient computations $\hat{K}:=Kb$
and the stepsize $\eta$ satisfy:
\begin{equation}
\hat{K}=\tilde{\bigO}\left(\frac{d(L\varepsilon^{8}+4)^{2}}{\varepsilon^{18}\mu^{3}}\right),
\qquad
\eta=\frac{\varepsilon^{18}\mu^{2}}{d(L\varepsilon^{8}+4)^{2}}.
\end{equation}
\end{prop}

In terms of the dependence on the condition number $\kappa:=L/\mu$, Proposition~\ref{prop:SGLD:convex} implies that the batch-size $b$ is of constant order, the stochastic gradient computations $\hat{K}=\tilde{\mathcal{O}}(\kappa^{2}/\mu)$ and the stepsize $\eta=\Theta(1/\kappa^{2})$.


\textbf{Penalized Stochastic Gradient Underdamped Langevin Monte Carlo.}
Next, we consider the extension to allow stochastic gradient, which we refer to 
as the stochastic gradient 
underdamped Langevin Monte Carlo (SGULMC). Such algorithms have been studied previously in the unconstrained setting in the literature \citep{chen2014stochastic,chen2015convergence,GGZ}. We propose the penalized stochastic gradient underdamped Langevin 
Monte Carlo (PSGULMC): 
\begin{align}
&v_{k+1}=\psi_{0}(\eta)v_{k}-\psi_{1}(\eta)\left(\nabla\tilde{f}(x_{k})+\frac{1}{\delta}\nabla S(x_{k})\right)+\sqrt{2\gamma}\xi_{k+1},\label{SGHMC:convex:1}
\\
&x_{k+1}=x_{k}+\psi_{1}(\eta)v_{k}-\psi_{2}(\eta)\left(\nabla\tilde{f}(x_{k})+\frac{1}{\delta}\nabla S(x_{k})\right)+\sqrt{2\gamma}\xi'_{k+1},\label{SGHMC:convex:2}
\end{align}
where $(\xi_{k},\xi'_{k})$ are i.i.d. $2d$-dimensional Gaussian noises 
independent of the initial condition $v_{0},x_{0}$,
centered with covariance matrix given in \eqref{defn:cov} and
$\psi_{k}(t)$ are defined in \eqref{defn:psi}, 
where we recall that the gradient noise satisfies Assumption~\ref{assump:grad:noise}.
Then we can provide the following proposition for the number of iterations we need to sample from the target distribution within $\varepsilon$ error using PSGULMC with a stochastic gradient that satisfies Assumption~\ref{assump:grad:noise}.

\begin{prop}\label{prop:SGHMC:convex}
Suppose Assumptions~\ref{assump:C}, \ref{assump:f:1} and \ref{assump:grad:noise} hold.
Let $\nu_{K}$ denote the distribution 
of the $K$-th iterate $x_{K}$ of penalized stochastic gradient underdamped Langevin Monte Carlo \eqref{SGHMC:convex:1}-\eqref{SGHMC:convex:2}
and $(v_{0},x_{0})$ follows the product distribution $\mathcal{N}(0,I_{d})\otimes\nu_{0}$.
We have $\mathcal{W}_{2}(\nu_{K},\pi)\leq\tilde{\mathcal{O}}(\varepsilon)$
provided that $\delta=\varepsilon^{8}$, and the batch-size $b$ satisfies: 
\begin{equation}
b=\Omega\left(\sigma^{-2}\right)
=\Omega\left(\frac{L^{2}(L\varepsilon^{8}+4)(\varepsilon^{16}d\mu+(L\varepsilon^{8}+4)^{2})}{\varepsilon^{26}\mu^{4}}\right),
\end{equation}
and the stochastic gradient computations $\hat{K}:=Kb$ and the stepsize $\eta$ satisfy:
\begin{align}
\hat{K}=\tilde{\bigO}
\Bigg(\frac{L^{2}(L\varepsilon^{8}+4)^{2}(\varepsilon^{16}d\mu+(L\varepsilon^{8}+4)^{2})\sqrt{(\mu+L)\varepsilon^{8}+4}}{\varepsilon^{39}\mu^{6}}
\max\left(\sqrt{d},\frac{\sqrt{(L+\mu)\varepsilon^{8}+4}}{\varepsilon^{3}}\right)\Bigg),\nonumber
\end{align}
and
\begin{equation}
\eta=\min\left(\frac{1}{\sqrt{d}}\frac{\varepsilon^{9}\mu}{(L\varepsilon^{8}+4)},\frac{1}{\sqrt{(\mu+L)\varepsilon^{8}+4}}\frac{\varepsilon^{12}\mu}{(L\varepsilon^{8}+4)}\right).\nonumber
\end{equation}
\end{prop}

In terms of the dependence on the condition number $\kappa=L/\mu$, Proposition~\ref{prop:SGHMC:convex} implies that the batch-size $b=\Omega(L^{5}/\mu^{4})=\Omega(L\kappa^{4})$, the stochastic gradient computations $\hat{K}=\tilde{\mathcal{O}}(\kappa^{6})$ and the stepsize $\eta=\Theta(1/(\sqrt{L}\kappa))$.


\subsubsection{Non-Convex Case}
This section discusses the case when $f$ is non-convex and smooth.

\textbf{Penalized Stochastic Gradient Langevin Dynamics.}
First, we consider 
the penalized stochastic gradient Langevin dynamics (PSGLD):
\begin{equation}\label{SGLD:nonconvex}
x_{k+1}=x_{k}-\eta\left(\nabla\tilde{f}(x_{k})+\frac{1}{\delta}\nabla S(x_{k})\right)+\sqrt{2\eta}\xi_{k+1},
\end{equation}
whereby following \citet{Raginsky} we assume
that the initial distribution $x_{0}$ satisfies the exponential integrability condition 
\begin{equation}\label{defn:kappa:0}
\kappa_{0}:=\log\mathbb{E}\left[e^{\Vert x_{0}\Vert^{2}}\right]<\infty,
\end{equation}
and we recall that the gradient noise satisfies Assumption~\ref{assump:grad:noise}. For instance, we could take $x_0$ to be a Dirac measure or any distribution that is compactly supported. Similar to Proposition~\ref{prop:SGLD:convex}, we have the following proposition about the complexity analysis of the PSGLD for the non-convex case.

\begin{prop}\label{prop:SGLD:nonconvex}
Suppose Assumptions~\ref{assump:S:0}, \ref{assump:C}, \ref{assump:grad:noise} and \ref{assump:f:2} hold.
Let $\nu_{K}$ be the distribution of the $K$-th iterate $x_{K}$ of penalized stochastic gradient Langevin dynamics \eqref{SGLD:nonconvex}.
We have $\mathcal{W}_{2}(\nu_{K},\pi)\leq\tilde{\mathcal{O}}(\varepsilon)$
provided that $\delta=\varepsilon^{8}$, the batch-size $b=\Omega(\eta^{-1})$ and the stochastic gradient computations
$\hat{K}:=Kb$ and the stepsize $\eta$ satisfy:
\begin{equation}
\hat{K}=\tilde{\mathcal{O}}\left(\frac{d^{17}\lambda_{\ast}^{-9}(\log(\lambda_{\ast}^{-1}))^{8}}{\varepsilon^{392}}\right),
\qquad
\eta=\tilde{\Theta}\left(\frac{\varepsilon^{196}}{d^{8}\lambda_{\ast}^{-4}(\log(\lambda_{\ast}^{-1}))^{4}}\right),
\end{equation}
where $\tilde{\mathcal{O}}$ and $\tilde{\Theta}$ ignore the dependence on $\log d$ and $\log(1/\varepsilon)$, and $\lambda_{*}$ is the spectral gap of the penalized overdamped Langevin SDE \eqref{penalized:overdamped:SDE}\footnote{This definition of the spectral gap can be found in \citet{Raginsky}.}:
\begin{equation}\label{defn:lambda:ast}
\lambda_{\ast}:=
\inf
\left\{
\frac{\int_{\mathbb{R}^{d}}\Vert\nabla g\Vert^{2}d\pi_{\delta}}{\int_{\mathbb{R}^{d}}g^{2}d\pi_{\delta}}:
g\in C^{1}(\mathbb{R}^{d})\cap L^{2}(\pi_{\delta}),
g\neq 0,
\int_{\mathbb{R}^{d}}gd\pi_{\delta}=0\right\}.
\end{equation}
Moreover, if we further assume that the assumptions of Corollary~\ref{cor:strongly:convex} hold,
then $\frac{1}{\lambda_{\ast}}\leq\mathcal{O}\left(1\right)$,    
and we have
$\hat{K}=\tilde{\mathcal{O}}\left(\frac{d^{17}}{\varepsilon^{392}}\right)$ and $\eta=\tilde{\Theta}\left(\frac{\varepsilon^{196}}{d^{8}}\right)$.
\end{prop}

\textbf{Penalized Stochastic Gradient Underdamped Langevin Monte Carlo.}
Next, we consider the extension of underdamped Langevin Monte Carlo to allow stochastic gradient, which we refer to as the stochastic gradient underdamped Langevin 
Monte Carlo (SGULMC). Such algorithms have been studied in the unconstrained setting in the literature \citep{chen2014stochastic,chen2015convergence,GGZ}.
We now consider the penalized stochastic gradient underdamped Langevin 
Monte Carlo (PSGULMC):
\begin{align}
&v_{k+1}=\psi_{0}(\eta)v_{k}-\psi_{1}(\eta)\left(\nabla\tilde{f}(x_{k})+\frac{1}{\delta}\nabla S(x_{k})\right)+\sqrt{2\gamma}\xi_{k+1},\label{SGHMC:nonconvex:1}
\\
&x_{k+1}=x_{k}+\psi_{1}(\eta)v_{k}-\psi_{2}(\eta)\left(\nabla\tilde{f}(x_{k})+\frac{1}{\delta}\nabla S(x_{k})\right)+\sqrt{2\gamma}\xi'_{k+1},\label{SGHMC:nonconvex:2}
\end{align}
where $(\xi_{k},\xi'_{k})$ are i.i.d. $2d$-dimensional Gaussian noises 
independent of the initial condition $v_{0},x_{0}$,
centered with covariance matrix given in \eqref{defn:cov} and
$\psi_{k}(t)$ are defined in \eqref{defn:psi},
and finally, we recall that the gradient noise satisfies Assumption~\ref{assump:grad:noise}.
We follow \citet{GGZ} by assuming that 
the probability law $\mu_{0}$ of the initial state $(x_{0},v_{0})$ satisfies
the following exponential integrability condition:
\begin{equation}\label{mu:0:integrability}
\int_{\mathbb{R}^{2d}}e^{\alpha\mathcal{V}(x,v)}\mu_{0}(dx,dv)<\infty\,,
\end{equation}
where $\mathcal{V}$ is a Lyapunov function:
\begin{equation}\label{eq:lyapunov}
\mathcal{V}(x,v):=f(x)+\frac{S(x)}{\delta}
+\frac{1}{4}\gamma^{2}\left(\left\Vert x+\gamma^{-1}v\right\Vert^{2}+\left\Vert\gamma^{-1}v\right\Vert^{2}-\lambda\Vert x\Vert^{2}\right)\,,
\end{equation}
and $\lambda$ is a positive constant less than $\min(1/4,m_{\delta}/(L_{\delta}+\gamma^{2}/2))$, and $\alpha= \lambda(1- 2\lambda)/12$,
where we recall from Lemma~\ref{lem:f:plus:S:dissipative} that $f+\frac{S}{\delta}$ is $(m_{\delta},b_{\delta})$-dissipative
where $m_{\delta},b_{\delta}$ are defined in \eqref{m:b:delta}.
Notice that there exists a constant $A\in(0,\infty)$ so that
\begin{equation}
\left\langle x,\nabla f(x)+\frac{\nabla S(x)}{\delta}\right\rangle\geq m_{\delta}\Vert x\Vert^{2}-b_{\delta}
\geq 2\lambda\left( f(x)+ \gamma^{2}\Vert x\Vert^{2}/4\right)-2A\,.
\label{eq:drift}
\end{equation}
Indeed, \citet{GGZ2} showed that one can take
\begin{align}
&\lambda:=\frac{1}{2}\min(1/4,m_{\delta}/(L_{\delta}+\gamma^{2}/2)),\label{def-lambda}
\\
&A:=\frac{m_{\delta}}{2L_{\delta}+\gamma^{2}}\left(\frac{\Vert\nabla f(0)\Vert^{2}}{2L_{\delta}+\gamma^{2}}+\frac{b_{\delta}}{m_{\delta}}\left(L_{\delta}+\frac{1}{2}\gamma^{2}\right)+f(0)\right),\label{def-A}
\end{align}
where we recall from Lemma~\ref{lem:f:plus:S:dissipative} that $f+\frac{S}{\delta}$ is $L_{\delta}$-smooth
with $L_{\delta}=L+\frac{\ell}{\delta}$.

The Lyapunov function \eqref{eq:lyapunov} is constructed in \citet{Eberle} as a key ingredient to show the convergence speed of the penalized underdamped Langevin SDE \eqref{eq:VL:penalized}-\eqref{eq:XL:penalized} to the Gibbs distribution $\pi_{\delta}(x,v)\propto\exp(-f(x)-\frac{1}{\delta}S(x)-\frac{1}{2}\Vert v\Vert^{2})$.
\citet{Eberle} shows that the convergence speed  of \eqref{eq:VL:penalized}-\eqref{eq:XL:penalized} to the Gibbs distribution $\pi_{\delta}$
is governed by 
\begin{align}
\mu_{\ast}:=\frac{\gamma}{768}\min\left\{\lambda L_{\delta}\gamma^{-2},\Lambda^{1/2}e^{-\Lambda}L_{\delta}
\gamma^{-2},\Lambda^{1/2}e^{-\Lambda}\right\}\label{def-mu-star-0},
\end{align}
where
\begin{align}
\Lambda:=\frac{12}{5}\left(1+2\alpha_{1}+2\alpha_{1}^{2}\right)(d+A)L_{\delta}\gamma^{-2}\lambda^{-1}(1-2\lambda)^{-1},
\qquad\alpha_{1}:=\left(1+\Lambda^{-1}\right)L_{\delta}\gamma^{-2}\label{def-capital-Lambda-alpha_1-0}.
\end{align}
The Lyapunov function \eqref{eq:lyapunov} also plays a key role 
in \citet{GGZ} that obtains the non-asymptotic convergence guarantees for (unconstrained) stochastic gradient underdamped Langevin Monte Carlo. We have the following proposition about the complexity of PSGULMC with stochastic gradient 
that satisfies Assumption~\ref{assump:grad:noise} for the non-convex case.

\begin{prop}\label{prop:SGHMC:nonconvex}
Suppose Assumptions~\ref{assump:S:0}, \ref{assump:C}, \ref{assump:grad:noise} and \ref{assump:f:2} hold.
Let $\nu_{K}$ be the distribution
of the $K$-th iterate $x_{K}$ of penalized stochastic gradient underdamped Langevin Monte Carlo \eqref{SGHMC:nonconvex:1}-\eqref{SGHMC:nonconvex:2}. 
We have $\mathcal{W}_{2}(\nu_{K},\pi)\leq\tilde{\mathcal{O}}(\varepsilon)$
provided that $\delta=\varepsilon^{8}$, the batch-size $b=\Omega(\eta^{-1})$ and the stochastic gradient computations $\hat{K}:=Kb$
and the stepsize $\eta$ satisfy:
\begin{equation}
\hat{K}=\tilde{\mathcal{O}}\left(\frac{d^{7}\left(\log(1/\mu_{\ast})\right)^{5}}{\varepsilon^{132}\mu_{\ast}^{3}}\right),
\qquad
\eta=\tilde{\Theta}\left(\frac{\varepsilon^{50}\mu_{\ast}}{d^{3}\left(\log(1/\mu_{\ast})\right)^{2}}\right),
\end{equation}
where $\tilde{\Theta}$ ignores the dependence on $\log d$ and $\log(1/\varepsilon)$.
\end{prop}


In Proposition~\ref{prop:SGHMC:nonconvex} (resp. Proposition~\ref{prop:SGLD:nonconvex}), $\mu_{\ast}$ (resp. $\lambda_{\ast}$) governs the speed
of convergence of the continuous-time penalized underdamped (resp. overdamped) Langevin SDEs
to the Gibbs distribution. 
It is shown in Proposition~1 in \citet{GGZ} that when
the surface of the target is relatively flat, 
$\mu_{\ast}$ can be better than $\lambda_{\ast}$ by a square root factor,
i.e. $1/\mu_{\ast}=\mathcal{O}\left(\sqrt{1/\lambda_{\ast}}\right)$.


\subsection{Avoiding Projections}\label{subsec-avoiding-projections}

We recall our discussion from Section \ref{subsubsec-penal_langevin} that the constraint set is often defined by functional inequalities of the form
$$\mathcal{C} := \{ x : h_i(x) \leq 0, \,\, \mbox{for} \,\, i=1,2,\dots,m\},$$
where $h_i(x):\mathbb{R}^n\to\mathbb{R}$ is convex and differentiable for every $i$. This would, for instance, be the case if $\mathcal{C}$ is the $\ell_p$-ball in $\mathbb{R}^d$ for $p\geq 1$. So far, our main complexity results involve the choice of $S(x)=\left(\delta_{\mathcal{C}}(x)\right)^{2}$ as a penalty function where computing $S(x)$ requires calculating projection of $x$ to the set $\mathcal{C}$.  Computing such projections can be carried out in polynomial time, but it can be costly in some cases, for instance, when the number of constraints $m$ is large or if the constraints are not simple. A natural question to ask is whether our results will hold if we use
$$ S(x) = \sum\nolimits_{i=1}^m \max\left(0, h_i(x)\right)^2,$$
as a penalty function and sample from the modified target
\begin{equation}
\pi_{\delta}(x)\propto\exp\left(-f(x)-\frac{1}{\delta}\sum\nolimits_{i=1}^{m}\max\left(0,h_{i}(x)\right)^{2}\right),\qquad x\in\mathbb{R}^{d},
\label{eq-target-general-constraints}
\end{equation}
instead. After all, this (alternative) choice of $S(x)$ would still satisfy our Assumption~\ref{assump:S:0}. In this section, we will show that this is indeed possible, provided that the functions $h_i(x)$ satisfy some growth conditions. The advantage of the formulation \eqref{eq-target-general-constraints} is that the modified target does not require computing the projection and the distance function $\delta_{\mathcal{C}}(x)$ to the constraint set as before, and it allows directly working with the functions that define the constraint set. This is computationally more efficient when computing the projections to the constraint set is not straightforward. For example, if $h_i(x)$'s are affine (in which case the constraint set $\mathcal{C}$ is a polyhedral set as an intersection of half-planes) and the number of constraints $m$ is large, computing the projection will be typically slower than evaluating the derivative of the penalized objective in \eqref{eq-target-general-constraints}.

 We first show that when $h_i(x)$'s are differentiable and convex for every $i$, then the function $S$ and therefore the density \eqref{eq-target-general-constraints} is differentiable despite the presence of the non-smooth $\max(0,\cdot)$ part in \eqref{eq-target-general-constraints}. Under some further assumptions, we also show in the next result that $S$ is $\ell$-smooth with appropriate constants.
\begin{lemm}\label{lemma:sum_gix_smooth}
If $h_i(x)$ is differentiable and convex on $\mathbb{R}^d$ for every $i=1,2,\dots,m$, then $\sum_{i=1}^{m}\max(0,h_{i}(x))^{2}$ is differentiable and
convex on $\mathbb{R}^d$. Furthermore, assume that on the set $\mathcal{B}_i:=\{x\in \R^d :\,\, h_i(x) \geq 0\}$, $h_i(x)$ satisfies the following three properties for every $i=1,2,\dots,m$: (i) $h_i(x)$ is continuously twice differentiable, (ii) the gradient of $h_i(x)$ is bounded $\lVert \nabla h_i(x) \rVert \leq N_i$, (iii) the Hessian of $h_i(x)$ satisfies $|h_i(x)|\nabla^2 h_i(x) \preceq P_iI$, i.e., the large eigenvalue of the matrix $|h_i(x)|\nabla^2 h_i(x)$ is smaller than or equal to a non-negative scalar $P_i$. Then, $\sum_{i=1}^{m}\max(0,h_{i}(x))^{2}$ is $\ell$-smooth, where $\ell:=2 \sum_{i=1}^m \left( N_i^2+P_i \right)$. 
\end{lemm}

The $\ell_p$ ball constraint arises in several applications that we will also discuss in the numerical experiments section (Section~\ref{sec:experiment}). Next, we show that the conditions in Lemma~\ref{lemma:sum_gix_smooth} can be satisfied for $\ell_p$ ball constraints.


\begin{coro}\label{prop:lp-norm}
If we choose $\mathcal{C}=\{x : h(x)\leq 0\}$ with $h(x)=\lVert x \rVert_p - R$ for a given $R>0$ with $p\geq 2$, then $\max(0,h(x))^2$ is $\ell$-smooth on $\mathbb{R}^d$, where $\ell:=\left(\frac{2}{R} + (d-1)\right) (p-1)$. 
\end{coro}

In the rest of this section, we will argue that our results can be extended to the penalty function $ S(x) = \sum\nolimits_{i=1}^m \max(0, h_i(x))^2,$ when $h_i$ satisfies certain growth conditions so that projections required by the distance-based penalty functions can be avoided. First of all, by applying the same arguments as in Lemma~\ref{lem:1}, 
we can show that for any $\delta>0$,
\begin{align*}
D(\pi\Vert\pi_{\delta})
\leq\frac{\int_{\mathbb{R}^{d}\backslash\mathcal{C}_{1}}e^{-\frac{1}{\delta}\sum_{i=1}^{m}\max(0,h_{i}(x))^{2}-f(x)}dx}{\int_{\mathcal{C}_{1}}e^{-f(x)}dx},
\end{align*}
where
\begin{equation}
\mathcal{C}_{1}:=\left\{x\in\mathbb{R}^{d}:\sum\nolimits_{i=1}^{m}\max(0,h_{i}(x))^{2} \leq 0\right\}
=\left\{x\in\mathbb{R}^{d}:\max\nolimits_{1\leq i\leq m}h_{i}(x)\leq 0\right\}.
\label{def-C1}
\end{equation}

Next, we provide an analog
of Lemma~\ref{lem:geometry}
that upper bounds of the Lebesgue measure 
of the set of all points that are outside $\mathcal{C}_{1}$
yet in a small neighborhood of $\mathcal{C}_{1}$. Consider the constraint map 
 $H:\mathbb{R}^{d}\rightarrow\mathbb{R}^m$ defined as
$H(x):=\left[h_{1}(x),h_{2}(x),\ldots,h_{m}(x)\right]^{\top}$.
We assume that $H$ is \emph{metrically subregular} everywhere
on the boundary of the constrained set, i.e. we assume there exists some constant $\bar{K}>0$ such that for any sufficiently small $\epsilon>0$, 
\begin{equation}
\left|x\in\mathbb{R}^{d}\backslash\mathcal{C}_{1}: \sum\nolimits_{i=1}^{m}\max(0,h_{i}(x))^{2}\leq\epsilon\right|
\leq
\left|x\in\mathbb{R}^{d}\backslash\mathcal{C}_{1}: \delta_{\mathcal{C}_{1}}(x)\leq\sqrt{\bar{K}\epsilon}\right|,
\end{equation}
(see, e.g., \citet{ioffe2015,ioffe2016metric} for more about metric subregularity and its consequences). For instance, the last inequality is satisfied when the constraint set is the $\ell_1$ ball with radius $R$ which is a a polyhedral set that can be expressed in the form \eqref{def-C-1} with affine choices of $h_i(x)$. Another example, would be the $\ell_p$ ball of radius $R$; i.e when $\mathcal{C}= \{ x: h(x)\leq 0 \}$ with $h(x) = \max_i h_i(x) = \|x\|_{p}-R$ and $p>1$. By applying the same arguments as in Lemma~\ref{lem:geometry}, we conclude that
there exists some constant $\bar{K}>0$ such that for any sufficiently small $\epsilon>0$, 
\begin{equation}
\left|x\in\mathbb{R}^{d}\backslash\mathcal{C}_{1}: \sum\nolimits_{i=1}^{m}\max(0,h_{i}(x))^{2}\leq\epsilon\right|
\leq
\left(\left(1+\sqrt{\bar{K}\epsilon}/r_{1}\right)^{d}-1\right)|\mathcal{C}_{1}|,
\end{equation}
where we assumed that $\mathcal{C}_{1}$ contains an open ball of radius $r_{1}$ centered at $0$.
Furthermore, Lemma~\ref{lem:D}, Lemma~\ref{lem:S} and Theorem~\ref{thm:final} still apply with minor modifications and it follows that as $\delta\rightarrow 0$,
$\mathcal{W}_{2}(\pi_{\delta},\pi)\leq\mathcal{O}\left(\left(\delta\log(1/\delta)\right)^{1/8}\right)$,
which is an analogue of Theorem~\ref{thm:final}. We can then 
obtain analogous results for PSGLD, 
and PSGULMC in Section~\ref{sec:stochastic}.
We can then utilize the conclusions of
previous sections to get the convergence rate and complexity by using penalized Langevin and underdamped Langevin Monte Carlo algorithms in this setting.

\begin{figure}[t]
\centering
    \subfigure[Penalized LD (PLD)]{
    \includegraphics[width=0.4\columnwidth]{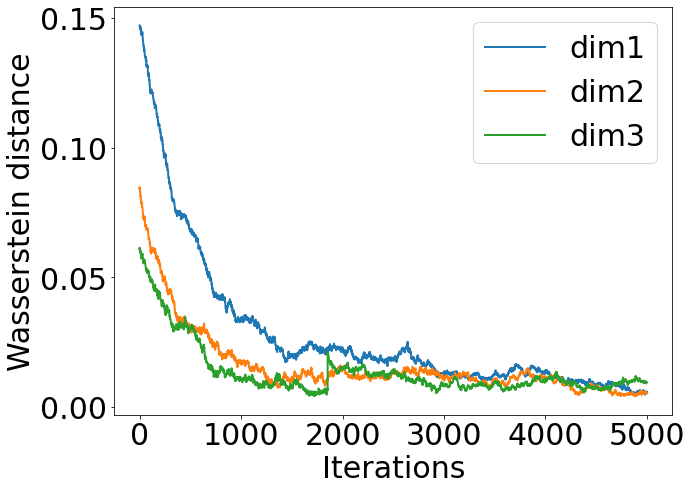}
    }
    \hfill
    \subfigure[Penalized Underdamped Langevin Monte Carlo (PULMC)]{
    \includegraphics[width=0.4\columnwidth]{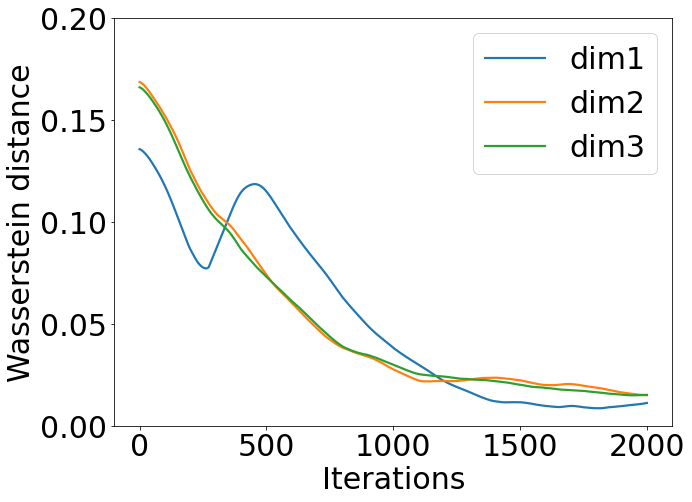} 
    }
    \hfill
\caption{Wasserstein distance between the target distribution and our proposed methods.}
\label{fig:Diri-wass}
\end{figure}

\begin{figure}[t]
\centering
    \subfigure[True distribution]{
    \includegraphics[width=0.3\columnwidth]{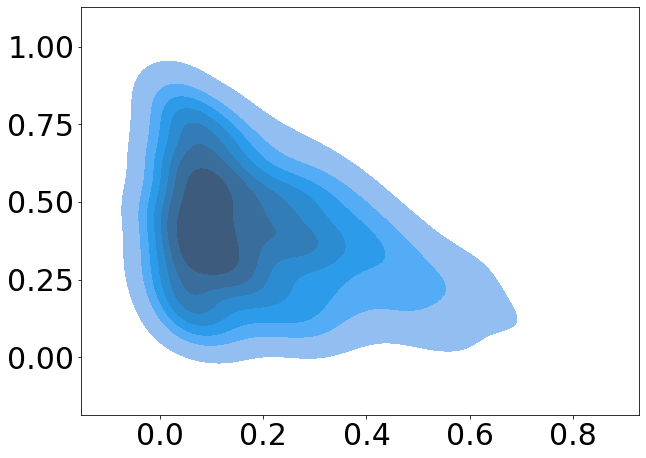}
    }
    \hfill
    \subfigure[Penalized LD (PLD)]{
    \includegraphics[width=0.3\columnwidth]{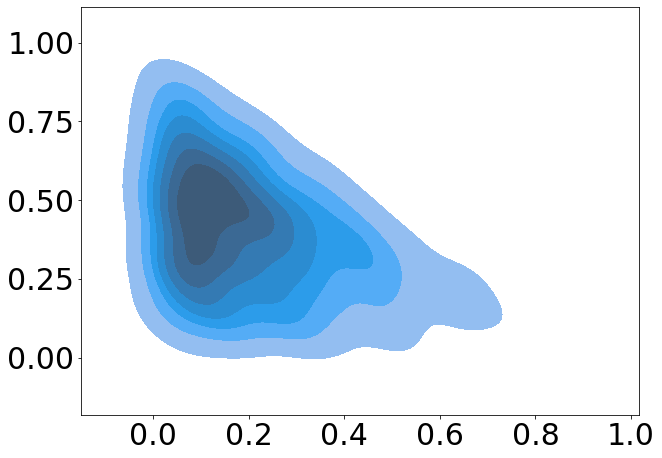}
    }
    \hfill
    \subfigure[Penalized ULMC (PULMC)]{
    \includegraphics[width=0.3\columnwidth]{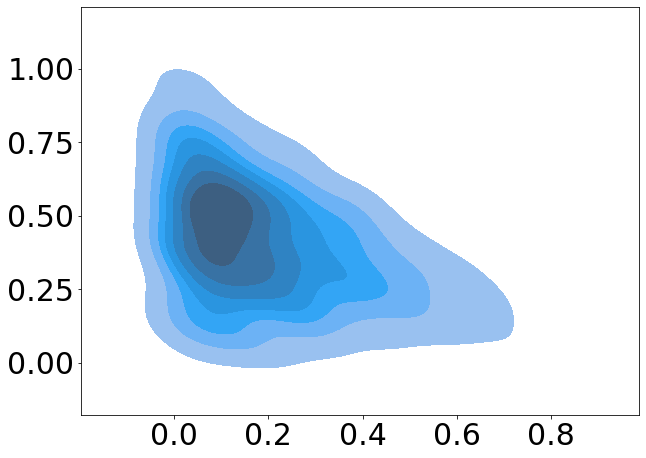} 
    }
\caption{Density plots of the target distribution and samples obtained by PLD and PULMC.}
\label{fig:Diri-density}
\end{figure}
\section{Numerical Experiments}\label{sec:experiment}

\subsection{Synthetic Experiment for Dirichlet Posterior}
As a toy experiment, we consider our proposed PLD and PULMC algorithms for sampling from a $3$-dimensional Dirichlet posterior distribution. The Dirichlet distribution is commonly used in machine learning, especially in Latent Dirichlet allocation (LDA) problems; see, e.g., \citet{Blei2003}. The Dirichlet distribution of dimension $K\geq 2$ with parameters $\alpha_1, \dots ,\alpha_K >0$ has a probability density function with respect to Lebesgue measure on $\mathbb{R}^{K-1}$ given by:
\begin{equation}\label{eqn:pdf}
    f(x_1, \dots , x_K; \alpha_1, \dots, \alpha_K) = \frac{1}{B(\alpha)}\prod_{i=1}^{K} x_i^{\alpha_i-1},
\end{equation}
where $\{x_k\}_{k=1}^{K}$ belongs to the standard $K-1$ simplex, i.e. $\sum_{i=1}^{K} x_i = 1 \text{ and } x_i\geq 0 \text{ for all } i\in \{1,\dots ,K\},$
and the normalizing constant $B(\alpha)$ in equation~\eqref{eqn:pdf} is the multivariate alpha function, which can be expressed as
$B(\alpha) = \frac{\prod_{i=1}^{K} \Gamma(\alpha_i)}{\Gamma(\sum_{i=1}^{K}\alpha_{i})}$, for any $\alpha := (\alpha_1, \dots, \alpha_K)\in\mathbb{R}_{\geq 0}^{K}$,
where $\Gamma(\cdot)$ denotes the gamma function.

\begin{figure}[t]
\centering
    \subfigure[PLD for Dirichlet sampling]{
    \includegraphics[width=0.32\columnwidth]{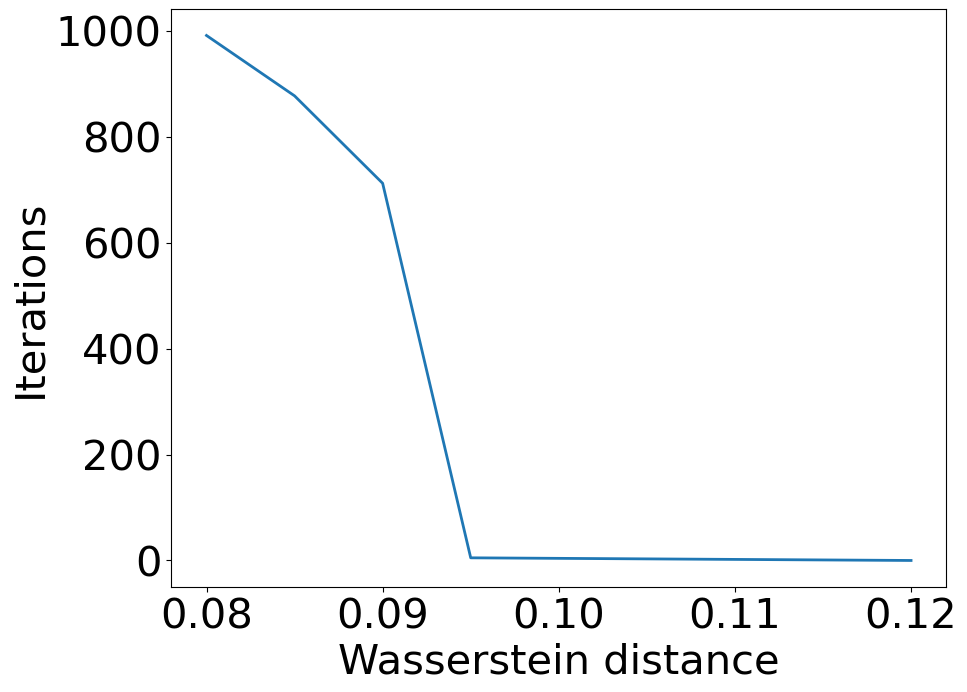}
    }
    \qquad \qquad
    \subfigure[PULMC for Dirichlet sampling]{
    \includegraphics[width=0.32\columnwidth]{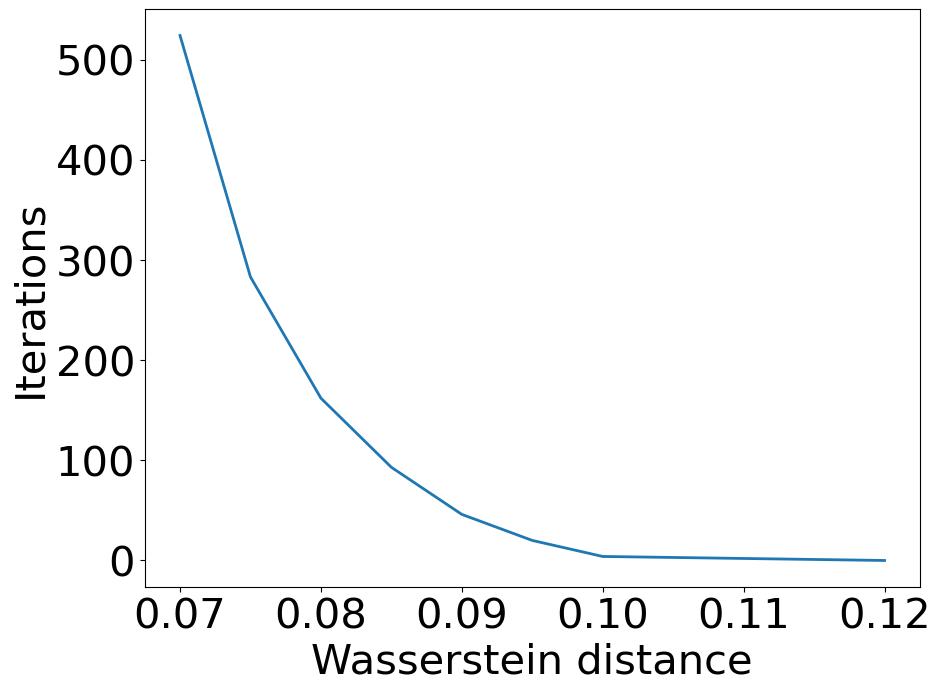}
    }
\caption{Average number of iterations required for achieving a target accuracy $\varepsilon$ (measured in terms of the Wasserstein distance) for the Dirichlet sampling problem as $\varepsilon$ is varied for PLD (left panel) and PULMC (right panel).}
\label{fig:var-dep1}
\end{figure}

In our experiment, we set $\alpha = (1,2,2)$ and use uniform distribution on the simplex as the prior distribution. 
For PLD, we set $\delta = 0.005$ and learning rate $\eta = 0.0001$, and $\eta$ is decreased by $25\%$ every $1000$ iterations. For PULMC, we set $\delta = 0.01$, $\gamma = 0.6$, and we take the learning rate $\eta = 0.0012$, where $\eta$ is decreased by $10\%$ every $200$ iterations. We obtain $1000$ samples from the posterior distribution using our methods and calculate the 2-Wasserstein distance for each of the three (coordinates) dimensions with respect to the true distribution based on $1000$ runs. The results in Figure~\ref{fig:Diri-wass} illustrates the convergence of our methods where we observe that the 2-Wasserstein distance decays to zero in each dimension for both PLD and PULMC methods. 
In Figure~\ref{fig:Diri-density}, on the left panel we illustrate the target distribution whereas in the middle and right panels, we illustrate the density of the samples obtained by PLD and PULMC methods, based on $1000$ samples. 
These figures illustrate that PLD and PULMC can sample successfully from the true Dirichlet distribution for this problem. In Figure \ref{fig:var-dep1}, we also plot the (expected) average number of iterations $k$ required for achieving an accuracy $\varepsilon$, i.e. for achieving $\mathcal{W}_2(\mathcal{L}(x_k), \pi) \leq \varepsilon$ where $x_k$ are the iterates and $\pi$ is the target Dirichlet distribution. In Figure \ref{fig:var-dep1}, the x-axis is the accuracy $\varepsilon$ whereas the y-axis is the number of iterations required. PULMC and PLD performs similarly, especially when the accuracy required is not small. It may be that both algorithms admit a better scaling in practice on this example with respect to $\varepsilon$ than the worst-case theoretical bounds we provide in Table \ref{tab:methods}. 
\begin{figure}[t!]
\centering
    \subfigure[PLD for Dirichlet sampling]{
    \includegraphics[width=0.32\columnwidth]{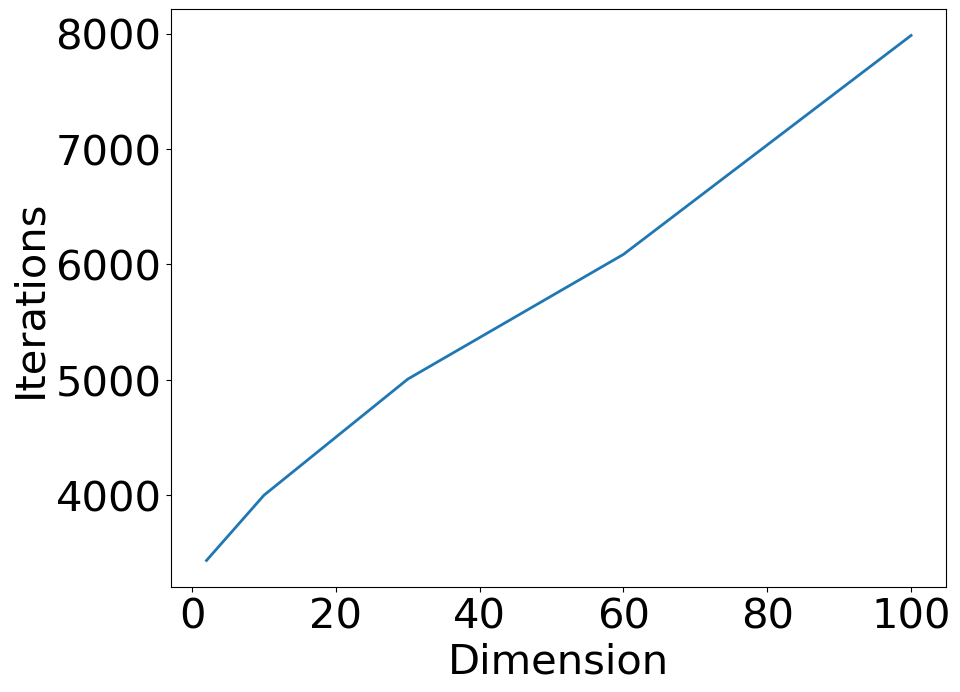}
    }
    \qquad\qquad
    \subfigure[PULMC for Dirichlet sampling]{
    \includegraphics[width=0.32\columnwidth]{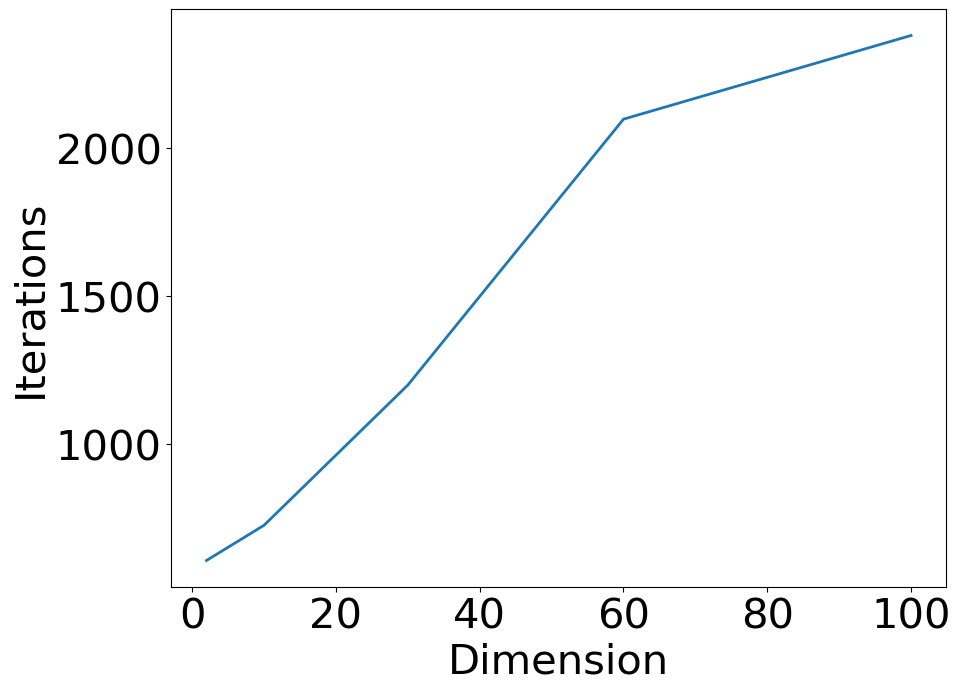}
    }
\caption{Dimension ($d$) dependency of PLD and PULMC on the Dirichlet distribution sampling problem.}
\label{fig:d-dep1}
\end{figure}In Figure~\ref{fig:d-dep1}, we also vary the dimension $d$ while keeping the target accuracy $\varepsilon$ fixed. More specifically, we report the (estimated) expected number of iterations needed to achieve the Wasserstein distance at most $\varepsilon=0.25$. 
The parameter $\alpha=(\alpha_1, \alpha_2,\dots,  \alpha_d)$ of the Dirichlet distribution in dimension $d$ is generated randomly, where $\alpha_i$ is set to a uniformly random integer from 1 to 5 independently for every $i=1,2,\dots,d$. We tuned the parameters for both algorithms. In the PLD case, we use $\delta=0.0004, \eta=0.0003/d$. 
In the PULMC case, we use $\delta=0.001, \eta=0.0012/d, \gamma=0.7$. We observe that the number of iterations required for PLD grows (approximately) linearly in the dimension $d$, whereas for PULMC we have roughly a sublinear growth in the dimension. The experimental results are more or less inline with our theoretical findings, where we prove that for the TV distance, PULMC admits better ($\mathcal{O}(\sqrt{d})$) guarantees compared to $\mathcal{O}(d)$ guarantees of PLD.

\subsection{Bayesian Constrained Linear Regression}

We consider Bayesian constrained linear regression models in our next set of experiments. Such models have many applications in data science and machine learning \citep{Brosse,bubeck2018sampling}. For example, if the constraint set is an $\ell_p$-ball around the origin, for $p=1$, we obtain the Bayesian Lasso regression, and for $p=2$, we get the Bayesian Ridge regression. We will consider both synthetic data and real-world data settings.
\begin{figure}[t]
\centering
    \subfigure[Prior]{
    \includegraphics[width=0.3\columnwidth]{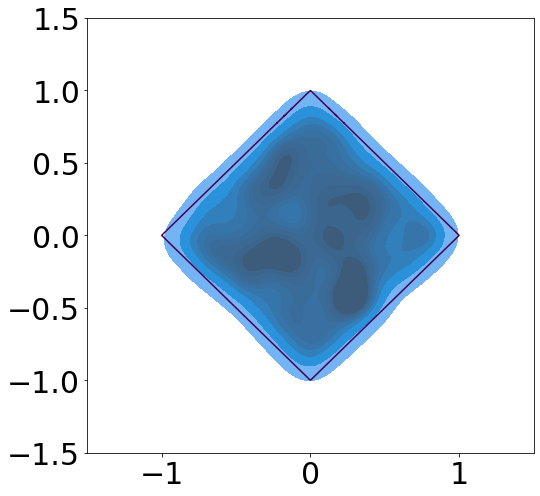}
    \label{fig:1-2d-prior}
    }
    \hfill
    \subfigure[Penalized SGLD]{
    \includegraphics[width=0.3\columnwidth]{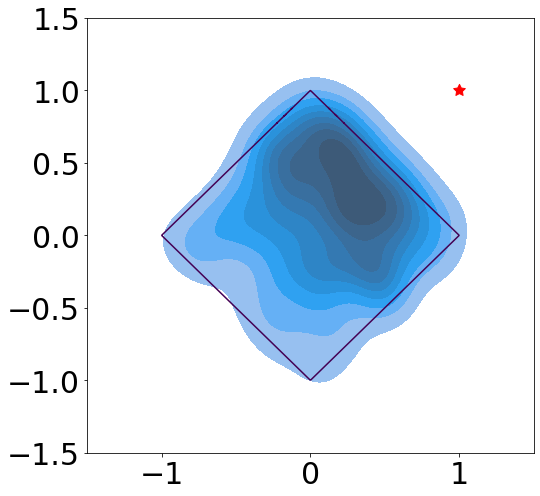} 
    }
    \hfill
    \subfigure[Penalized SGULMC]{
    \includegraphics[width=0.3\columnwidth]{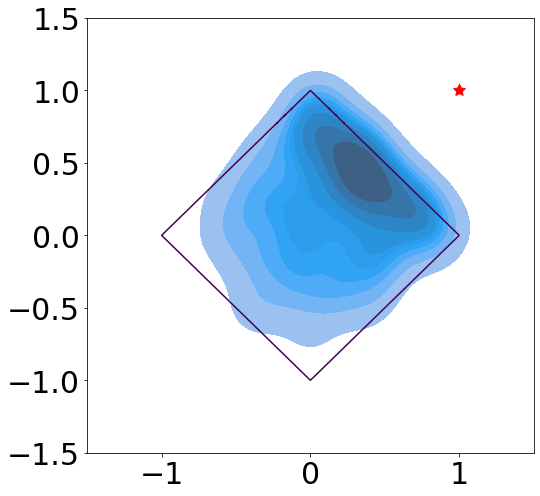} 
    }
\caption{Prior and posterior distribution with $1$-norm constraint in dimension 2.}
\label{fig:1-2d-linreg}
\end{figure}
\subsubsection{Synthetic 2-Dimensional Problem}

In our first set of experiments, we will consider the case when $p=1$, which corresponds to the Bayesian Lasso regression \citep{hans2009bayesian}. For better visualization, we start with a synthetic 2-dimensional problem. We generate 10,000 data points $(a_j, y_j)$ according to the model:
\begin{equation}
    \delta_j \sim \mathcal{N}\left(0,0.25\right),\quad  a_j \sim \mathcal{N}(0,I),\quad  y_j={x_{\star}}^{\top}a_j+\delta_j,\quad x_{\star}=[1,1]^{\top}.
\label{num:lin_gen}
\end{equation}
We take the constraint set to be 
\begin{equation*}
\mathcal{C}=\left\{x:\|x\|_1\leq 1\right\}.
\end{equation*}
The prior distribution is the uniform distribution, where the constraints are satisfied. This is illustrated in Figure~\ref{fig:1-2d-prior}. The posterior distribution of this model is given by 
\begin{equation*}
\pi(x)\propto e^{\sum_{j=1}^n-\frac{1}{2}(y_j-x^{\top}a_j)^2}\cdot \mathbbm{1}_{\mathcal{C}}, 
\end{equation*}
where $\mathbbm{1}_{\mathcal{C}}$ is the indicator function for the constraint set $\mathcal{C}$ and $n=10,000$ is the number of data points. For this set of experiments, we take the batch size $b=50$ and run PSGLD with
$\delta=0.001$, the learning rate $\eta= 10^{-5}$ where we reduce $\eta$ by $15\%$ every 5000 iterations. The total number of iterations is set to 50,000. For PSGULMC, we have a similar setting with $\delta=0.001$, $\gamma = 0.1$, and learning rate $\eta=0.0001$, where we reduce $\eta$ by $15\%$ every 5000 iterations. The results are shown in Figure~\ref{fig:1-2d-linreg} where the point $x_{\star}$ is marked with a red asterisk. In Figure~\ref{fig:1-2d-linreg}, we estimate the density of the samples obtained by both PSGLD and PSGULMC methods based on 500 runs. We see that the densities obtained by PSGLD and PSGULMC algorithms are compatible with the constraints, and they sample from a target distribution that puts higher weights into regions closer to $x_{\star}$ as expected (where without any constraints $x_{\star}$ would be the peak of the target).

\begin{figure}[t]
\centering
    \subfigure[PSGLD for ellipsoid constraints]{
    \includegraphics[width=0.4\columnwidth]{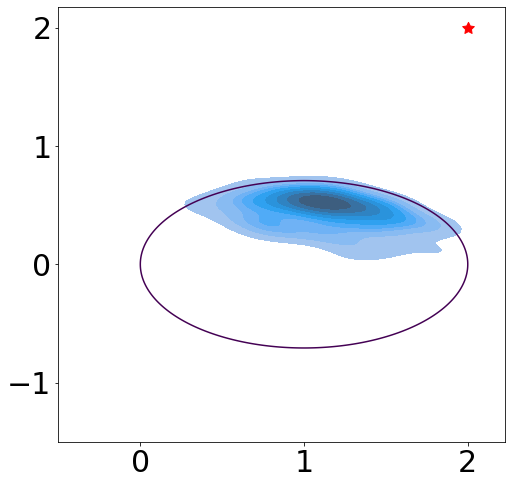}
    }
    \hfill
    \subfigure[PSGULMC for ellipsoid constraints]{
    \includegraphics[width=0.4\columnwidth]{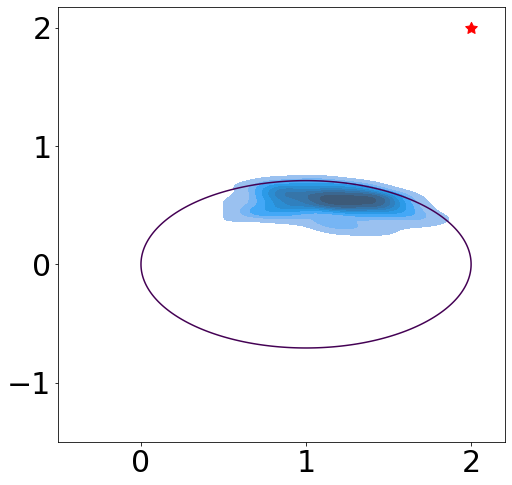}
    }
\caption{The density plot of the posterior distribution with ellipsoid constraints.}
\label{fig:quadratic-cons}
\end{figure}

We also consider an ellipsoidal constraint set 
\begin{equation*}
\mathcal{C} := \left\{ x~:~ (x-\bar{a}_1)^{\top} Q_1 (x-\bar{a}_1) \leq \bar{b}_1\right\},
\end{equation*}
for the same posterior distribution 
\begin{equation*}
\pi(x)\propto e^{\sum_{j=1}^n-\frac{1}{2}(y_j-x^{\top}a_j)^2}\cdot \mathbbm{1}_{\mathcal{C}}, 
\end{equation*}
where $Q_1 \in \mathbb{R}^{2\times 2}$ is positive definite, $\bar{a}_1 \in \mathbb{R}^2$ is a real vector and $\bar{b}_1 >0$ is a real scalar.
We take $x_{\star}=[2,2]^{\top}$ and 
$$
\bar{a}_1=[1,0]^{\top},\quad \bar{b}_1=1,\quad 
Q_1 = \begin{pmatrix}
  1 & 0\\ 
  0 & 2
\end{pmatrix}. 
$$
If we use the squared distances 
$S(x) = (\min_{y\in \mathcal{C}}\|x-y\|)^2$ as a penalty, then this will necessitate calculating projections to the ellipsoid constraint. 
However, we can avoid projections by following the methodology described in Section~\ref{subsec-avoiding-projections}. Namely, we take 
$$
S(x) = \max\left(0,(x-\bar{a}_1)^{\top} Q_1 (x-\bar{a}_1) - \bar{b}_1\right).
$$
The ellipsoid constraint set and a contour plot of the densities obtained by PSGLD and PSGULMC algorithms are reported in Figure~\ref{fig:quadratic-cons}, where the lighter colors in the contour plots (including the white and light blue colors) correspond to regions with a smaller estimated density compared to darker blue regions. In these experiments, we tuned the parameters for each algorithm: For PSGLD, we set $\delta=0.0001, \eta = 0.00005$, the number of iterations $k=20,000$ and we reduce the stepsize $\eta$  
by 15\% every $10,000$ iterations. For PSGULMC, we set $\delta=0.001, \gamma = 0.1, \eta = 0.00001$, the number of iterations $k=17,000$ where the stepsize $\eta$ is reduced by 5\% every $5,000$ iterations. 
We see that the densities lie within the constraints, and PSGLD and PSGULMC 
sample from a target distribution that puts higher weights into regions closer to $x_{\star}$ as expected.

\begin{figure}[t]
\centering
    \subfigure[PLD for sampling a Gaussian mixture subject to ellipsoid constraints]{
    \includegraphics[width=0.25\columnwidth]{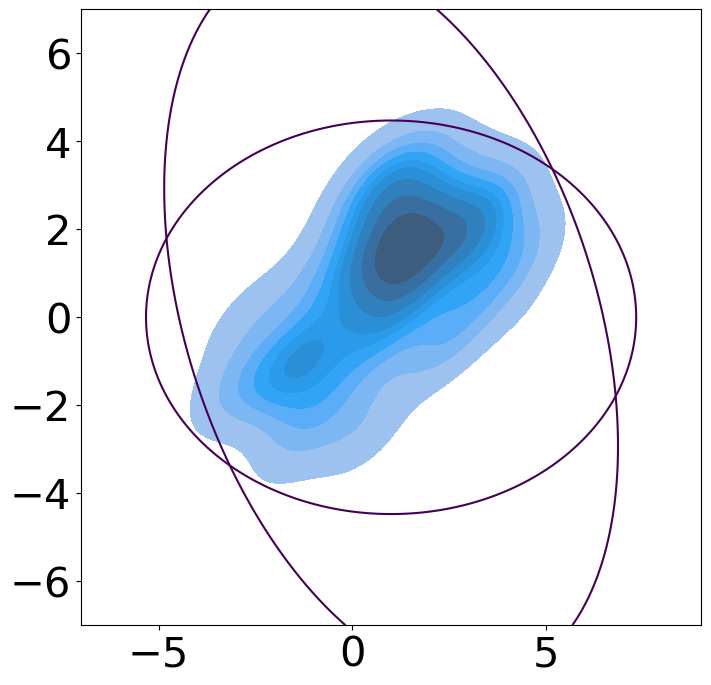}
    }
    \qquad \qquad
    \subfigure[PULMC for sampling a Gaussian mixture subject to ellipsoid constraints]{
    \includegraphics[width=0.25\columnwidth]{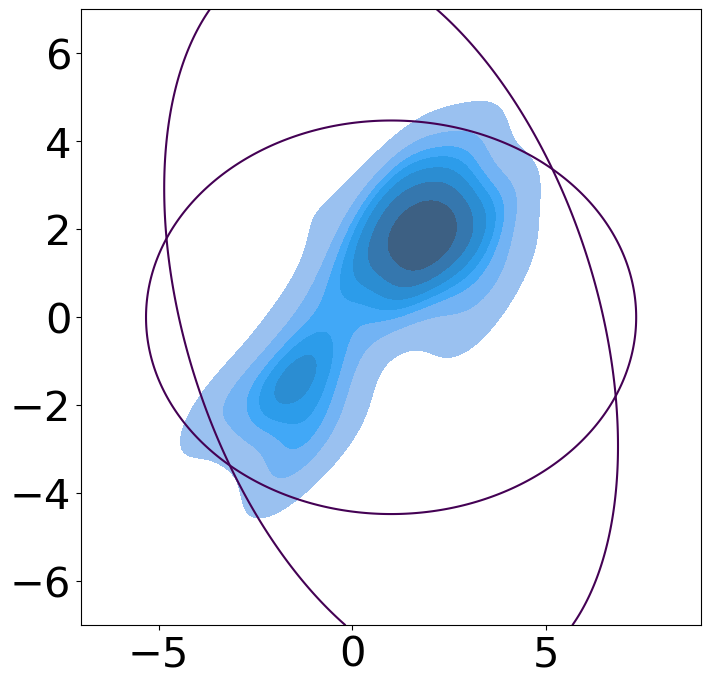}
    }
\caption{The contour plot of the density for a Gaussian mixture with ellipsoid constraints.}
\label{fig:gau-mix}
\end{figure}

We also considered another example where the aim is to sample from a Gaussian mixture 
\begin{equation*}
\pi(x) \propto \frac{2}{3}\exp(-\|x-z_1\|^2/2) + \frac{1}{3}\exp(-\|x-z_2\|^2/2), 
\end{equation*}
where $z_1 = [2,2]^{\top}, z_2 = [-2,-2]^{\top}$. We consider the non-convex constraint set $\mathcal{C}= \mathcal{C}_1 \cap \mathcal{C}_2$ obtained by intersecting the ellipsoids
$$
\mathcal{C}_i := \left\{x~:~ (x-\bar{a}_i)^{\top} Q_i (x-\bar{a}_i) - \bar{b}_1 \leq 0 \right\},\qquad\text{for $i=1,2$.} 
$$
We take  
$$\bar{a}_1=[1,0]^{\top},\quad \bar{a}_2=[1,0]^{\top},\quad \bar{b}_2 = 60,\quad \bar{b}_1=40, \quad Q_1 = \begin{pmatrix}
  1 & 0\\ 
  0 & 2
\end{pmatrix}, \quad Q_2 = \begin{pmatrix}
  2 & 1\\ 
  0 & 1
\end{pmatrix},$$ 
and consider the penalty 
$$
S(x) = \max\left(0,(x-\bar{a}_1)^{\top} Q_1 (x-\bar{a}_1) - \bar{b}_1\right) + \max\left(0,(x-\bar{a}_2)^{\top} Q_2 (x-\bar{a}_2) - \bar{b}_2\right).
$$
 For both PLD and PULMC, we take 10,000 iterations. The results are given in Figure~\ref{fig:gau-mix} where we densities of the distributions that are outputs of PLD and PULMC algorithms are given as a contour plot and where the constraint set $\mathcal{C}$ is the intersection of the two ellipsoids displayed in the figure. We can see that the output distributions obtained PLD and PULMC are within both of the constraints, and the peaks of the two Gaussian distributions that are part of the mixture can be clearly observed in the figures.

\subsubsection{Diabetes Dataset Experiment}

\begin{figure}[t]
\centering
    \subfigure[Penalized SGLD]{
    \includegraphics[width=0.3\columnwidth]{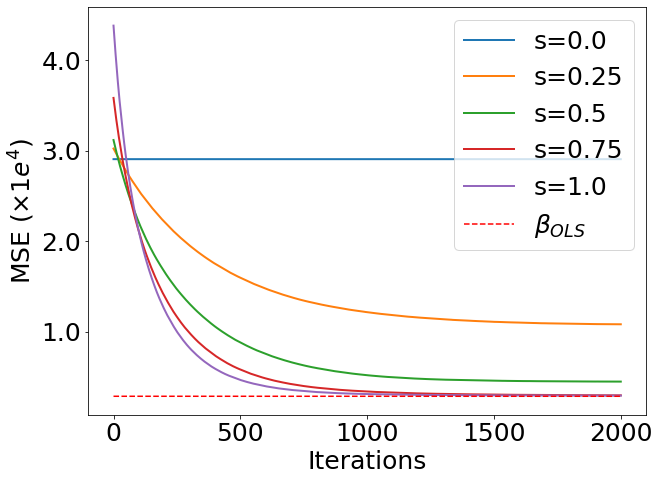}\label{fig:linreg_diabete_sgld}
    }
    \hfill
    \subfigure[Penalized SGULMC]{
    \includegraphics[width=0.3\columnwidth]{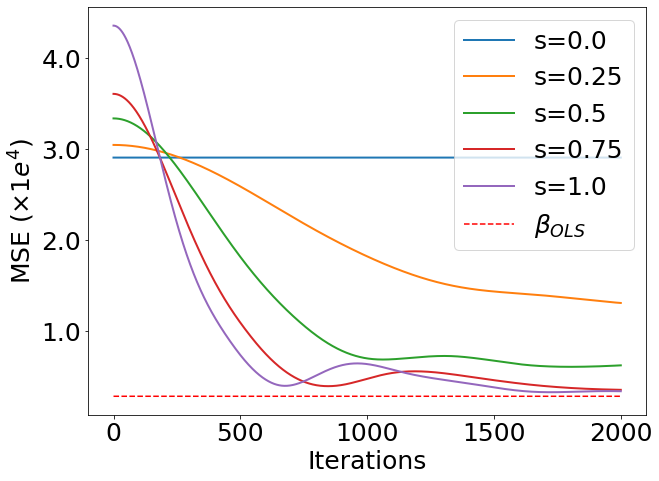} \label{fig:linreg_diabete_sghmc}
    }
    \hfill
    \subfigure[Norm of parameters]{
    \includegraphics[width=0.3\columnwidth]{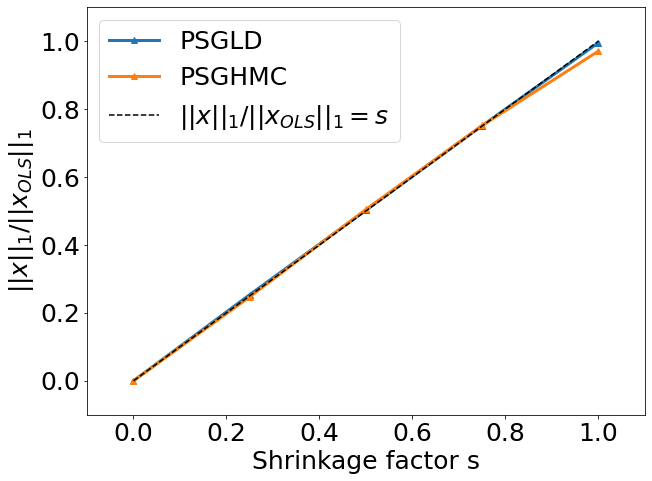}
    \label{fig:linreg_diabete_norm_SGLD}
    }
\caption{Penalized SGLD and Penalized SGULMC results for Diabetes dataset with 1-norm ball constraints in dimension 2.}
\label{fig:linreg_diabete_SGLD}
\end{figure}

\begin{figure}[t!]
\centering
    \subfigure[PSGLD for the Diabetes dataset]{
    \includegraphics[width=0.4\columnwidth]{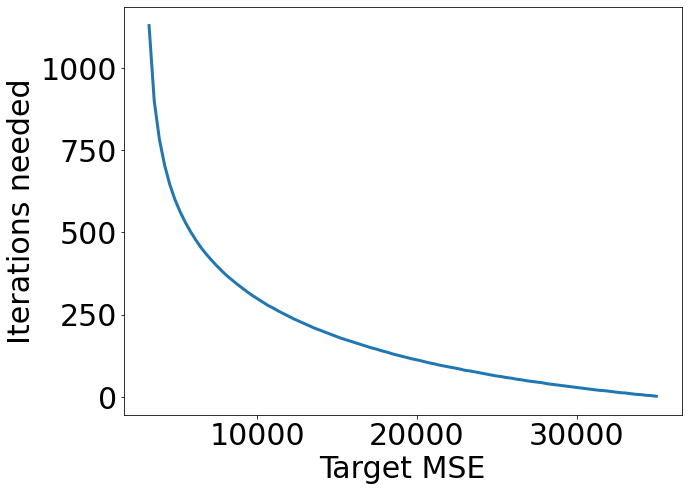}
    }
    \hfill
    \subfigure[PSGULMC for the Diabetes dataset]{
    \includegraphics[width=0.4\columnwidth]{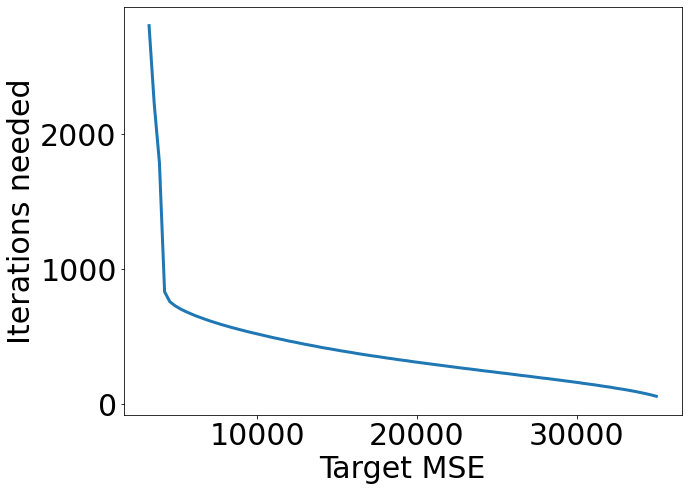}
    }
\caption{Target MSE vs. average number of iterations needed to reach the target for the Diabetes dataset.}
\label{fig:var-dep}
\end{figure}

Besides the synthetic dataset, we consider the Bayesian constrained linear regression on the Diabetes dataset.\protect\footnote{This dataset is available online at \url{https://archive.ics.uci.edu/ml/datasets/diabetes}.} 
Similar to \citet{Brosse}, we take the constraint set to be 
\begin{equation*}
\mathcal{C}=\left\{x:\|x\|_1 \leq s\|x_{\text{OLS}}\|_1\right\}, 
\end{equation*}
where $s$ is the shrinkage factor and $x_{\text{OLS}}$ is the solution to the ordinary least squares problem without any constraints. 
The posterior distribution of this model is given by 
\begin{equation*}
\pi(x)\propto e^{\sum_{j=1}^n-\frac{1}{2}(y_j-x^{\top} a_j)^2}\cdot \mathbbm{1}_{\mathcal{C}}, 
\end{equation*}
where $\mathbbm{1}_{\mathcal{C}}$ is the indicator function for the constraint set $\mathcal{C}$, and $(a_j,y_j),j=1,2,\dots n$ are the data points in the Diabetes dataset.
We experiment with different choices of $s$ ranging from 0 to 1. 
For penalized SGLD, we set $\eta=s\|x_{\text{OLS}}\|\times 10^{-5},b = 50$, and $\delta=0.05$, for penalized SGULMC, we set $\eta=s\|x_{\text{OLS}}\|\times 10^{-5},b = 50,\delta=0.05$, and $\gamma=0.6$. We take the prior distribution to be the uniform distribution on $\mathcal{C}$. We run our algorithms 100 times, and for the $\ell$-th run, we let $x_k^{(\ell)}$ denote the $k$-th iterate of the $\ell$-th run of our algorithms. First, we compute the mean squared error 
$\mathrm{MSE}_k^{(\ell)}:=\frac{1}{n}\sum_{j=1}^n (y_j-(x_k^{(\ell)})^{\top}a_j)^2$
corresponding to the $k$-th iterate of the $\ell$-th run. In Figure~\ref{fig:linreg_diabete_sgld} and ~\ref{fig:linreg_diabete_sghmc}, we report the average of the mean squared error values of each iteration, averaged over 100 runs, i.e. we plot $\text{MSE}_k:=\frac{1}{100}\sum_{\ell=1}^{100}  \text{MSE}_k^{(\ell)}$ over the iterations $k$.
 The results of averaged MSE over 100 samples are shown in Figure~\ref{fig:linreg_diabete_sgld} and ~\ref{fig:linreg_diabete_sghmc}. We can observe from these figures that with $s$ increasing from $0$ to $1$, the average mean squared error will decrease to the mean squared error of $x_{\text{OLS}}$ for $p=1$ as expected. 
As the number of iterations increases, the error of iterates decreases to a steady state. To illustrate that the final iterates of both algorithms are still lying in the constraint set $\mathcal{C}$, we plot the maximum values of $\|x_{\text{last}}\|_1/\|x_{\text{OLS}}\|_1$ calculated among 100 samples in Figure~\ref{fig:linreg_diabete_norm_SGLD}, where $x_{\text{last}}$ is the last iterates of each sample from both algorithms, against the shrinkage factor $s$. The results from PSGLD and PSGULMC are shown as the blue and orange lines in the figure, where we plot the equation $\|x_{\text{last}}\|_1/\|x_{\text{OLS}}\|_1=s$ as a dashed black line. We can observe that $\|x_{\text{last}}\|_1/\|x_{\text{OLS}}\|_1$ is always smaller than $s$ for various $s$ values. We illustrates that the final iterates for both PSGLD and PSGULMC stay in the constrained set $\mathcal{C}$ as expected. In Figure~\ref{fig:var-dep}, we also plotted the (expected) average number of iterations required for achieving a target MSE value, as the target value is varied for both PSGLD and PSGULMC algorithms. Here, the dimension $d$ is fixed and is determined by the Diabetes dataset. Although we do not provide theoretical guarantees for the MSE, comparing both algorithms in practice, PSGLD admits slightly better accuracy (measured in terms of MSE) on this example for the same number of iterations.
\subsection{Bayesian Constrained Deep Learning}

Non-Bayesian formulation of deep learning is based on minimizing the so-called \emph{empirical risk} $f:=\frac{1}{n}\sum_{i=1}^n f_i(x,z_i)$ where $f_i$ is the loss function corresponding to the $i$-the data point based on the dataset $z=(z_1,z_2,\dots ,z_n)$ and has a particular structure as a composition of non-linear but smooth functions when smooth activation functions (such as the sigmoid function or the ELU function) are used \citep{clevert2015fast}. Furthermore, here $x$ denotes the weights of the neural network and is a concatenation of vectors $x=\begin{bmatrix} x^{(1)},  x^{(2)}, \dots,  x^{(I)} \end{bmatrix}$ \citep{hu2020non} where $x^{(i)}$ are the (vectorized) weights of the $i$-th layer for $i=1,2,\dots, I$ and $I$ is the number of layers. We refer the reader to \citet{deisenroth2020mathematics} for the details. 

Constraining the weights $x$ to lie on a compact set has been proposed in the deep learning practice for regularization purposes \citep{goodfellow2016regularization}. From the Bayesian sampling perspective, instead of minimizing the empirical risk function, we are interested in sampling from the posterior distribution $\pi(x)\propto e^{-f}$ (see, e.g., \citet{gurbuzbalaban2021decentralized} for a similar approach) subject to constraints. We first consider the unconstrained setting where we run SGLD for 400 epochs and draw 20 samples from the posterior. We let $x_{\text{optimal}} =\begin{bmatrix} x_{\text{optimal}}^{(1)},  x_{\text{optimal}}^{(2)}, \dots,  x_{\text{optimal}}^{(I)} \end{bmatrix}$ denote the average of the samples, which is an approximation to the solution of the unconstrained minimization problem.
We consider the constraints $\|x^{(i)}\|_p \leq s\|x^{(i)}_{\text{optimal}}\|_p$ for the $i$-th layer in the network with $p=1$. Since $\ell_1$-norm promotes sparsity \citep{hastie2009elements}, by adding these layer-wise constraints, we expect to get a sparser model compared to the original model. Sparser models can be preferable as they would be more memory efficient, if they have similar prediction power \citep{srivastava2014dropout}.

Note that $f$ will be smooth on the constraint set if smooth activation functions are used in which case our theory will apply. In our experiments, we use a four-layer fully connected network with hidden layer width $d=400$ on the MNIST dataset.\protect\footnote{This dataset is available online at \url{http://yann.lecun.com/exdb/mnist/}.} The results are shown in Table~\ref{tab:fcn_MNIST_SGLD} and Table~\ref{tab:fcn_MNIST_SGHMC}, 
where the results are based on the average of 20 independent samples. We set the stepsize $\eta=10^{-7}$ for PSGLD and SGLD methods, and we decay $\eta$ by $10\%$ every 100 epochs. We set penalty term $\delta=0.1$ and report results after 350 epochs. 
For PSGULMC and SGULMC methods, we set $\gamma=0.1$ and the stepsize $\eta=5\times 10^{-8}$, which decreases $10\%$ every 100 epochs. We set penalty term $\delta = 100$ and report results after 400 epochs. For both PSGLD and PSGULMC, we use the batch size $b=128$. In Table~\ref{tab:fcn_MNIST_SGLD} 
we report the accuracy of the prediction in the training and test datasets, where we compared SGLD (without any constraints) to PSGLD algorithms with constraints defined by $s=0.9$ and $s=0.8$. We also report the maximum values of $\hat{s}$ among 20 samples, where $\hat{s}$ is defined as $\hat{s}:=\frac{\left\|x\right\|_{1}}{\left\|x_{\text{optimal}}\right\|_{1}}$ and $x$ are the parameters from the last iteration, after running the algorithms for 400 epochs. 
We can see from the results that for different $s$ values, the value of $\hat{s}$ is always smaller than $s$, which indicates that the parameters of our algorithms satisfy the constraints. Table \ref{tab:fcn_MNIST_SGHMC} reports similar results for PSGULMC. Basically, by enforcing $\ell_1$ constraints, we can make the models sparser with a smaller $\ell_1$ constraint at the cost of a relatively small decrease in training and test accuracy.


\begin{table}[h!]
\centering 
\begin{tabular}{lccc}
\toprule
\textbf{}           & \begin{tabular}{@{}c@{}}\textbf{training} \\ \textbf{accuracy}\end{tabular} & \begin{tabular}{@{}c@{}}\textbf{testing} \\ \textbf{accuracy}\end{tabular} &\textbf{$\hat{s}:= \frac{\left\|x\right\|_{1}}{\left\|x_{\text{optimal}}\right\|_{1}}$} \\ \midrule
SGLD                & 90.60\%           & 89.95\%   & 1        \\
PSGLD ($s$=0.9)     & 89.37\%           & 88.89\%   & 0.8954     \\
PSGLD ($s$=0.8)     & 87.35\%           & 87.80\%   & 0.7999     \\ \bottomrule
\end{tabular}
\caption{Training and testing accuracy of fully-connected network with different constraints using PSGLD based on 20 samples.}
\label{tab:fcn_MNIST_SGLD}
\end{table}

\begin{table}[h!]
\centering 
\begin{tabular}{lccc}
\toprule
\textbf{}           & \begin{tabular}{@{}c@{}}\textbf{training} \\ \textbf{accuracy}\end{tabular} & \begin{tabular}{@{}c@{}}\textbf{testing} \\ \textbf{accuracy}\end{tabular} &\textbf{$\hat{s}:= \frac{\left\|x\right\|_{1}}{\left\|x_{\text{optimal}}\right\|_{1}}$} \\ \midrule
SGULMC                & 89.88\%           & 90.22\%   & 1        \\
PSGULMC ($s$=0.9)    & 89.72\%           & 89.49\%   & 0.8918 \\
PSGULMC ($s$=0.8)    & 87.28\%           & 87.80\%          & 0.7931  \\ \bottomrule
\end{tabular}
\caption{Training and testing accuracy of fully-connected network with different constraints using PSGULMC based on 20 samples.}
\label{tab:fcn_MNIST_SGHMC}
\end{table}

\section{Conclusion}
In this paper, we considered the problem of constrained sampling where the goal is to sample from a target distribution $\pi(x)\propto e^{-f(x)}$ when $x$ is constrained to lie on a convex body $\mathcal{C}$. We proposed and studied penalty-based overdamped Langevin and underdamped Langevin Monte Carlo (ULMC) methods. We considered targets where $f$ is smooth and strongly convex as well as the more general case where $f$ can be non-convex. In both cases, under some assumptions, we characterized the number of iterations and samples required to sample the target up to an $\varepsilon$-error while the error is measured in terms of the 2-Wasserstein or the total variation distance. Our methods improve upon the dimension dependency of the existing approaches in a number of settings and to our knowledge provides the first convergence results for ULMC-based methods for non-convex $f$ in the context of constrained sampling. Our methods can also handle unbiased stochastic noise on the gradients that arise in machine learning applications. Finally, we illustrated the efficiency of our methods on the Bayesian Lasso linear regression and Bayesian deep learning problems.

\section*{Acknowledgements}
The authors thank the acting editor and two anonymous referees for helpful comments and suggestions.
The authors also thank Sam Ballas and Andrzej Ruszczy\'{n}ski for helpful discussions.
Mert G\"urb\"uzbalaban and Yuanhan Hu's research is partly supported by
the grants Office of Naval Research Award Number N00014-21-1-2244, National Science Foundation (NSF) CCF-1814888, NSF DMS-1723085, NSF DMS-2053485. 
Lingjiong Zhu is partially supported by the grants NSF DMS-2053454, NSF DMS-2208303, and a Simons Foundation Collaboration Grant. 

\bibliography{langevin}

\newpage

\appendix


\section{Notations}\label{sec:notations}

A function $f:\mathbb{R}^{d}\rightarrow\mathbb{R}$ is said to be $\mu$-strongly convex
if there exists $\mu>0$ such that for any $x,y\in\mathbb{R}^{d}$,
$$
f(x)-f(y)- g^{\top}(x-y)
\geq
\frac{\mu}{2}\Vert x-y\Vert^{2},
\qquad\text{for all $g\in \partial f(y)$},
$$
where $\partial f$ denotes the subdifferential. If $f$ is differentiable at $y$, then $\partial f(y)=\{\nabla f(y)\}$ is a singleton set. If the latter inequality holds for $\mu=0$, we say $f$ is merely convex (see e.g. \cite{nesterov2013introductory}).

The function $f:\mathbb{R}^{d}\rightarrow\mathbb{R}$ is $L$-smooth
if for any $x,y\in\mathbb{R}^{d}$, the gradients $\nabla f(x),\nabla f(y)$ exist and satisfy $\|\nabla f(x) - \nabla f(y) \|\leq L\|x-y\|$. If $f$ is both $\mu$-strongly convex and $L$-smooth, it holds that (see e.g. \citet{bubeck-mono}):
$$
\frac{\mu}{2}\|x-y\|^2 \leq f(x)-f(y)-\nabla f(y)^{\top}(x-y) \leq\frac{L}{2}\Vert x-y\Vert^{2},
\qquad\text{for any $x,y\in\mathbb{R}^{d}$}.
$$
We say that a function $f:\mathbb{R}^{d}\rightarrow\mathbb{R}$ is $(m,b)$-dissipative
if for some $m,b>0$,
$\langle\nabla f(x),x\rangle\geq m\Vert x\Vert^{2}-b, \,\text{for any $x\in\mathbb{R}^{d}$}.$

For any $x,y\in\mathbb{R}$, $x\vee y$ denotes $\max(x,y)$
and $x\wedge y$ denotes $\min(x,y)$.
For any $x=(x_{1},\ldots,x_{d})\in\mathbb{R}^{d}$, its $\ell_{p}$-norm (also referred to as $p$-norm)
is denoted by $\Vert x\Vert_{p}:=\left(\sum_{i=1}^{d}|x_{i}|^{p}\right)^{1/p}$.
For any measurable set $\mathcal{A}\subset\mathbb{R}^{d}$, 
we use $|\mathcal{A}|$ to denote the Lebesgue measure
of $\mathcal{A}$. For a set $\mathcal{A}$, the indicator function $1_\mathcal{A}(y)=1$ for $y\in \mathcal{A}$ and 
$1_\mathcal{A}(y)=0$ otherwise. We denote $\mathbb{R}_{\geq 0}$ the set of non-negative real scalars.

A subset $\mathcal{C}$ of $\mathbb{R}^d$ is called a hypersurface of class $C^k$, if for every $x_0\in \mathcal{C}$ there is an open set $V\subset\mathbb{R}^d$ containing $x_0$ and a real-valued function $\phi\in C^k(V)$ such that $\nabla \phi$ is non-vanishing on $S\cap V = \{x\in V:\phi(x)=0\}$, where $C^k(V)$ is the set of functions defined on $V$ with $k$ continuous derivatives. We denote $D n(\xi)$ and $D^2 n(\xi)$ as the first and second-order derivatives of unit normal vector $n$ in the sense of \citet{leobacher2021existence}.

Next, we introduce three standard notions often used
to quantify the distances between two probability measures.
For a survey on distances between two probability measures, 
we refer to \citet{GS2002}.

\textbf{Wasserstein metric.}
For any $p\geq 1$, define $\mathcal{P}_{p}(\mathbb{R}^{d})$
as the space consisting of all the Borel probability measures $\nu$
on $\mathbb{R}^{d}$ with the finite $p$-th moment
(based on the Euclidean norm).
For any two Borel probability measures $\nu_{1},\nu_{2}\in\mathcal{P}_{p}(\mathbb{R}^{d})$, 
we define the standard $p$-Wasserstein
metric \citep{villani2008optimal}:
$\mathcal{W}_{p}(\nu_{1},\nu_{2}):=\left(\inf\mathbb{E}\left[\Vert Z_{1}-Z_{2}\Vert^{p}\right]\right)^{1/p},$
where the infimum is taken over all joint distributions of the random variables $Z_{1},Z_{2}$ with marginal distributions
$\nu_{1},\nu_{2}$.

\textbf{Kullback-Leibler (KL) divergence.}
KL divergence, also known as relative entropy,
between two probability measures  
$\mu$ and $\nu$ on $\mathbb{R}^{d}$, where $\mu$ is absolutely continuous with respect to $\nu$, is defined as:
$D(\mu\Vert\nu):=\int_{\mathbb{R}^{d}}\frac{d\mu}{d\nu}\log\left(\frac{d\mu}{d\nu}\right)d\nu.$

\textbf{Total variation distance.}
The total variation (TV) distance between two probability measures $P$ and $Q$ on a sigma-algebra $\mathcal{F}$ is defined as $\sup_{A\in \mathcal{F}} |P(A)-Q(A)|$.



\section{Weighted Csisz\'{a}r-Kullback-Pinsker Inequality}

The KL divergence can bound the Wasserstein distances
on $\mathbb{R}^{d}$ under some technical conditions, known as 
the weighted Csisz\'{a}r-Kullback-Pinsker (W-CKP) inequality.

\begin{lemm}[page 337 in \citet{BV}]\label{lem:BV}
For any two probability measures $\mu$ and $\nu$ on $\mathbb{R}^{d}$,
we have
\begin{equation}
\mathcal{W}_{2}(\mu,\nu)
\leq\hat{C}\left(D(\mu\Vert\nu)^{\frac{1}{2}}+\left(\frac{D(\mu\Vert\nu)}{2}\right)^{\frac{1}{4}}\right),
\end{equation}
where 
$\hat{C}:=2\inf_{\hat{x}\in\mathbb{R}^{d},\hat{\alpha}>0}\left(\frac{1}{\hat{\alpha}}\left(\frac{3}{2}+\log\int_{\mathbb{R}^{d}}e^{\hat{\alpha}\Vert x-\hat{x}\Vert^{2}}d\nu(x)\right)\right)^{\frac{1}{2}}$,
provided that there exists some $\hat{\alpha}>0$ and $\hat{x}\in\mathbb{R}^{d}$ such that $\int_{\mathbb{R}^{d}}e^{\hat{\alpha}\Vert x-\hat{x}\Vert^{2}}d\nu(x)<\infty$.
\end{lemm}


\section{Technical Lemmas}

In this section, we provide some technical lemmas that are used in the proofs of the main results.
The proofs of these technical lemmas will be provided in Appendix~\ref{sec:proofs}.

\begin{lemm}\label{lem-S-dissipative} 
If Assumption~\ref{assump:C} holds, 
then the penalty function $S(x)=\left(\delta_{\mathcal{C}}(x)\right)^{2}$ is continuously differentiable, $\ell$-smooth with $\ell=4$
and $(m_S, b_S)$-dissipative with $m_S = 1, b_S = R^2/4$, i.e.
$\langle x,\nabla S(x)\rangle\geq m_{S}\Vert x\Vert^{2}-b_{S},\ \text{for any $x\in\mathbb{R}^{d}$}$.
\end{lemm}

\begin{lemm}\label{lem:f:plus:S:dissipative}
If Assumption~\ref{assump:f:2} and Assumption~\ref{assump:C} hold, 
then $f+\frac{1}{\delta}S$ 
is $L_{\delta}$-smooth, 
with $L_{\delta}:=L+\frac{\ell}{\delta}$
and moreover $f+\frac{1}{\delta}S$  is $(m_{\delta},b_{\delta})$-dissipative 
with 
\begin{equation}\label{m:b:delta}
m_{\delta}:=-L-\frac{1}{2}+\frac{m_{S}}{\delta}>0,
\qquad
b_{\delta}:=\frac{1}{2}\Vert\nabla f(0)\Vert^{2}+\frac{b_{S}}{\delta},
\end{equation}
provided that $\delta<m_{S}/(L+\frac{1}{2})$, where $m_{S},b_{S}$ are defined in Lemma~\ref{lem-S-dissipative}.
\end{lemm}

\begin{lemm}\label{lem-dissipative-implies-lower-bound}
Under Assumption~\ref{assump:f:2} and Assumption~\ref{assump:C}, then
$f+\frac{S}{\delta}$ is lower bounded for $S(x)=\left(\delta_{\mathcal{C}}(x)\right)^{2}$, i.e. there exists a real non-negative scalar $M$ such that $f(x)+\frac{S(x)}{\delta}\geq -M$ for any $x\in\mathbb{R}^{d}$, 
where we can take
\begin{align}\label{M:eqn}
M:=-f(0)+\frac{1}{2}\Vert\nabla f(0)\Vert^{2}+\frac{b_{S}}{2\delta}\log 3,
\end{align}
provided that $\delta\leq\frac{2m_{S}}{3(1+L)}$ where $m_{S},b_{S}$ are defined in Lemma~\ref{lem-S-dissipative}.
\end{lemm}

\begin{lemm}\label{condition:thm:final}
If Assumption~\ref{assump:f:2} and Assumption~\ref{assump:C} hold, 
then the conditions in Theorem~\ref{thm:final} are satisfied
with $\hat{\alpha}=\frac{m_{\delta}}{6}$ and $\hat{x}=0$, 
where $m_{\delta}$ is defined in \eqref{m:b:delta}.
\end{lemm}

\begin{lemm}\label{condition:thm:final:2}
If Assumptions~\ref{assump:f:1} and \ref{assump:C} hold, 
then the assumptions in Theorem~\ref{thm:final} are satisfied with $\hat{\alpha}=\frac{\mu}{4}$ and $\hat{x}=x_{\ast}$, 
where $x_{\ast}$ is the unique minimizer of $f$.
\end{lemm}

\section{Technical Proofs}\label{sec:proofs}

In this section, we provide technical proofs of the main results in our paper.

%



\subsection*{Proof of Lemma~\ref{lem:1}}

Note that $\pi$ is supported on $\mathcal{C}$
whereas $\pi_{\delta}$ is supported on $\mathbb{R}^{d}$,
and $\pi$ is absolutely continuous with respect to $\pi_{\delta}$. 
We can compute that the KL divergence between $\pi$ and $\pi_{\delta}$ is given by
\begin{align}
&D(\pi\Vert\pi_{\delta})
\nonumber
\\
&=\int_{\mathbb{R}^{d}}\log\left(\frac{\pi(x)}{\pi_{\delta}(x)}\right)\pi(x)dx
=\int_{\mathcal{C}}\log\left(e^{\frac{1}{\delta}S(x)}\frac{\int_{\mathbb{R}^{d}}e^{-f(y)-\frac{1}{\delta}S(y)}dy}{\int_{\mathcal{C}}e^{-f(y)}dy}\right)\frac{e^{-f(x)}}{\int_{\mathcal{C}}e^{-f(y)}dy}dx
\label{we:used:1}
\\
&=\int_{\mathcal{C}}\log\left(\frac{\int_{\mathbb{R}^{d}}e^{-f(y)-\frac{1}{\delta}S(y)}dy}{\int_{\mathcal{C}}e^{-f(y)}dy}\right)\frac{e^{-f(x)}}{\int_{\mathcal{C}}e^{-f(y)}dy}dx,
\label{we:used:2}
\\
&=\log\left(\frac{\int_{\mathbb{R}^{d}}e^{-f(y)-\frac{1}{\delta}S(y)}dy}{\int_{\mathcal{C}}e^{-f(y)}dy}\right),\label{further:compute}
\end{align}
where we used the definition of $\pi$ and $\pi_{\delta}$ to obtain \eqref{we:used:1}
and the fact that $S(x)=0$ for any $x\in\mathcal{C}$ to obtain \eqref{we:used:2}.
We can further compute from \eqref{further:compute} that
\begin{align}
D(\pi\Vert\pi_{\delta})
&=\log\left(\frac{\int_{\mathcal{C}}e^{-f(y)-\frac{1}{\delta}S(y)}dy+\int_{\mathbb{R}^{d}\backslash\mathcal{C}}e^{-f(y)-\frac{1}{\delta}S(y)}dy}{\int_{\mathcal{C}}e^{-f(y)}dy}\right)
\nonumber
\\
&=\log\left(1+\frac{\int_{\mathbb{R}^{d}\backslash\mathcal{C}}e^{-f(y)-\frac{1}{\delta}S(y)}dy}{\int_{\mathcal{C}}e^{-f(y)}dy}\right)
\leq\frac{\int_{\mathbb{R}^{d}\backslash\mathcal{C}}e^{-\frac{1}{\delta}S(y)-f(y)}dy}{\int_{\mathcal{C}}e^{-f(y)}dy},
\label{we:used:3}
\end{align}
where we used the fact that $S(y)=0$ for any $y\in\mathcal{C}$ to obtain the equality in \eqref{we:used:3}
and $\log(1+x)\leq x$ for any $x\geq 0$ to obtain the inequality in \eqref{we:used:3}. 
This completes the proof.
\hfill $\Box$


\subsection*{Proof of Lemma~\ref{lem:geometry}}
By the definitions of $S(y)$ and $g$, we have
\begin{equation*}
\left|y\in\mathbb{R}^{d}\backslash\mathcal{C}: S(y)\leq\epsilon\right|
=\left|y\in\mathbb{R}^{d}\backslash\mathcal{C}: \delta_{\mathcal{C}}(y)\leq\delta\right|,
\end{equation*}
with
$\delta:= g^{-1}(\epsilon)$,
where $g^{-1}$ denotes the inverse function of $g$ which exists due to the assumptions
on $g$.
Translate $\mathcal{C}$ so that the largest ball it contains is centered
at $0$. The set
$\left(1+\frac{\delta}{r}\right)\mathcal{C}
=\mathcal{C}+\frac{\delta}{r}\mathcal{C}$
contains the $\delta$-neighborhood of $\mathcal{C}$
since $\frac{\delta}{r}\mathcal{C}$ contains
a ball of radius $\delta$. The volume of 
$\left(1+\frac{\delta}{r}\right)\mathcal{C}$ is $(1+\delta/r)^{d}|\mathcal{C}|$, 
where we used the fact that for any Lebesgue measurable set $A$ in $\mathbb{R}^{d}$ the dilation of $A$ by $\lambda>0$ defined as $\lambda A$
is also Lebesgue measurable with the Lebesgue measure $\lambda^{d}|A|$. 
Therefore, the volume of the set of all points that do not belong to $\mathcal{C}$
but lie within distance at most $\delta$ from it has volume at most
$\left(\left(1+\frac{\delta}{r}\right)^{d}-1\right)|\mathcal{C}|$.
The proof is complete.
\hfill $\Box$


\subsection*{Proof of Lemma~\ref{lem:D}}

First, we recall from Lemma~\ref{lem:1} that the KL divergence between $\pi$ and $\pi_{\delta}$ is bounded by:
$D(\pi\Vert\pi_{\delta})
\leq\frac{\int_{\mathbb{R}^{d}\backslash\mathcal{C}}e^{-\frac{1}{\delta}S(y)-f(y)}dy}{\int_{\mathcal{C}}e^{-f(y)}dy}$.
It is easy to compute that for any $\theta>0$,
\begin{align}
&\int_{\mathbb{R}^{d}\backslash\mathcal{C}}e^{-\frac{1}{\delta}S(y)-f(y)}dy
\nonumber
\\
&=\int_{y\in\mathbb{R}^{d}\backslash\mathcal{C}: S(y)\leq\theta}e^{-\frac{1}{\delta}S(y)-f(y)}dy
+\int_{y\in\mathbb{R}^{d}\backslash\mathcal{C}: S(y)>\theta}e^{-\frac{1}{\delta}S(y)-f(y)}dy
\nonumber
\\
&\leq\left|y\in\mathbb{R}^{d}\backslash\mathcal{C}: S(y)\leq\theta\right|e^{-\inf_{y\in\mathbb{R}^{d}\backslash\mathcal{C}:S(y)\leq\theta}f(y)}
+e^{-\frac{\theta}{\delta}}\int_{\mathbb{R}^{d}\backslash\mathcal{C}}e^{-\frac{1}{\delta}S(y)-f(y)}dy,\label{obtain:ineq}
\end{align}
where we used $S(y)\geq 0$ for any $y\in\mathbb{R}^{d}$ 
to obtain the inequality \eqref{obtain:ineq}.

By taking $\theta=\tilde{\alpha}\delta\log(1/\delta)$ with $\tilde{\alpha}>0$ in \eqref{obtain:ineq}, and by applying Lemma~\ref{lem:geometry}, we have
\begin{align*}
&\int_{\mathbb{R}^{d}\backslash\mathcal{C}}e^{-\frac{1}{\delta}S(y)-f(y)}dy
\\
&\leq\left|y\in\mathbb{R}^{d}\backslash\mathcal{C}: S(y)\leq\tilde{\alpha}\delta\log(1/\delta)\right|
e^{-\inf_{y\in\mathbb{R}^{d}\backslash\mathcal{C}:S(y)\leq\tilde{\alpha}\delta\log(1/\delta)}f(y)}
+\delta^{\tilde{\alpha}}\int_{\mathbb{R}^{d}\backslash\mathcal{C}}e^{-\frac{1}{\delta}S(y)-f(y)}dy
\\
&\leq
\left(\left(1+\frac{g^{-1}(\tilde{\alpha}\delta\log(1/\delta))}{r}\right)^{d}-1\right)|\mathcal{C}|
e^{-\inf_{y\in\mathbb{R}^{d}\backslash\mathcal{C}:S(y)\leq\tilde{\alpha}\delta\log(1/\delta)}f(y)}
\\
&\qquad\qquad\qquad\qquad\qquad\qquad
+\delta^{\tilde{\alpha}}\int_{\mathbb{R}^{d}\backslash\mathcal{C}}e^{-\frac{1}{\delta}S(y)-f(y)}dy
\\
&\leq
\left(\left(1+\frac{g^{-1}(\tilde{\alpha}\delta\log(1/\delta))}{r}\right)^{d}-1\right)\frac{\pi^{d/2}}{\Gamma(\frac{d}{2}+1)}R^{d}
e^{-\inf_{y\in\mathbb{R}^{d}\backslash\mathcal{C}:S(y)\leq\tilde{\alpha}\delta\log(1/\delta)}f(y)}
\\
&\qquad\qquad\qquad\qquad\qquad\qquad\qquad\qquad\qquad
+\delta^{\tilde{\alpha}}\int_{\mathbb{R}^{d}\backslash\mathcal{C}}e^{-\frac{1}{\delta}S(y)-f(y)}dy,
\end{align*}
where we used the fact that $\mathcal{C}$ is contained in an Euclidean ball with radius $R$ (Assumption~\ref{assump:C})
so that $|\mathcal{C}|$ is less than or equal to the volume of a Euclidean ball with radius $R$
which is $\frac{\pi^{d/2}}{\Gamma(\frac{d}{2}+1)}R^{d}$, where $\Gamma$ denotes the gamma function. 
The proof is complete.
\hfill $\Box$

\subsection*{Proof of Lemma~\ref{lem:S}} 

Since $\mathcal{C}$ is convex, for every $x \in \mathbb{R}^d$ there exists a unique point of $\mathcal{C}$
nearest to $x$. 
Then the fact that $S(x)=\left(\delta_{\mathcal{C}}(x)\right)^2$
is convex, $\ell$-smooth and continuously differentiable with a gradient $\nabla S(x) = 2 (x - \mathcal{P}_{\mathcal{C}}(x))$ is a direct consequence of \citet[Theorem~4.8]{federer1959curvature}. 
To show that $S(x)$ is convex, consider two points $x_1$ and $x_2 \in \mathbb{R}^d$, and their projections $c_1$ and $c_2$ to the set $\mathcal{C}$. By the convexity of the set $\mathcal{C}$, we have $\bar{c}:=\frac{(c_1 + c_2)}{2} \in \mathcal{C}$ and by the definition of $S$, we obtain
\begin{align*} 
S\left( \frac{x_1 + x_2}{2}\right) &\leq  \left\|\frac{x_1 + x_2}{2} - \bar{c}\right\|^2 = \frac{\| (x_1 - c_1) + (x_2 - c_2)\|^2}{4} \\
&\leq \frac{2\|x_1 - c_1\|^2 + 2 \|x_2 - c_2\|^2}{4} = \frac{S(x_1) + S(x_2)}{2}, 
\end{align*}
where we used the inequality $\Vert a+b\Vert^2 \leq 2\Vert a\Vert^2 + 2\Vert b\Vert^2$ for any two vectors $a,b$ in the last inequality. Finally, note that by the triangle inequality, 
$$\left\| \nabla S(y) - \nabla S(x) \right\| \leq 2 \|y-x\|  + 2 \left\|P_\mathcal{C}(y)-P_\mathcal{C}(x)\right\| \leq 4 \|y-x\|,$$
where we used the non-expansiveness of the projection step. Therefore, we can take the smoothness constant of $S(x)$ to be $\ell = 4$. This completes the proof.
\hfill $\Box$


\subsection*{Proof of Theorem~\ref{thm:final}}

By weighted Csisz\'{a}r-Kullback-Pinsker (W-CKP) inequality (see Lemma~\ref{lem:BV}), we have
\begin{equation}\label{ineq:CKP}
\mathcal{W}_{2}(\pi,\pi_{\delta})
\leq\hat{C}\left(D(\pi\Vert\pi_{\delta})^{\frac{1}{2}}+\left(\frac{D(\pi\Vert\pi_{\delta})}{2}\right)^{\frac{1}{4}}\right),
\end{equation}
where 
$\hat{C}:=2\inf_{\hat{x}\in\mathbb{R}^{d},\hat{\alpha}>0}\left(\frac{1}{\hat{\alpha}}\left(\frac{3}{2}+\log\int_{\mathbb{R}^{d}}e^{\hat{\alpha}\Vert x-\hat{x}\Vert^{2}}d\pi_{\delta}(x)\right)\right)^{\frac{1}{2}}<\infty$,
provided that there exists some $\hat{\alpha}>0$ and $\hat{x}\in\mathbb{R}^{d}$ so that 
$\int_{\mathbb{R}^{d}}e^{\hat{\alpha}\Vert x-\hat{x}\Vert^{2}}d\pi_{\delta}(x)<\infty$.
Furthermore, we can compute the following:
\begin{equation}
\int_{\mathbb{R}^{d}}e^{\hat{\alpha}\Vert x-\hat{x}\Vert^{2}}d\pi_{\delta}(x)
=\frac{\int_{\mathbb{R}^{d}}e^{\hat{\alpha}\Vert x-\hat{x}\Vert^{2}}e^{-f(x)-\frac{S(x)}{\delta}}dx}
{\int_{\mathbb{R}^{d}}e^{-f(x)-\frac{S(x)}{\delta}}dx}
\leq
\frac{\int_{\mathbb{R}^{d}}e^{\hat{\alpha}\Vert x-\hat{x}\Vert^{2}}e^{-f(x)-\frac{S(x)}{\delta}}dx}
{\int_{\mathcal{C}}e^{-f(x)}dx}<\infty,
\end{equation}
provided that $\int_{\mathbb{R}^{d}}e^{\hat{\alpha}\Vert x-\hat{x}\Vert^{2}}e^{-f(x)-\frac{S(x)}{\delta}}dx<\infty$
which is increasing in $\delta$ and hence uniformly bounded as $\delta\rightarrow 0$.

We now take $g(x)=x^{2}$ in Lemma~\ref{lem:D}
with $S(x)=(\delta_{\mathcal{C}}(x))^{2}$
so that by Lemma~\ref{lem:S}, we have that $S$ is convex, $\ell$-smooth
and continuously differentiable. Moreover,  
since $S(x)=(\delta_{\mathcal{C}}(x))^{2}$ and $f$ is continuous
and the set $\{y\in\mathbb{R}^{d}:S(y)\leq\tilde{\alpha}\delta\log(1/\delta)\}$ is compact and
we have that in equation~\eqref{second:upper:bound} in Lemma~\ref{lem:D},
\begin{equation*}
\inf_{y\in\mathbb{R}^{d}\backslash\mathcal{C}:S(y)\leq\tilde{\alpha}\delta\log(1/\delta)}f(y)
\geq\inf_{y\in\mathbb{R}^{d}:S(y)\leq\tilde{\alpha}\delta\log(1/\delta)}f(y)
=\min_{y\in\mathbb{R}^{d}:S(y)\leq\tilde{\alpha}\delta\log(1/\delta)}f(y)>-\infty,
\end{equation*}
and it is uniform in $\delta$ as $\delta\rightarrow 0$
and hence by applying Lemma~\ref{lem:D} we obtain
\begin{equation}\label{D:ineq}
D(\pi\Vert\pi_{\delta})
\leq\mathcal{O}\left(\left(\delta\log(1/\delta)\right)^{1/2}\right).
\end{equation}
Finally, we get the desired result by plugging \eqref{D:ineq} into W-CKP inequality \eqref{ineq:CKP}. The proof is complete.
\hfill $\Box$


\subsection*{Proof of Lemma~\ref{lemma:O:alpha}}
Recall that we have the representation $\mathcal{C} = \{ x: h(x)\leq 0\}$ given in \eqref{def-constraint-set-with-ineq} where $h(x)=\max_{1\leq i \leq m} h_i(x)$ for some $m\geq 1$ with $h_i:\mathbb{R}^d\to\mathbb{R}$ being convex for $i=1,2,\dots,m$. Furthermore, $h(0)\leq 0$ as we assumed in Assumption~\ref{assump:C} that $0 \in \mathcal{C}$. We first define the function $p_m: \mathbb{R}^d \to \mathbb{R}_{\geq 0}$, 
$$ p_m(x) := \inf \{t\geq 0: h_i(x/t) \leq 0 \mbox{ for every } i=1,2,\dots,m\}.
$$
By the convexity of $h_i$, it is easy to check that $p_m$ is subadditive satisfying $p_m(x+y) \leq p_m(x)+p_m(y)$ for every $x$ and $y$, and it is homogeneous with $p_m(sx)=s p_m(x)$ for any $x$ and scalar $s\geq 0$. Therefore, $p_m$ is convex and consequently locally Lipschitz continuous \citep{cvx-locally-lip} and Lipschitz continuous on compact sets. The function $h(x)$ is also convex; hence, there exists a positive constant $B$ such that $\| y\|\leq B$ for any $y \in \partial h(x)$ and $x\in \mathcal{C}$.
We note that there exist
some constants $c_{K},C_{K}>0$
such that 
\begin{equation*}
c_{K} p_m(x) \leq\Vert x\Vert\leq C_{K} p_m(x),\qquad\text{for any $x\in\mathbb{R}^{d}$},
\end{equation*}
where $\Vert x\Vert$ is the Euclidean norm of $x\in\mathbb{R}^{d}$. 
To show this, let $\text{bd}(\mathcal{C})$ denote
the boundary of $\mathcal{C}$ and let
\begin{equation*}
c_{K}:=\min\{\Vert x\Vert: x\in\text{bd}(\mathcal{C})\}
\qquad\text{and}\qquad
C_{K}:=\max\{\Vert x\Vert: x\in\text{bd}(\mathcal{C})\}.
\end{equation*}
Note that $p_m(x)=1$ for $x\in\text{bd}(\mathcal{C})$
and furthermore $p_m$ is homogeneous. 
For any $x$, there exists $t>0$ such that $tx\in\text{bd}(\mathcal{C})$. 
Moreover, $c_{K}\leq\Vert tx\Vert\leq C_{K}$ and $p_m( tx)=1$. 
Therefore, $p_m(x)=1/t$ and $c_{K}/t\leq\Vert x\Vert\leq C_{K}/t$. 
Hence, we showed that 
\begin{equation*}
c_{K}p_m(x)\leq p_m(tx) \leq C_{K}p_m(x).
\end{equation*}
Next, we can compute that
\begin{equation*}
\left|x:-\frac{\alpha}{2}\Vert x\Vert^{2}\leq h(x)\leq 0\right|
=\left|x:1-\frac{\alpha}{2}\Vert x\Vert^{2}\leq p_m(x) \leq 1\right|.
\end{equation*}
For any $x$ such that $p_m(x)\leq 1$, we have $\Vert x\Vert\leq C_{K}$. Thus,
for any $x$ such that 
$p_m(x)\geq 1-\frac{\alpha}{2}\Vert x\Vert^{2}$ and $p_m(x)\leq 1$, we have
$p_m(x)\geq 1-\frac{\alpha}{2}C_{K}^{2}$, 
which implies that
\begin{equation*}
\left|x:1-\frac{\alpha}{2}\Vert x\Vert^{2}\leq p_m(x) \leq 1\right|
\leq
\left|x:1-\frac{\alpha}{2}C_{K}^{2}\leq p_m(x)\leq 1\right|,
\end{equation*}
provided that $\alpha<2/C_{K}^2$. Furthermore, by the definition of the functional $p_m(x)$, 
we have $p_m(x)\leq 1$ if and only if $x\in\mathcal{C}$. 
Therefore, 
\begin{align*}
\left|x:1-\frac{\alpha}{2}C_{K}^{2}\leq p_m(x) \leq 1\right|
&=\left|x: p_m(x) \leq 1\right|
-\left|x: p_m(x) \leq 1-\frac{\alpha}{2}C_{K}^{2}\right|
\\
&=|\mathcal{C}|-\left(1-\frac{\alpha}{2}C_{K}^{2}\right)^{d}|\mathcal{C}|.
\end{align*}
On the other hand, 
\begin{align*}
\left|x:h(x)+\frac{\alpha}{2}\Vert x\Vert^{2}\leq 0\right|
&=\left|x: p_m(x) +\frac{\alpha}{2}\Vert x\Vert^{2}\leq 1\right|
\\
&\geq\left|x: p_m(x) +\frac{\alpha}{2}C_{K}^{2}\Vert x\Vert_{K}^{2}\leq 1\right|
\\
&\geq\left|x: p_m(x) \leq 1-\frac{\alpha}{2}C_{K}^{2}\right|
=\left(1-\frac{\alpha}{2}C_{K}^{2}\right)^{d}|\mathcal{C}|,
\end{align*}
provided that $\alpha<2/C_{K}^{2}$.
Hence, we conclude that
\begin{equation*}
\frac{|\mathcal{C}\backslash\mathcal{C}^{\alpha}|}{|\mathcal{C}^{\alpha}|}
\leq\frac{1-\left(1-\frac{\alpha}{2}C_{K}^{2}\right)^{d}}{\left(1-\frac{\alpha}{2}C_{K}^{2}\right)^{d}}
\leq\mathcal{O}(\alpha),
\end{equation*}
as $\alpha\rightarrow 0$. Therefore, the inequality \eqref{C:alpha:assump} is satisfied, and the proof is complete.
\hfill $\Box$



\subsection*{Proof of Proposition~\ref{cor:LD:convex:outside}}

Before we proceed to the proof of 
Proposition~\ref{cor:LD:convex:outside}, 
we first state a few technical lemmas
whose proofs will be provided at the end
of the Appendix.
The next technical lemma states that
the penalty function $S^{\alpha}(x)$ is strongly-convex
outside a compact domain for $\alpha\geq 0$ if the boundary function $h(x)$ is strongly convex, or for $\alpha>0$ when $h$ is merely convex. 
Before we proceed, let us recall that
since the function $h(x)$ is also convex, there exists a positive constant $B$ such that $\| y\|\leq B$ for any $y \in \partial h(x)$ and $x\in \mathcal{C}$.

\begin{lemm}\label{lemm:strongly convex}
Consider the constrained set $\mathcal{C}^{\alpha}$
that is defined in \eqref{C:alpha} for $\alpha \geq 0$. Let $\beta$ be the strong convexity constant of $h$ with the convention that $\beta=0$ if $h$ is merely convex. If $\alpha +\beta>0$, 
then the penalty function $S^\alpha(x)$ is strongly convex with constant $\frac{2(\alpha+\beta) \rho}{B+(\alpha +\beta) \rho}$ on the set $\mathbb{R}^{d}\backslash U(\mathcal{C}^\alpha,\rho)$, where $U(\mathcal{C}^\alpha, \rho)$ is the open $\rho$-neighborhood of $\mathcal{C}^\alpha$ i.e. $U(\mathcal{C}^\alpha, \rho):=\{x:\text{dist}(x,\mathcal{C}^\alpha)<\rho\}$.
\end{lemm}


We have the following corollary as an immediate consequence of Lemma~\ref{lemm:strongly convex}.

\begin{coro}\label{cor:strongly:convex}
Under Assumption~\ref{assump:f:2} and the assumptions of Lemma~\ref{lemm:strongly convex}, $f+\frac{S^\alpha}{\delta}$ is $\mu_{\delta}$-strongly convex
outside an Euclidean ball with radius $R+\rho$, where $\mu_{\delta}:=\frac{2(\alpha+\beta) \rho}{\delta(B+(\alpha+\beta) \rho)}-L$
provided that $\delta<\frac{2(\alpha+\beta)\rho}{L(B+(\alpha+\beta)\rho)}$.
\end{coro}

When $f+S^{\alpha}/\delta$ is strongly convex outside a compact domain, one can leverage
the non-asymptotic guarantees in \citet{sampling-can-be-faster} for Langevin dynamics
to obtain the following performance guarantees for the penalized Langevin dynamics. 
Before we proceed, we introduce
the following technical lemma, which 
states that $f+\frac{S^{\alpha}}{\delta}$ is close
to a strongly-convex function.

\begin{lemm}\label{lem:close-piecewise-quadratic}
Under the assumptions in Corollary~\ref{cor:strongly:convex}, for any given $m>0$, there exists a $C^{1}$ function $U$
such that $U$ is $s_{0}$-strongly convex on $\mathbb{R}^{d}$ with
\begin{align*}
&\sup_{x\in\mathbb{R}^{d}}\left(U(x)-\left(f(x)+\frac{S^{\alpha}(x)}{\delta}\right)\right)-\inf_{x\in\mathbb{R}^{d}}\left(U(x)-\left(f(x)+\frac{S^{\alpha}(x)}{\delta}\right)\right)
\\
&\leq R_{0}:= 2(R+\rho)^2 \left( \frac{m+L}{2} + \frac{(m+L)^2}{\mu_{\delta}} \right),
\end{align*}
where $s_0 := \min(m, \mu_{\delta}/2)$.
\end{lemm}



Finally, we can proceed to the proof of Proposition~\ref{cor:LD:convex:outside}. 
We first consider the case that $\mathcal{C}=\{x:h(x)\leq 0\}$, 
where $h$ is $\beta$-strongly convex.

First of all, by running the penalized Langevin dynamics \eqref{eq:LD:nonconvex}, we have
\begin{equation*}
\text{TV}(\nu_{K},\pi)\leq \text{TV}(\nu_{k},\pi_{\delta})+\text{TV}(\pi_{\delta},\pi),
\end{equation*}
where $\text{TV}$ standards for the total variation distance.
We recall from \eqref{D:ineq} that in KL divergence:
$D(\pi\Vert\pi_{\delta})
\leq\mathcal{O}\left(\left(\delta\log(1/\delta)\right)^{1/2}\right)$.
By Pinsker's inequality, we have
\begin{equation*}
\text{TV}(\pi_{\delta},\pi)
\leq\sqrt{\frac{1}{2}D(\pi\Vert\pi_{\delta})}
\leq\mathcal{O}\left(\left(\delta\log(1/\delta)\right)^{1/4}\right).
\end{equation*}
Therefore, $\text{TV}(\pi_{\delta},\pi)\leq\tilde{\mathcal{O}}(\varepsilon)$
provided that $\delta=\varepsilon^{4}$.


By Lemma~\ref{lem:f:plus:S:dissipative}, $f+S/\delta$ is $L_{\delta}$-smooth with $L_{\delta}:=L+\frac{\ell}{\delta}$.
Note that in Lemma~\ref{lem:close-piecewise-quadratic}, we showed that there exists a $C^{1}$ function $U$
that is $s_{0}$-strongly convex and satisfies
\begin{equation*}
\sup_{x\in\mathbb{R}^{d}}\left(U(x)-f(x)-\frac{S(x)}{\delta}\right)-\inf_{x\in\mathbb{R}^{d}}\left(U(x)-f(x)-\frac{S(x)}{\delta}\right)\leq R_{0}.
\end{equation*}
By Proposition~2 in \citet{sampling-can-be-faster}, $\pi_{\delta}$ satisfies a log-Sobolev inequality
with constant $\rho_{\ast}\geq s_{0}e^{-R_{0}}$. 
Moreover, with $\delta=\varepsilon^{4}$,
we recall from Lemma~\ref{lem:f:plus:S:dissipative}
that $s_{0}=\min(m,\mu_{\delta}/2)$
and $R_{0}:=2(R+\rho)^{2}\left(\frac{m+L}{2}+\frac{(m+L)^{2}}{\mu_{\delta}}\right)$ so that $\mu_{\delta}=\frac{2\beta\rho}{\delta(B+\beta\rho)}-L=\Theta\left(\frac{1}{\varepsilon^{4}}\right)$
and thus $s_{0}=\Theta(1)$, $R_{0}=\Theta(1)$.
By the proof of Theorem~1 in \citet{sampling-can-be-faster}, 
$\text{TV}(\nu_{K},\pi_{\delta})\leq\tilde{\mathcal{O}}(\varepsilon)$
provided that
\begin{equation*}
\eta=\mathcal{O}\left(\frac{\rho_{\ast}}{L_{\delta}^{2}}\frac{\varepsilon^{2}}{d}\right)
=\mathcal{O}\left(\frac{\varepsilon^{10}}{d}\right),
\quad
\text{and} 
\quad
K=\tilde{O}\left(\frac{1}{\rho_{\ast}\eta}\right)
=\tilde{O}\left(\frac{L_{\delta}^{2}d}{\rho_{\ast}^{2}\varepsilon^{2}}\right)
=\tilde{\mathcal{O}}\left(\frac{d}{\varepsilon^{10}}\right).
\end{equation*}
This completes the proof
when $h$ is $\beta$-strongly convex.

Indeed, we can see that the leading-order term for $K$ we derived above
does not depend on $\beta$. However, we can also spell out the dependence on $\beta$ through
the second-order term as follows. 
Notice that, by taking into account $\beta$, we have $\mu_{\delta}=\Theta\left(\frac{\beta}{\varepsilon^{4}}\right)$ and 
thus $s_{0}=\Theta(1)$ and $R_{0}=\Theta(1)+\Theta\left(\frac{\varepsilon^{4}}{\beta}\right)$.
Then, we have $\text{TV}(\nu_{K},\pi_{\delta})\leq\tilde{\mathcal{O}}(\varepsilon)$
provided that
\begin{equation*}
\eta=\mathcal{O}\left(\frac{\rho_{\ast}}{L_{\delta}^{2}}\frac{\varepsilon^{2}}{d}\right),
\quad
\text{and} 
\quad
K=\tilde{O}\left(\frac{1}{\rho_{\ast}\eta}\right)
=\tilde{O}\left(\frac{L_{\delta}^{2}d}{\rho_{\ast}^{2}\varepsilon^{2}}\right)
=\tilde{\mathcal{O}}\left(\frac{d}{\varepsilon^{10}}\right)+\tilde{\mathcal{O}}\left(\frac{d}{\beta\varepsilon^{6}}\right),
\end{equation*}
where we ignored the dependence on the other constants $B,m,L,\rho$ when
we consider the second-order dependence on $\beta$.

Next, we consider the case when $h$
is merely convex so that 
\begin{equation*}
h^{\alpha}(x):=h(x)+\frac{\alpha}{2}\Vert x\Vert^{2} 
\end{equation*}
is $\alpha$-strongly convex.
By the previous discussions, 
$\text{TV}(\nu_{K},\pi_{\delta}^{\alpha})\leq\tilde{\mathcal{O}}(\varepsilon)$
provided that
\begin{equation*}
\eta=\mathcal{O}\left(\frac{\rho_{\ast}}{L_{\delta}^{2}}\frac{\varepsilon^{2}}{d}\right),
\quad\text{and}\quad 
K=\tilde{O}\left(\frac{1}{\rho_{\ast}\eta}\right)
=\tilde{O}\left(\frac{L_{\delta}^{2}d}{\rho_{\ast}^{2}\varepsilon^{2}}\right),
\end{equation*}
where we can take $\rho_{\ast}=s_{0}e^{-R_{0}}$.
Next, we can compute that
\begin{align*}
D(\pi^{\alpha}\Vert\pi)
&=\int_{\mathbb{R}^{d}}\log\left(\frac{\pi^{\alpha}(x)}{\pi(x)}\right)\pi^{\alpha}(x)dx
=\int_{\mathcal{C}^{\alpha}}\log\left(\frac{\int_{\mathcal{C}}e^{-f(x)}dx}{\int_{\mathcal{C}^{\alpha}}e^{-f(x)}dx}\right)
\frac{e^{-f(x)}}{\int_{\mathcal{C}^{\alpha}}e^{-f(y)}dy}dx
\\
&=\log\left(\frac{\int_{\mathcal{C}}e^{-f(x)}dx}{\int_{\mathcal{C}^{\alpha}}e^{-f(x)}dx}\right)
=\log\left(1+\frac{\int_{\mathcal{C}\backslash\mathcal{C}^{\alpha}}e^{-f(x)}dx}{\int_{\mathcal{C}^{\alpha}}e^{-f(x)}dx}\right)
\\
&\leq
\frac{\int_{\mathcal{C}\backslash\mathcal{C}^{\alpha}}e^{-f(x)}dx}{\int_{\mathcal{C}^{\alpha}}e^{-f(x)}dx}
\leq
e^{\sup_{x\in\mathcal{C}}f(x)-\inf_{x\in\mathcal{C}}f(x)}\frac{|\mathcal{C}\backslash\mathcal{C}^{\alpha}|}{|\mathcal{C}^{\alpha}|},
\end{align*}
where $\sup_{x\in\mathcal{C}}f(x)-\inf_{x\in\mathcal{C}}f(x)$ is finite since $\mathcal{C}$ is compact. 
We recall from Lemma~\ref{lemma:O:alpha} that
$\frac{|\mathcal{C}\backslash\mathcal{C}^{\alpha}|}{|\mathcal{C}^{\alpha}|}
\leq\mathcal{O}(\alpha)$,
as $\alpha\rightarrow 0$.
Finally, by Pinsker's inequality, 
\begin{equation*}
\text{TV}(\pi^{\alpha},\pi)\leq\sqrt{\frac{1}{2}D(\pi^{\alpha}\Vert\pi)}
\leq\mathcal{O}(\sqrt{\alpha}),
\end{equation*}
as $\alpha\rightarrow 0$.
Therefore $\text{TV}(\pi^{\alpha},\pi)\leq\mathcal{O}(\varepsilon)$
provided that $\alpha=\varepsilon^{2}$. 
We recall from Lemma~\ref{lem:f:plus:S:dissipative}
that 
$s_{0}=\min(m,\mu_{\delta}/2)$,
and
$R_{0}=2(R+\rho)^{2}\left(\frac{m+L}{2}+\frac{(m+L)^{2}}{\mu_{\delta}}\right)$, 
so that $\mu_{\delta}=\frac{2\alpha\rho}{\delta(B+\alpha\rho)}-L
=\Theta\left(\frac{1}{\varepsilon^{2}}\right)$
with the choice of $\alpha=\varepsilon^{2}$
and $\delta=\varepsilon^{4}$. 
Hence, we conclude that 
$\text{TV}(\nu_{K},\pi)\leq\tilde{\mathcal{O}}(\varepsilon)$
provided that $\delta=\varepsilon^{4}$, $\alpha=\varepsilon^{2}$ and 
\begin{equation*}
\eta=\mathcal{O}\left(\frac{\rho_{\ast}}{L_{\delta}^{2}}\frac{\varepsilon^{2}}{d}\right)
=\mathcal{O}\left(\frac{\varepsilon^{10}}{d}\right),
\quad\text{and}\quad 
K=\tilde{O}\left(\frac{1}{\rho_{\ast}\eta}\right)
=\tilde{O}\left(\frac{L_{\delta}^{2}d}{\rho_{\ast}^{2}\varepsilon^{2}}\right)
=\tilde{\mathcal{O}}\left(\frac{d}{\varepsilon^{10}}\right).
\end{equation*}
This completes the proof.
\hfill $\Box$

\subsection*{Proof of Lemma~\ref{lemma:Frobenius-lipschitz}}

Since $\mathcal{C}$ is a convex set, every point in $\mathbb{R}^d$ has an unique projection on $\mathcal{C}$, which leads to $reach(\mathcal{C})=\infty$, 
where
\begin{equation}
reach(\mathcal{C}):=\sup\left\{\zeta\in [0,\infty]: \text{every point in }\mathcal{C}^{\zeta} \text{ has unique projection on }\mathcal{C}\right\},
\end{equation}
with $\mathcal{C}^{\zeta}:= \{x\in\mathbb{R}^d: \inf\{\|x-\xi\|:\xi\in\mathcal{C}\}<\zeta\}$. According to Corollary~4 in \citet{leobacher2021existence}, we can get that $D^2\mathcal{P}_{\mathcal{C}}$ is bounded on $\mathbb{R}^d$, where $\mathcal{P}_{\mathcal{C}}$ is the projection operator on $\mathcal{C}$. Then there exists some constant $M_{\mathcal{P}}>0$ such that $\|D^2\mathcal{P}_{\mathcal{C}}\|_F \leq M_{\mathcal{P}}$, where $\|\cdot\|_F$ is the Frobenius norm.
Moreover, we can compute that $S(x)=(x-\mathcal{P}_{\mathcal{C}}(x))^2$, $\nabla S(x)=2(x-\mathcal{P}_{\mathcal{C}}(x))$ and $\nabla^2 S(x) = 2(I - D\mathcal{P}_{\mathcal{C}}(x))$. Note that for $x,y\in \mathbb{R}^d$,
\begin{equation*}
    \left\|\nabla^2S(x)-\nabla^2S(y)\right\|_F = 2\left\|D\mathcal{P}_{\mathcal{C}}(x) - D\mathcal{P}_{\mathcal{C}}(y)\right\|_F \leq 2 M_{\mathcal{P}} \|x-y\|.
\end{equation*}
The proof is complete.
\hfill $\Box$

\subsection*{Proof of Corollary~\ref{cor:Frobenius-lipschitz}}

The result follows from Lemma~\ref{lemma:Frobenius-lipschitz} immediately.
\hfill $\Box$


\subsection*{Proof of Proposition~\ref{cor:HMC:convex:outside}}

We first consider the case that $\mathcal{C}=\{x:h(x)\leq 0\}$, 
where $h(x)$ is $\beta$-strongly convex.
First of all, by running the penalized underdamped Langevin Monte Carlo \eqref{HMC:nonconvex:1}-\eqref{HMC:nonconvex:2}, 
in total variation distance (TV), we have
\begin{equation*}
\text{TV}(\nu_{K},\pi)\leq \text{TV}(\nu_{k},\pi_{\delta})+\text{TV}(\pi_{\delta},\pi).
\end{equation*}
We recall from \eqref{D:ineq} that the KL divergence between $\pi$ and $\pi_{\delta}$ can be bounded as:
$D(\pi\Vert\pi_{\delta})
\leq\mathcal{O}\left(\left(\delta\log(1/\delta)\right)^{1/2}\right)$.
By Pinsker's inequality, we have
\begin{equation*}
\text{TV}(\pi_{\delta},\pi)
\leq\sqrt{\frac{1}{2}D(\pi\Vert\pi_{\delta})}
\leq\mathcal{O}\left(\left(\delta\log(1/\delta)\right)^{1/4}\right).
\end{equation*}
Therefore, $\text{TV}(\pi_{\delta},\pi)\leq\tilde{\mathcal{O}}(\varepsilon)$
provided that $\delta=\varepsilon^{4}$.
Moreover, by Corollary~\ref{cor:strongly:convex}, 
$f+S/\delta$ is $\mu_{\delta}$ strongly convex outside an Euclidean ball with radius $R+\rho$ with $\mu_{\delta}:=\frac{2\beta \rho}{\delta(B+\beta\rho)}-L$
and by Lemma~\ref{lem:f:plus:S:dissipative}, $f+S/\delta$ is $L_{\delta}$-smooth with $L_{\delta}:=L+\frac{\ell}{\delta}$.
By Theorem~1 in \citet{Ma2019} and Pinsker's inequality, we have
\begin{equation*}
\text{TV}(\nu_{K},\pi_{\delta})
\leq\sqrt{\frac{1}{2}D(\nu_{K}\Vert\pi_{\delta})}
\leq\tilde{\mathcal{O}}(\varepsilon),
\end{equation*}
provided that
\begin{equation}
K=\tilde{\mathcal{O}}\left(\max\left\{\frac{L_{\delta}^{3/2}}{\hat{\mu}_{\ast}^{2}},\frac{M_{\delta}}{\hat{\mu}_{\ast}^{2}}\right\}\frac{\sqrt{d}}{\varepsilon}\right),
\end{equation}
where $\hat{\mu}_{\ast}=\min\left\{\rho_{\ast},1\right\}$, where $\rho_{\ast}$ is the log-Sobolev constant
for $\pi_{\delta}$. 
Note that in Lemma~\ref{lem:close-piecewise-quadratic}, we showed that there exists a $C^{1}$ function $U$
that is $s_{0}$-strongly convex and satisfies:
\begin{equation*}
\sup_{x\in\mathbb{R}^{d}}\left(U(x)-f(x)-\frac{S(x)}{\delta}\right)-\inf_{x\in\mathbb{R}^{d}}\left(U(x)-f(x)-\frac{S(x)}{\delta}\right)\leq R_{0}.
\end{equation*}
Therefore, by Holley-Stroock perturbation principle (see \citet{holley-stroock}), the log-Sobolev constant for $\pi_{\delta}$ can be lower bounded as $\rho_{\ast}\geq s_{0}e^{-R_{0}}$
where we recall from Lemma~\ref{lem:f:plus:S:dissipative}
that $s_{0}=\min(m,\mu_{\delta}/2)$
and $R_{0}:=2(R+\rho)^{2}\left(\frac{m+L}{2}+\frac{(m+L)^{2}}{\mu_{\delta}}\right)$,
so that we can take
$\hat{\mu}_{\ast}=\min\left\{s_{0}e^{-R_{0}},1\right\}$.
Finally, we notice that $L_{\delta}=L+\frac{\ell}{\delta}=\mathcal{O}\left(\frac{1}{\varepsilon^{4}}\right)$
and $M_{\delta}=M_{f}+\frac{M_{S}}{\delta}=\mathcal{O}\left(\frac{1}{\varepsilon^{4}}\right)$ with the choice
$\delta=\varepsilon^{4}$. 
Moreover, $\mu_{\delta}=\frac{2\beta\rho}{\delta(B+\beta\rho)}-L=\mathcal{O}\left(\frac{1}{\varepsilon^{4}}\right)$ so that $s_{0}=\Theta(1)$ and $R_{0}=\Theta(1)$ and thus $\hat{\mu}_{\ast}=\Theta(1)$. 
Hence, we conclude that
$\text{TV}(\nu_{K},\pi)\leq\tilde{\mathcal{O}}(\varepsilon)$
provided that 
$K=\tilde{\mathcal{O}}\left(\frac{\sqrt{d}}{\varepsilon^{7}}\right)$. 

Indeed, we can see that the leading-order term for $K$ we derived above
does not depend on $\beta$. However, we can also spell out the dependence on $\beta$ through
the second-order term as follows. 
Notice that, by taking into account $\beta$, we have $\mu_{\delta}=\Theta\left(\frac{\beta}{\varepsilon^{4}}\right)$ and 
thus $s_{0}=\Theta(1)$ and $R_{0}=\Theta(1)+\Theta\left(\frac{\varepsilon^{4}}{\beta}\right)$.
Then, we have $\text{TV}(\nu_{K},\pi_{\delta})\leq\tilde{\mathcal{O}}(\varepsilon)$
provided that
$K=\tilde{\mathcal{O}}\left(\frac{\sqrt{d}}{\varepsilon^{7}}\right)+\tilde{\mathcal{O}}\left(\frac{\sqrt{d}}{\beta\varepsilon^{3}}\right)$,
where we ignored the dependence on the other constants $B,m,L,\rho$ when
we consider the second-order dependence on $\beta$.

Next, we consider the case when $h$
is merely convex so that 
\begin{equation*}
h^{\alpha}(x):=h(x)+\frac{\alpha}{2}\Vert x\Vert^{2} 
\end{equation*}
is $\alpha$-strongly convex and consider the constraint set
\begin{equation*}
\mathcal{C}^{\alpha}:=\left\{x:h(x)+\frac{\alpha}{2}\Vert x\Vert^{2}\leq 0\right\}.
\end{equation*}
In the previous discussions,
we showed that $\text{TV}(\nu_{K},\pi_{\delta}^{\alpha})\leq\tilde{\mathcal{O}}(\varepsilon)$
provided that
$K=\tilde{\mathcal{O}}\left(\max\left\{\frac{L_{\delta}^{3/2}}{\hat{\mu}_{\ast}^{2}},\frac{M_{\delta}}{\hat{\mu}_{\ast}^{2}}\right\}\frac{\sqrt{d}}{\varepsilon}\right)$,
where we can take $\hat{\mu}_{\ast}=\min\left\{s_{0}e^{-R_{0}},1\right\}$.
By following the proof of Proposition~\ref{cor:LD:convex:outside}, we have
$\text{TV}(\pi^{\alpha},\pi)\leq\mathcal{O}(\varepsilon)$
provided that $\alpha=\varepsilon^{2}$. 
We recall from Lemma~\ref{lem:f:plus:S:dissipative}
that 
$s_{0}=\min(m,\mu_{\delta}/2)$
and
$R_{0}=2(R+\rho)^{2}\left(\frac{m+L}{2}+\frac{(m+L)^{2}}{\mu_{\delta}}\right)$, 
so that $\mu_{\delta}=\frac{2\alpha\rho}{\delta(B+\alpha\rho)}-L
=\Theta\left(\frac{1}{\varepsilon^{2}}\right)$
with the choice of $\alpha=\varepsilon^{2}$
and $\delta=\varepsilon^{4}$
so that $s_{0}=\Theta(1)$ and $R_{0}=\Theta(1)$
and $\hat{\mu}_{\ast}=\Theta(1)$. 
Finally, we notice that $L_{\delta}=L+\frac{\ell}{\delta}=\mathcal{O}\left(\frac{1}{\varepsilon^{4}}\right)$
and $M_{\delta}=M_{f}+\frac{M_{S}}{\delta}=\mathcal{O}\left(\frac{1}{\varepsilon^{4}}\right)$ with the choice
$\delta=\varepsilon^{4}$. 
Hence, we conclude that 
$\text{TV}(\nu_{K},\pi)\leq\tilde{\mathcal{O}}(\varepsilon)$
provided that $\delta=\varepsilon^{4}$ and $\alpha=\varepsilon^{2}$
and 
$K=\tilde{\mathcal{O}}\left(\frac{\sqrt{d}}{\varepsilon^{7}}\right)$. 
This completes the proof.
\hfill $\Box$

\subsection*{Proof of Lemma~\ref{lem-str-cvx-implies-lower-bound}}

Since $f$ is strongly convex, it admits a unique minimizer, say $x_{\ast,f}$. 
If $x_{\ast,f}\in\mathcal{C}$, then for any $x\notin\mathcal{C}$, $S(x)>0$
and 
\begin{equation*}
f(x)+\frac{S(x)}{\delta}>f(x_{\ast,f})+\frac{S(x_{\ast,f})}{\delta}=f(x_{\ast,f}),
\end{equation*}
which implies the minimizer of $f+\frac{S}{\delta}$ must lie within $\mathcal{C}$ and hence
the conclusion follows.
If $x_{\ast,f}\notin\mathcal{C}$, then $S(x_{\ast,f})=(\delta_{\mathcal{C}}(x_{\ast,f}))^{2}>0$. 
Then, for any $x$ such that $S(x)>S(x_{\ast,f})$, 
we have 
\begin{equation*}
f(x)+\frac{S(x)}{\delta}>f(x_{\ast,f})+\frac{S(x_{\ast,f})}{\delta}=f(x_{\ast,f}),
\end{equation*}
which implies that any minimizer $x_{\ast}$ of $f+\frac{S}{\delta}$
must satisfy $S(x_{\ast})\leq S(x_{\ast,f})$ so that $\delta_{\mathcal{C}}(x_{\ast})\leq\delta_{\mathcal{C}}(x_{\ast,f})$.
Since $\mathcal{C}$ is contained in a Euclidean ball centered at $0$ with radius $R>0$, 
we conclude that $\Vert x_{\ast}\Vert\leq R+\delta_{\mathcal{C}}(x_{\ast,f})$, which is
independent of $\delta$.
This completes the proof.
\hfill $\Box$



\subsection*{Proof of Proposition~\ref{prop:SGLD:convex}}
First of all, we notice that
with $S(x)=\left(\delta_{\mathcal{C}}(x)\right)^{2}$, by Lemma~\ref{lem:S}, 
$S$ is convex, $\ell$-smooth (with $\ell=4$) and continuously differentiable.

We will first show that we can uniformly bound the variance
of the gradient noise.
Let $x_{\ast}$ be the unique minimizer of $f(x)+\frac{1}{\delta}S(x)$
(the minimizer is unique since $f(x)+\frac{1}{\delta}S(x)$ is strongly convex by Assumption~\ref{assump:f:1} and Lemma~\ref{lem:S}). 
By Lemma~\ref{lem-str-cvx-implies-lower-bound}, $\Vert x_{\ast}\Vert\leq(1+c)R$
for some $c,R\geq 0$.
This implies that for any $\frac{\eta L_{\delta}}{2}<1$, 
\begin{align*}
&\mathbb{E}\Vert x_{k+1}-x_{\ast}\Vert^{2}
\\
&=\mathbb{E}\left\Vert x_{k}-x_{\ast}-\eta\left(\nabla f(x_{k})+\frac{1}{\delta}\nabla S(x_{k})\right)\right\Vert^{2}
+\eta^{2}\mathbb{E}\left\Vert\nabla\tilde{f}(x_{k})-\nabla f(x_{k})\right\Vert^{2}
+\mathbb{E}\left\Vert\sqrt{2\eta}\xi_{k+1}\right\Vert^{2}
\\
&\leq
\mathbb{E}\left\Vert x_{k}-x_{\ast}\right\Vert^{2}
-2\eta\mathbb{E}\left\langle x_{k}-x_{\ast},\nabla f(x_{k})+\frac{1}{\delta}\nabla S(x_{k})\right\rangle
+\eta^{2}\mathbb{E}\left\Vert\nabla f(x_{k})+\frac{1}{\delta}\nabla S(x_{k})\right\Vert^{2}
\\
&\qquad\qquad
+2\eta^{2}\sigma^{2}\left(L^{2}\mathbb{E}\Vert x_{k}\Vert^{2}+\Vert\nabla f(0)\Vert^{2}\right)
+2\eta d
\\
&\leq
\mathbb{E}\left\Vert x_{k}-x_{\ast}\right\Vert^{2}
-2\eta\left(1-\frac{\eta L_{\delta}}{2}\right)\mathbb{E}\left\langle x_{k}-x_{\ast},\nabla f(x_{k})+\frac{1}{\delta}\nabla S(x_{k})\right\rangle
\\
&\qquad\qquad
+2\eta^{2}\sigma^{2}\left(L^{2}\mathbb{E}\Vert x_{k}\Vert^{2}+\Vert\nabla f(0)\Vert^{2}\right)
+2\eta d
\\
&\leq
(1-2\eta\mu+\eta^{2}\mu L_{\delta})\mathbb{E}\left\Vert x_{k}-x_{\ast}\right\Vert^{2}
+2\eta^{2}\sigma^{2}\left(L^{2}\mathbb{E}\Vert x_{k}\Vert^{2}+\Vert\nabla f(0)\Vert^{2}\right)
+2\eta d
\\
&\leq
(1-2\eta\mu+\eta^{2}\mu L_{\delta})\mathbb{E}\left\Vert x_{k}-x_{\ast}\right\Vert^{2}
\\
&\qquad\qquad
+2\eta^{2}\sigma^{2}\left(2L^{2}\mathbb{E}\Vert x_{k}-x_{\ast}\Vert^{2}+2L^{2}(1+c)^{2}R^{2}+\Vert\nabla f(0)\Vert^{2}\right)
+2\eta d,
\end{align*}
where we used $\frac{\eta L_{\delta}}{2}<1$, and the fact that $f+\frac{1}{\delta}S$ is $\mu$-strongly convex
and $L_{\delta}$-smooth.
Hence, for any $\eta\leq\frac{\mu}{\mu L_{\delta}+4\sigma^{2}L^{2}}$ and $\frac{\eta L_{\delta}}{2}<1$, we get
\begin{equation*}
\mathbb{E}\Vert x_{k+1}-x_{\ast}\Vert^{2}
\leq(1-\eta\mu)\mathbb{E}\left\Vert x_{k}-x_{\ast}\right\Vert^{2}
+2\eta^{2}\sigma^{2}\left(2L^{2}(1+c)^{2}R^{2}+\Vert\nabla f(0)\Vert^{2}\right)
+2\eta d,
\end{equation*}
which implies that
\begin{align}
\mathbb{E}\Vert x_{k}\Vert^{2}
&\leq
2\mathbb{E}\Vert x_{k}-x_{\ast}\Vert^{2}
+2(1+c)^{2}R^{2}
\nonumber
\\
&
\leq
\frac{4\eta\sigma^{2}}{\mu}\left(2L^{2}(1+c)^{2}R^{2}+\Vert\nabla f(0)\Vert^{2}\right)
+\frac{4d}{\mu}+2(1+c)^{2}R^{2}.
\end{align}
Hence, we conclude that
\begin{align}
\mathbb{E}\left\Vert\nabla\tilde{f}(x_{k})-\nabla f(x_{k})\right\Vert^{2}
\leq
2\sigma^{2}\left(L^{2}\mathbb{E}\Vert x_{k}\Vert^{2}+\Vert\nabla f(0)\Vert^{2}\right)
\leq\sigma_{V}^{2}d,\label{grad:noise:assumption:friendly}
\end{align}
where
\begin{equation}\label{sigma:V}
\sigma_{V}^{2}:=\sigma^{2}\left(\frac{8\eta\sigma^{2}L^{2}}{\mu d}\left(2L^{2}(1+c)^{2}R^{2}+\Vert\nabla f(0)\Vert^{2}\right)
+\frac{8L^{2}}{\mu}+\frac{4L^{2}(1+c)^{2}R^{2}}{d}
+\frac{2\Vert\nabla f(0)\Vert^{2}}{d}\right).
\end{equation}

Let $\nu_{K}$ be the distribution
of the $K$-th iterate of the penalized stochastic gradient Langevin dynamics
given by \eqref{SGLD:convex}.
By applying Theorem~4 in \citet{DK2017}, 
under the assumption that $f(x)+\frac{1}{\delta}S(x)$ is $\mu$-strongly convex and $L_{\delta}$-smooth and
the variance of the gradient noise is uniformly bounded (i.e., \eqref{grad:noise:assumption:friendly}) 
and the stepsize satisfies
$\eta\leq\frac{\mu}{\mu L_{\delta}+4\sigma^{2}L^{2}}$
and $\eta<\min\left(\frac{\mu}{\mu L_{\delta}+4\sigma^{2}L^{2}},\frac{2}{L_{\delta}}\right)$ (so that \eqref{grad:noise:assumption:friendly} holds), we have
\begin{equation}
\mathcal{W}_{2}(\nu_{K},\pi_{\delta})
\leq(1-\mu\eta)^{K}\mathcal{W}_{2}(\nu_{0},\pi_{\delta})
+\frac{1.65L_{\delta}}{\mu}\sqrt{\eta d}
+\frac{\sigma_{V}^{2}\sqrt{\eta d}}{1.65L_{\delta}+\sigma_{V}\sqrt{\mu}},
\end{equation}
where $\sigma_{V}$ is defined in \eqref{sigma:V}, 
so that together with Theorem~\ref{thm:final} we have
\begin{equation}
\mathcal{W}_{2}(\nu_{K},\pi)
\leq(1-\mu\eta)^{K}\mathcal{W}_{2}(\nu_{0},\pi_{\delta})
+\frac{1.65L_{\delta}}{\mu}\sqrt{\eta d}
+\frac{\sigma_{V}^{2}\sqrt{\eta d}}{1.65L_{\delta}+\sigma_{V}\sqrt{\mu}}
+\mathcal{O}\left(\left(\delta\log(1/\delta)\right)^{1/8}\right).
\end{equation}

Moreover, we can compute that
\begin{equation*}
\mathcal{W}_{2}(\nu_{0},\pi_{\delta})
\leq\left(\mathbb{E}_{X\sim\nu_{0}}\Vert X\Vert^{2}\right)^{1/2}
+\left(\mathbb{E}_{X\sim\pi_{\delta}}\Vert X\Vert^{2}\right)^{1/2},
\end{equation*}
and by the definition of $\pi_{\delta}$,
\begin{align}\label{eqn:upper:bound:indep}
\mathbb{E}_{X\sim\pi_{\delta}}\Vert X\Vert^{2}
=\frac{\int_{\mathbb{R}^{d}}\Vert x\Vert^{2}e^{-f(x)-\frac{S(x)}{\delta}}dx}{\int_{\mathbb{R}^{d}}e^{-f(x)-\frac{S(x)}{\delta}}dx}
\leq
\frac{\int_{\mathbb{R}^{d}}\Vert x\Vert^{2}e^{-f(x)}dx}{\int_{\mathcal{C}}e^{-f(x)}dx},
\end{align}
where the upper bound in \eqref{eqn:upper:bound:indep} 
is finite and independent of $\delta$ since $f$ is $\mu$-strongly convex.

By taking $\delta=\varepsilon^{8}$, 
$\eta=\frac{\varepsilon^{18}\mu^{2}}{d(L\varepsilon^{8}+\ell)^{2}}$,
and
$K=\tilde{\bigO}\left(\frac{d(L\varepsilon^{8}+\ell)^{2}}{\varepsilon^{18}\mu^{3}}\right)$,
we get
\begin{align*}
\mathcal{W}_{2}(\nu_{K},\pi)
&\leq
\tilde{\mathcal{O}}(\varepsilon)
+\frac{\sigma_{V}^{2}\sqrt{\eta d}}{1.65L_{\delta}+\sigma_{V}\sqrt{\mu}}
\\
&\leq
\tilde{\mathcal{O}}(\varepsilon)
+\tilde{\mathcal{O}}\left(\frac{\sigma_{V}^{2}\sqrt{\eta d}\varepsilon^{8}}{L\varepsilon^{8}+\ell}\right)
\leq
\tilde{\mathcal{O}}(\varepsilon)
+\tilde{\mathcal{O}}\left(\frac{\sigma_{V}^{2}\varepsilon^{17}\mu}{(L\varepsilon^{8}+\ell)^{2}}\right).
\end{align*}
Therefore $\mathcal{W}_{2}(\nu_{K},\pi)\leq\tilde{\mathcal{O}}(\varepsilon)$
provided that
$\sigma_{V}^{2}=\tilde{\mathcal{O}}\left(\frac{(L\varepsilon^{8}+\ell)^{2}}{\varepsilon^{16}\mu}\right)$.
This implies that $\sigma_{V}^{2}$ and hence $\sigma^{2}$ 
and the batch-size $b$ can simply be taken as the constant order,
and therefore the stochastic gradient computations satisfy:
$\hat{K}:=Kb=\tilde{\bigO}\left(\frac{d(L\varepsilon^{8}+\ell)^{2}}{\varepsilon^{18}\mu^{3}}\right)$.
Finally, by Lemma~\ref{lem:S}, we can take $\ell=4$.
The proof is complete.
\hfill $\Box$



\subsection*{Proof of Proposition~\ref{prop:SGHMC:convex}}

Before we proceed to the technical proof of Proposition~\ref{prop:SGHMC:convex}, we make the following remark regarding Lemma~\ref{lem-dissipative-implies-lower-bound}.

\begin{rema}\label{remark:f:0}
Note that in Lemma~\ref{lem-dissipative-implies-lower-bound} without loss of generality, we can always assume $M=0$ so that $f+\frac{S}{\delta}\geq 0$. This is because, if $M>0$, we can always consider the ``shifted" function $\hat{f}:= f+ M$ which will satisfy $\hat{f}\geq 0$ and then apply the proof arguments to 
$e^{-\hat{f}(x)-\frac{S(x)}{\delta}}/\int_{x\in\mathcal{C}} e^{-\hat{f}(x)-\frac{S(x)}{\delta}}dx$ which will be proportional to $e^{-f(x)-\frac{S(x)}{\delta}}$. Therefore, in the rest of the paper and the proofs, we will assume $M=0$ in Lemma~\ref{lem-dissipative-implies-lower-bound}.
\end{rema}

Now, we are ready to present the technical proof of Proposition~\ref{prop:SGHMC:convex}.

First of all, we notice that
with $S(x)=\left(\delta_{\mathcal{C}}(x)\right)^{2}$,
by Lemma~\ref{lem:S}, $S$ is convex, $\ell$-smooth and continuously differentiable.
One technical challenge is that 
we cannot apply the results directly from \citet{dalalyan2018kinetic}
because the results in \citet{dalalyan2018kinetic} are for
the underdamped Langevin Monte Carlo without the gradient noise. 
Therefore, we need to adapt their approach
to allow the additional gradient noise.
First, we will obtain uniform $L^{2}$ bounds
on penalized SGULMC $v_{k}$ and $x_{k}$ in \eqref{SGHMC:convex:1}--\eqref{SGHMC:convex:2}.

Under Assumption~\ref{assump:f:1} and by Lemma~\ref{lem:S}, 
$f+\frac{S}{\delta}$ is $\mu$-strongly convex so that
we have
\begin{equation}\label{lower:f:S}
\left\langle\nabla f(x)+\frac{1}{\delta}\nabla S(x),x-x_{\ast}\right\rangle\geq\mu\Vert x-x_{\ast}\Vert^{2},
\end{equation}
where $x_{\ast}$ is the unique minimizer of $f+\frac{1}{\delta}S$.
By Lemma~\ref{lem-str-cvx-implies-lower-bound}, $\Vert x_{\ast}\Vert\leq(1+c)R$
for some $c,R\geq 0$.
On the other hand, 
\begin{equation}
\left|\left\langle\nabla f(x)+\frac{1}{\delta}\nabla S(x),x_{\ast}\right\rangle\right|
\leq L_{\delta}\Vert x_{\ast}\Vert\cdot\Vert x-x_{\ast}\Vert 
\leq L_{\delta}(1+c)R\Vert x-x_{\ast}\Vert, 
\end{equation}
which together with \eqref{lower:f:S} implies that
\begin{align*}
\left\langle\nabla f(x)+\frac{1}{\delta}\nabla S(x),x\right\rangle
&\geq\mu\Vert x-x_{\ast}\Vert^{2}
-L_{\delta}(1+c)R\Vert x-x_{\ast}\Vert
\\
&\geq
\frac{\mu}{2}\Vert x-x_{\ast}\Vert^{2}-\frac{L_{\delta}^{2}(1+c)^{2}R^{2}}{2\mu}
\\
&\geq
\frac{\mu}{4}\Vert x\Vert^{2}-\frac{\mu}{2}\Vert x_{\ast}\Vert^{2}-\frac{L_{\delta}^{2}(1+c)^{2}R^{2}}{2\mu}
\\
&\geq
\frac{\mu}{4}\Vert x\Vert^{2}-\frac{\mu}{2}(1+c)^{2}R^{2}-\frac{L_{\delta}^{2}(1+c)^{2}R^{2}}{2\mu},
\end{align*}
and therefore $f+\frac{1}{\delta}S$ is $(m_{0},b_{0})$-dissipative 
with
\begin{equation}\label{m:b:eqn}
m_{0}:=\frac{\mu}{4},\qquad b_{0}:=\frac{\mu}{2}(1+c)^{2}R^{2}+\frac{L_{\delta}^{2}(1+c)^{2}R^{2}}{2\mu},
\end{equation}
and moreover 
by Lemma~\ref{lem-dissipative-implies-lower-bound} and Remark~\ref{remark:f:0},
$f+\frac{1}{\delta}S\geq 0$, 
and it follows from Lemma~EC.5 in \citet{GGZ} that uniformly in $k$, we have
\begin{align}
&\mathbb{E}\Vert x_{k}\Vert^{2}\leq C_{x}^{d}
:=\frac{\int_{\mathbb{R}^{2d}}\mathcal{V}(x,v)\mu_{0}(dx,dv)
+\frac{4(d+A)}{\lambda}}{\frac{1}{8} (1-2 \lambda)  \gamma^2 },\nonumber
\\
&\mathbb{E}\Vert v_{k}\Vert^{2}\leq C_{v}^{d}
:=\frac{\int_{\mathbb{R}^{2d}}\mathcal{V}(x,v)\mu_{0}(dx,dv)
+\frac{4(d+A)}{\lambda}}{\frac{1}{4}(1-2\lambda)},\label{def-Cx-d-strongly-convex}
\end{align}
where
\begin{align}
&\lambda:=\frac{1}{2}\min(1/4,m_{0}/(L_{\delta}+\gamma^{2}/2)),\label{lambda:eqn:recall}
\\
&A:=\frac{m_{0}}{2L_{\delta}+\gamma^{2}}\left(\frac{\Vert\nabla f(0)\Vert^{2}}{2L_{\delta}+\gamma^{2}}+\frac{b_{0}}{m_{0}}\left(L_{\delta}+\frac{1}{2}\gamma^{2}\right)+f(0)\right),\label{A:eqn:recall}
\end{align}
and $\mu_{0}$ is the distribution of $(x_{0},v_{0})$ and
\begin{equation}
\mathcal{V}(x,v):=f(x)+\frac{S(x)}{\delta}
+\frac{1}{4}\gamma^{2}\left(\left\Vert x+\gamma^{-1}v\right\Vert^{2}+\left\Vert\gamma^{-1}v\right\Vert^{2}-\lambda\Vert x\Vert^{2}\right)\,.
\end{equation}

Next, we will bound the difference
between $v_{k+1},x_{k+1}$ of the penalized SGULMC 
and $\tilde{v}_{k+1},\tilde{x}_{k+1}$, which
are the penalized ULMC without gradient noise
that also start
from $v_{k},x_{k}$ of the penalized SGULMC at $k$-th iterate.
We recall from \eqref{SGHMC:convex:1}-\eqref{SGHMC:convex:2} that
\begin{align}
&v_{k+1}=\psi_{0}(\eta)v_{k}-\psi_{1}(\eta)\left(\nabla\tilde{f}(x_{k})+\frac{1}{\delta}\nabla S(x_{k})\right)+\sqrt{2\gamma}\xi_{k+1},
\\
&x_{k+1}=x_{k}+\psi_{1}(\eta)v_{k}-\psi_{2}(\eta)\left(\nabla\tilde{f}(x_{k})+\frac{1}{\delta}\nabla S(x_{k})\right)+\sqrt{2\gamma}\xi'_{k+1},
\end{align}
and next, we define
\begin{align}
&\tilde{v}_{k+1}:=\psi_{0}(\eta)v_{k}-\psi_{1}(\eta)\left(\nabla f(x_{k})+\frac{1}{\delta}\nabla S(x_{k})\right)+\sqrt{2\gamma}\xi_{k+1},
\\
&\tilde{x}_{k+1}:=x_{k}+\psi_{1}(\eta)v_{k}-\psi_{2}(\eta)\left(\nabla f(x_{k})+\frac{1}{\delta}\nabla S(x_{k})\right)+\sqrt{2\gamma}\xi'_{k+1},
\end{align}
so that one can easily check that
\begin{equation}
\mathbb{E}\Vert v_{k+1}-\tilde{v}_{k+1}\Vert^{2}
\leq(\psi_{1}(\eta))^{2}2\sigma^{2}\left(L^{2}\mathbb{E}\Vert x_{k}\Vert^{2}+\Vert\nabla f(0)\Vert^{2}\right)
\leq 2\eta^{2}\sigma^{2}\left(L^{2}C_{x}^{d}+\Vert\nabla f(0)\Vert^{2}\right),\label{v:difference}
\end{equation}
and moreover,
\begin{equation}
\mathbb{E}\Vert x_{k+1}-\tilde{x}_{k+1}\Vert^{2}
\leq(\psi_{2}(\eta))^{2}2\sigma^{2}\left(L^{2}\mathbb{E}\Vert x_{k}\Vert^{2}+\Vert\nabla f(0)\Vert^{2}\right)
\leq 2\eta^{4}\sigma^{2}\left(L^{2}C_{x}^{d}+\Vert\nabla f(0)\Vert^{2}\right).\label{x:difference}
\end{equation}
Since $\tilde{v}_{k+1},\tilde{x}_{k+1}$ are the updates without the gradient noise, 
by using the synchronous coupling and following the same argument as in the proof of Theorem~2 in \citet{dalalyan2018kinetic}, one can show that

\begin{align*}
&\left(\mathbb{E}\left[\left\Vert P^{-1}
\left[
\begin{array}{c}
\tilde{v}_{k+1}-V((k+1)\eta)
\\
\tilde{x}_{k+1}-X((k+1)\eta)
\end{array}
\right]\right\Vert^{2}\right]\right)^{1/2}
\\
&\leq
\left(1-\frac{0.75\mu\eta}{\gamma}\right)\left(\mathbb{E}\left[\left\Vert P^{-1}
\left[
\begin{array}{c}
v_{k}-V(k\eta)
\\
x_{k}-X(k\eta)
\end{array}
\right]\right\Vert^{2}\right]\right)^{1/2}
+0.75L_{\delta}\eta^{2}\sqrt{d},
\end{align*}
where $(X(t),V(t))$ is the continuous-time penalized underdamped Langevin diffusion \eqref{eq:VL:penalized}-\eqref{eq:XL:penalized}
starting from the Gibbs distribution $\pi_{\delta}$ and
\begin{equation}
P:=\frac{1}{\gamma}
\left[\begin{array}{cc}
0_{d\times d} & -\gamma I_{d}
\\
I_{d} & I_{d}
\end{array}
\right].
\end{equation}
This implies that
\begin{align*}
&\left(\mathbb{E}\left[\left\Vert P^{-1}
\left[
\begin{array}{c}
v_{k+1}-V((k+1)\eta)
\\
x_{k+1}-X((k+1)\eta)
\end{array}
\right]\right\Vert^{2}\right]\right)^{1/2}
\\
&\leq
\left(1-\frac{0.75\mu\eta}{\gamma}\right)\left(\mathbb{E}\left[\left\Vert P^{-1}
\left[
\begin{array}{c}
v_{k}-V(k\eta)
\\
x_{k}-X(k\eta)
\end{array}
\right]\right\Vert^{2}\right]\right)^{1/2}
+0.75L_{\delta}\eta^{2}\sqrt{d}
\\
&\qquad\qquad\qquad\qquad\qquad
+\left(\mathbb{E}\left[\left\Vert P^{-1}
\left[
\begin{array}{c}
\tilde{v}_{k+1}-v_{k+1}
\\
\tilde{x}_{k+1}-x_{k+1}
\end{array}
\right]\right\Vert^{2}\right]\right)^{1/2},
\end{align*}
where we can compute from \eqref{v:difference} and \eqref{x:difference} that
\begin{align*}
\left(\mathbb{E}\left[\left\Vert P^{-1}
\left[
\begin{array}{c}
\tilde{v}_{k+1}-v_{k+1}
\\
\tilde{x}_{k+1}-x_{k+1}
\end{array}
\right]\right\Vert^{2}\right]\right)^{1/2}
&\leq
\left\Vert P^{-1}\right\Vert
\left(\mathbb{E}\Vert\tilde{v}_{k+1}-v_{k+1}\Vert^{2}+\mathbb{E}\Vert\tilde{x}_{k+1}-x_{k+1}\Vert^{2}\right)^{1/2}
\\
&\leq
2\eta\sigma\left\Vert P^{-1}\right\Vert
\left(L^{2}C_{x}^{d}+\Vert\nabla f(0)\Vert^{2}\right)^{1/2},
\end{align*}
provided that $\eta\leq 1$, 
which implies that
\begin{align*}
A_{k+1}
\leq
\left(1-\frac{0.75\mu\eta}{\gamma}\right)A_{k}
+0.75L_{\delta}\eta^{2}\sqrt{d}
+2\eta\sigma\left\Vert P^{-1}\right\Vert
\left(L^{2}C_{x}^{d}+\Vert\nabla f(0)\Vert^{2}\right)^{1/2},
\end{align*}
where
\begin{equation}
A_{k}:=\left(\mathbb{E}\left[\left\Vert P^{-1}
\left[
\begin{array}{c}
v_{k}-V(k\eta)
\\
x_{k}-X(k\eta)
\end{array}
\right]\right\Vert^{2}\right]\right)^{1/2}.
\end{equation}
This implies that
\begin{align*}
\mathcal{W}_{2}(\nu_{k},\pi)
&\leq\gamma^{-1}\sqrt{2}A_{k}
\\
&\leq
\frac{L_{\delta}\eta\sqrt{2d}}{\mu}
+\frac{8\sqrt{2}}{3\mu}\sigma\left\Vert P^{-1}\right\Vert
\left(L^{2}C_{x}^{d}+\Vert\nabla f(0)\Vert^{2}\right)^{1/2}
+\sqrt{2}\left(1-\frac{0.75\mu\eta}{\gamma}\right)^{k}\frac{A_{0}}{\gamma}
\\
&=\frac{L_{\delta}\eta\sqrt{2d}}{\mu}
+\frac{8\sqrt{2}}{3\mu}\sigma\left\Vert P^{-1}\right\Vert
\left(L^{2}C_{x}^{d}+\Vert\nabla f(0)\Vert^{2}\right)^{1/2}
\\
&\qquad\qquad\qquad\qquad\qquad\qquad
+\sqrt{2}\left(1-\frac{0.75\mu\eta}{\gamma}\right)^{k}\mathcal{W}_{2}(\nu_{0},\pi),
\end{align*}
where $\nu_{K}$ denotes the distribution 
of the $K$-th iterate $x_{K}$ of penalized stochastic gradient underdamped Langevin Monte Carlo \eqref{SGHMC:convex:1}-\eqref{SGHMC:convex:2}.
By the same argument as in the proof of Proposition~\ref{prop:SGLD:convex}, 
we can show that $\mathcal{W}_{2}(\nu_{0},\pi_{\delta})$
can be bounded uniformly in $\delta$.

Hence, by taking $\delta=\varepsilon^{8}$, we get
\begin{equation}
\mathcal{W}_{2}(\nu_{K},\pi)\leq\tilde{\mathcal{O}}(\varepsilon)
+\frac{8\sqrt{2}}{3\mu}\sigma\left\Vert P^{-1}\right\Vert
\left(L^{2}C_{x}^{d}+\Vert\nabla f(0)\Vert^{2}\right)^{1/2},
\end{equation}
where $\tilde{\mathcal{O}}$ ignores the dependence on $\log(1/\varepsilon)$, 
provided that
\begin{equation}
\eta=\min\left(\frac{1}{\sqrt{d}}\frac{\varepsilon^{9}\mu}{(L\varepsilon^{8}+\ell)},\frac{1}{\sqrt{(\mu+L)\varepsilon^{8}+\ell}}\frac{\varepsilon^{12}\mu}{(L\varepsilon^{8}+\ell)}\right),
\end{equation}
and
\begin{equation}
K=\tilde{\bigO}\left(\frac{\sqrt{(\mu+L)\varepsilon^{8}+\ell}(L\varepsilon^{8}+\ell)}{\varepsilon^{13}\mu^{2}}\max\left(\sqrt{d},\frac{\sqrt{(L+\mu)\varepsilon^{8}+\ell}}{\varepsilon^{3}}\right)\right),
\end{equation}
where $\tilde{\bigO}$ ignores the dependence on $\log(1/\varepsilon)$. 
Next, we recall from \eqref{def-Cx-d-strongly-convex} that
\begin{equation}\label{follow-C-x-d}
C_{x}^{d}=\frac{\int_{\mathbb{R}^{2d}}\mathcal{V}(x,v)\mu_{0}(dx,dv)
+\frac{4(d+A)}{\lambda}}{\frac{1}{8} (1-2 \lambda)  \gamma^2 }
\end{equation}
where we recall from \eqref{lambda:eqn:recall}-\eqref{A:eqn:recall} that
\begin{align}
&\lambda=\frac{1}{2}\min(1/4,m_{0}/(L_{\delta}+\gamma^{2}/2)),\label{lambda:proof}
\\
&A=\frac{m_{0}}{2L_{\delta}+\gamma^{2}}\left(\frac{\Vert\nabla f(0)\Vert^{2}}{2L_{\delta}+\gamma^{2}}+\frac{b_{0}}{m_{0}}\left(L_{\delta}+\frac{1}{2}\gamma^{2}\right)+f(0)\right),\label{A:proof}
\end{align}
where we recall from \eqref{m:b:eqn} that
$m_{0}=\frac{\mu}{4}$ and $b_{0}=\frac{\mu}{2}(1+c)^{2}R^{2}+\frac{L_{\delta}^{2}(1+c)^{2}R^{2}}{2\mu}$.
Since $L_{\delta}=L+\frac{\ell}{\delta}=L+\frac{\ell}{\varepsilon^{8}}$, 
we conclude from \eqref{lambda:proof} and \eqref{A:proof} that
$\lambda=\Omega\left(\frac{\mu\varepsilon^{8}}{\varepsilon^{8}L+\ell}\right)$
and
$A=\mathcal{O}\left(\left(L+\frac{\ell}{\varepsilon^{8}}\right)^{2}\frac{1}{\mu}\right)$,
and it follows from \eqref{follow-C-x-d} that
$C_{x}^{d}=\mathcal{O}\left(\frac{\varepsilon^{16}d\mu(L\varepsilon^{8}+\ell)+(L\varepsilon^{8}+\ell)^{3}}{\mu^{2}\varepsilon^{24}}\right)$,
which implies that $\mathcal{W}_{2}(\nu_{K},\pi)\leq\tilde{\mathcal{O}}(\varepsilon)$
provided that
$\sigma=\mathcal{O}\left(\frac{\varepsilon^{13}\mu^{2}}{L\sqrt{L\varepsilon^{8}+\ell}\sqrt{\varepsilon^{16}d\mu+(L\varepsilon^{8}+\ell)^{2}}}\right)$,
so that we can take
\begin{equation}
b=\Omega\left(\sigma^{-2}\right)
=\Omega\left(\frac{L^{2}(L\varepsilon^{8}+\ell)(\varepsilon^{16}d\mu+(L\varepsilon^{8}+\ell)^{2})}{\varepsilon^{26}\mu^{4}}\right).
\end{equation}
Hence, $\mathcal{W}_{2}(\nu_{K},\pi)\leq\tilde{\mathcal{O}}(\varepsilon)$
with stochastic gradient computations $\hat{K}:=Kb$:
\begin{align}
\hat{K}=\tilde{\bigO}
\Bigg(\frac{L^{2}(L\varepsilon^{8}+\ell)^{2}(\varepsilon^{16}d\mu+(L\varepsilon^{8}+\ell)^{2})\sqrt{(\mu+L)\varepsilon^{8}+\ell}}{\varepsilon^{39}\mu^{6}}
\max\left(\sqrt{d},\frac{\sqrt{(L+\mu)\varepsilon^{8}+\ell}}{\varepsilon^{3}}\right)\Bigg).
\end{align}
Finally, by Lemma~\ref{lem:S}, we can take $\ell=4$.
The proof is complete.
\hfill $\Box$




\subsection*{Proof of Proposition~\ref{prop:SGLD:nonconvex}}

Let $\nu_{K}$ be the distribution of the $K$-th iterate $x_{K}$ of penalized stochastic gradient Langevin dynamics \eqref{SGLD:nonconvex}.
We recall from Lemma~\ref{lem:f:plus:S:dissipative}
that $f+\frac{1}{\delta}S$ 
is $(m_{\delta},b_{\delta})$-dissipative 
with
$m_{\delta}:=-L-\frac{1}{2}+\frac{m_{S}}{\delta}$
and $b_{\delta}:=\frac{1}{2}\Vert\nabla f(0)\Vert^{2}+\frac{b_{S}}{\delta}$ from \eqref{m:delta:dissipative}
and $f+\frac{1}{\delta}S$ is $L_{\delta}$-smooth with $L_{\delta}:=L+\frac{\ell}{\delta}$
and we also recall from Lemma~\ref{lem-dissipative-implies-lower-bound} and Remark~\ref{remark:f:0}
that $f+\frac{1}{\delta}S\geq 0$.
Under Assumption~\ref{assump:f:2} and the assumption that $\eta\in(0,1\wedge\frac{m_{\delta}}{4L_{\delta}^{2}})$ and $k\eta\geq 1$,
by Proposition~10 in \citet{Raginsky}, we have
\begin{equation}\label{eqn:control}
\mathcal{W}_{2}(\nu_{K},\pi)
\leq
\left(\tilde{C}_{0}\sigma^{1/2}+\tilde{C}_{1}\eta^{1/4}\right)(K\eta)+\tilde{C}_{2}e^{-K\eta/c_{LS}}
+\mathcal{O}\left(\left(\delta\log(1/\delta)\right)^{1/8}\right),
\end{equation}
where $\tilde{C}_{1}$, $\tilde{C}_{2}$ are defined as, 
\begin{align}
&\tilde{C}_{0}:=(12+8(\kappa_{0}+2b_{\delta}+2d))\left(C_{0}+\sqrt{C_{0}}\right),
\\
&\tilde{C}_{1}:=\left(12+8(\kappa_{0}+2b_{\delta}+2d)\right)\left(6L_{\delta}^{2}(C_{0}+d)+\sqrt{6L_{\delta}^{2}(C_{0}+d)}\right),\label{eqn:tilde:C:1}
\\
&\tilde{C}_{2}:=\sqrt{2c_{LS}}
\left(\log\Vert p_{0}\Vert_{\infty}+\frac{d}{2}\log\frac{3\pi}{m_{\delta}}+\frac{L_{\delta}\kappa_{0}}{3}+\Vert\nabla f(0)\Vert\sqrt{\kappa_{0}}+f(0)+\frac{b_{\delta}}{2}\log 3\right)^{1/2},\label{eqn:tilde:C:2}
\end{align}
where $\kappa_{0}$ is given in \eqref{defn:kappa:0} and
\begin{equation}\label{def-C-0}
C_{0}:=L_{\delta}^{2}\left(\kappa_{0}+2\left(1\vee\frac{1}{m_{\delta}}\right)\left(b_{\delta}+2\Vert\nabla f(0)\Vert^{2}+d\right)\right)+\Vert\nabla f(0)\Vert^{2},
\end{equation}
where $p_{0}$ is the density of $x_{0}$,
$\kappa_{0}$ is defined in \eqref{defn:kappa:0}
and $c_{LS}$ is the constant for the logarithmic Sobolev inequality
that $\pi_{\delta}$ satisfies which can be bounded as
\begin{equation*}
c_{LS}\leq\frac{2m_{\delta}^{2}+8L_{\delta}^{2}}{m_{\delta}^{2}L_{\delta}}+\frac{1}{\lambda_{\ast}}\left(\frac{6L_{\delta}(d+1)}{m_{\delta}}+2\right),
\end{equation*}
where $\lambda_{*}$ is the spectral gap of the penalized overdamped Langevin SDE \eqref{penalized:overdamped:SDE}
that is defined in \eqref{defn:lambda:ast}.
Moreover, we observe that 
$\tilde{C}_{0}=\mathcal{O}(\tilde{C}_{1})$.
By \eqref{eqn:control}, we have $\mathcal{W}_{2}(\nu_{K},\pi)\leq\tilde{\mathcal{O}}(\varepsilon)$ 
with 
\begin{align*}
&\eta=\Theta\left(\frac{\varepsilon^{4}}{\tilde{C}_{1}^{4}c_{LS}^{4}(\log\tilde{C}_{2})^{4}}\right)
=\tilde{\Theta}\left(\frac{\varepsilon^{196}}{d^{8}\lambda_{\ast}^{-4}(\log(\lambda_{\ast}^{-1}))^{4}}\right),
\\
&K=\tilde{\mathcal{O}}\left(\frac{d^{9}\lambda_{\ast}^{-5}(\log(\lambda_{\ast}^{-1}))^{4}}{\varepsilon^{196}}\right),
\end{align*}
and
\begin{equation*}
\sigma^{2}=\mathcal{O}(\eta)
=\tilde{\Theta}\left(\frac{\varepsilon^{196}}{d^{8}\lambda_{\ast}^{-4}(\log(\lambda_{\ast}^{-1}))^{4}}\right),
\end{equation*}
where $\lambda_{\ast}$ is defined in \eqref{defn:lambda:ast}
so that
\begin{equation*}
b=\Omega\left(\sigma^{-2}\right)
=\tilde{\Omega}\left(\frac{d^{8}\lambda_{\ast}^{-4}(\log(\lambda_{\ast}^{-1}))^{4}}{\varepsilon^{196}}\right).
\end{equation*}
Hence, the stochastic gradient computations require
\begin{equation*}
\hat{K}=Kb=\tilde{\mathcal{O}}\left(\frac{d^{17}\lambda_{\ast}^{-9}(\log(\lambda_{\ast}^{-1}))^{8}}{\varepsilon^{392}}\right)
.
\end{equation*}

Finally, under the further assumptions of Corollary~\ref{cor:strongly:convex},  $f+\frac{S}{\delta}$ is $\mu_{\delta}$-strongly convex (with $\mu_{\delta}:=\frac{2\alpha\rho}{\delta(B+\alpha\rho)}-L$) outside
of an Euclidean ball with radius $R+\rho$ and by Lemma~\ref{lem:f:plus:S:dissipative}, $f+S/\delta$ is $L_{\delta}$-smooth with $L_{\delta}:=L+\frac{\ell}{\delta}$.
By applying Lemma~\ref{lem:close-piecewise-quadratic}, 
there exists a $C^{1}$ function $U$
such that $U$ is $s_{0}$-strongly convex on $\mathbb{R}^{d}$ with
\begin{equation*}
\sup_{x\in\mathbb{R}^{d}}\left(U(x)-\left(f(x)+\frac{S(x)}{\delta}\right)\right)-\inf_{x\in\mathbb{R}^{d}}\left(U(x)-\left(f(x)+\frac{S(x)}{\delta}\right)\right)
\leq R_{0},
\end{equation*}
where $s_{0},R_{0}$ are defined in Lemma~\ref{lem:close-piecewise-quadratic}.
We define $\pi_{U}$ as the Gibbs measure such that $\pi_{U}\propto e^{-U(x)}$ and we also define:
\begin{equation}
\lambda_{U}:=
\inf
\left\{
\frac{\int_{\mathbb{R}^{d}}\Vert\nabla g\Vert^{2}d\pi_{U}}{\int_{\mathbb{R}^{d}}g^{2}d\pi_{U}}:
g\in C^{1}(\mathbb{R}^{d})\cap L^{2}(\pi_{U}),
g\neq 0,
\int_{\mathbb{R}^{d}}gd\pi_{U}=0\right\}.
\end{equation}
Since $U$ is $s_{0}$-strongly convex, by Bakry-\'{E}mery criterion (see Corollary~4.8.2 in \citet{analysis-and-geometry}), we have $\frac{1}{\lambda_{U}}\leq\frac{1}{s_{0}}$. 
Finally, by the Holley-Stroock perturbation principle (see \citet{holley-stroock} and 
Proposition~5.1.6 and the discussion thereafter in \citet{analysis-and-geometry}), we have
$\frac{1}{\lambda_{\ast}}\leq\frac{1}{s_{0}}e^{R_{0}}
\leq\mathcal{O}\left(1\right)$,    
which is a dimension-free bound, where we chose $\delta=\varepsilon^{8}$. 
Hence, we have
$\hat{K}=\tilde{\mathcal{O}}\left(\frac{d^{17}}{\varepsilon^{392}}\right)$
and $\eta=\tilde{\Theta}\left(\frac{\varepsilon^{196}}{d^{8}}\right)$.
The proof is complete.
\hfill $\Box$


\subsection*{Proof of Proposition~\ref{prop:SGHMC:nonconvex}}

Let $\nu_{k}$ be the distribution
of the $k$-th iterate $x_{k}$ of penalized stochastic gradient underdamped Langevin Monte Carlo \eqref{SGHMC:nonconvex:1}-\eqref{SGHMC:nonconvex:2}. 
We recall from Lemma~\ref{lem:f:plus:S:dissipative}
that $f+\frac{1}{\delta}S$ 
is $(m_{\delta},b_{\delta})$-dissipative 
with
$m_{\delta}:=-L-\frac{1}{2}+\frac{m_{S}}{\delta}$
and $b_{\delta}:=\frac{1}{2}\Vert\nabla f(0)\Vert^{2}+\frac{b_{S}}{\delta}$ from \eqref{m:delta:dissipative}
and $f+\frac{1}{\delta}S$ is $L_{\delta}$-smooth with $L_{\delta}:=L+\frac{\ell}{\delta}$
and we also recall from Lemma~\ref{lem-dissipative-implies-lower-bound} and Remark~\ref{remark:f:0}
that $f+\frac{1}{\delta}S\geq 0$.
Then, under Assumption~\ref{assump:f:2}, 
it follows from Theorem~EC.1 and Lemma~EC.6 in \citet{GGZ} that
when the stepsize $\eta\leq\min\{1,\frac{\gamma}{\hat{K}_{2}}(d+A),\frac{\gamma\lambda}{2\hat{K}_{1}},\frac{2}{\gamma\lambda}\}$
where $\lambda,A$ are defined in \eqref{def-lambda}-\eqref{def-A}
where $\hat{K}_{1}:=K_{1}+Q_{1}\frac{4}{1-2\lambda}+Q_{2}\frac{8}{(1-2\lambda)\gamma^{2}}$
and $\hat{K}_{2}:=K_{2}+Q_{3}$, where
$Q_{1}=\Theta(L_{\delta}),
\
Q_{2}=\Theta\left((L_{\delta})^{3}\right),
\
Q_{3}=\Theta\left(L_{\delta}d\right),
\
K_{1}=\Theta\left((L_{\delta})^{2}\right),
\
K_{2} = \Theta\left(1\right)$,
(see Lemma~EC.6 in \citet{GGZ} for the precise definitions of $K_{1},K_{2}$ and $Q_{1},Q_{2},Q_{3}$) and $k\eta\geq e$,
we have
\begin{equation*}
\mathcal{W}_{2}(\nu_{k},\pi_{\delta})
\leq
\left(C_{0}\sigma^{1/2}+C_{1}\eta^{1/2}\right)\cdot(k\eta)^{1/2}\cdot\sqrt{\log(k\eta)}
+C\sqrt{\overline{\mathcal{H}}_{\rho}(\mu_{0})}e^{-\mu_{\ast}k\eta},
\end{equation*}
where $C_{1}$ is given by
\begin{align}
&C_{1}:=\hat{\gamma}
\cdot\Bigg(\frac{3L_{\delta}^{2}}{2\gamma}
\bigg(C_{v}^{d}+\left(2L_{\delta}^{2}C_{x}^{d}+2\Vert\nabla f(0)\Vert^{2}\right)+\frac{2d\gamma}{3}\bigg)
\nonumber
\\
&\qquad\qquad\qquad
+\sqrt{\frac{3L_{\delta}^{2}}{2\gamma}
\bigg(C_{v}^{d}+\left(2L_{\delta}^{2}C_{x}^{d}+2\Vert\nabla f(0)\Vert^{2}\right)+\frac{2d\gamma}{3}\bigg)}\Bigg)^{1/2},\label{def-C-1}
\end{align}
where $\hat{\gamma}$ is given by:
\begin{align}
\hat{\gamma}:=\frac{2\sqrt{2}}{\sqrt{\alpha}}\left(\frac{5}{2}+\log\left(\int_{\mathbb{R}^{2d}}e^{\frac{1}{4}\alpha\mathcal{V}(x,v)}\mu_{0}(dx,dv)
+\frac{1}{4}e^{\frac{\alpha(d+A)}{3\lambda}}\alpha\gamma(d+A)\right)
\right)^{1/2},
\label{def-hat-gamma}
\end{align}
where $\mu_{0}$ is the initial distribution for $(x_{0},v_{0})$ and 
$\lambda,A$ are defined in \eqref{def-lambda}-\eqref{def-A} and 
$\alpha:=\lambda(1-2\lambda)/12$ and
$\mathcal{V}(x,v)$ is the Lyapunov function defined in \eqref{eq:lyapunov}
and moreover
\begin{align}
C_{x}^{d}
:=\frac{\int_{\mathbb{R}^{2d}}\mathcal{V}(x,v)\mu_{0}(dx,dv)
+\frac{4(d+A)}{\lambda}}{\frac{1}{8} (1-2 \lambda)  \gamma^2 },
\quad
C_{v}^{d}
:=\frac{\int_{\mathbb{R}^{2d}}\mathcal{V}(x,v)\mu_{0}(dx,dv)
+\frac{4(d+A)}{\lambda}}{\frac{1}{4}(1-2\lambda)},\label{def-Cxv-d}
\end{align}
where $\hat{\gamma},C_{x}^{d},C_{v}^{d}$ are finite due to \eqref{mu:0:integrability}
and furthermore,
\begin{align}
&\mu_{\ast}:=\frac{\gamma}{768}\min\left\{\lambda L_{\delta}\gamma^{-2},\Lambda^{1/2}e^{-\Lambda}L_{\delta}
\gamma^{-2},\Lambda^{1/2}e^{-\Lambda}\right\}\label{def-mu-star},
\\
&C:=\sqrt{2}e^{1+\frac{\Lambda}{2}}\frac{1+\gamma}{\min\{1,\alpha_{1}\}}
\sqrt{\max\{1,4(1+2\alpha_{1}+2\alpha_{1}^{2})(d+A)\gamma^{-1}\mu_{\ast}^{-1}/\min\{1,R_{1}\}\}},\nonumber 
\\
&\Lambda:=\frac{12}{5}(1+2\alpha_{1}+2\alpha_{1}^{2})(d+A)L_{\delta}\gamma^{-2}\lambda^{-1}(1-2\lambda)^{-1},
\quad\alpha_{1}:=(1+\Lambda^{-1})L_{\delta}\gamma^{-2},\nonumber 
\\
&\varepsilon_{1}:=4\gamma^{-1}\mu_{\ast}/(d+A),
\quad
R_{1}:=4\cdot(6/5)^{1/2}(1+2\alpha_{1}+2\alpha_{1}^{2})^{1/2}(d+A)^{1/2} \gamma^{-1}(\lambda-2\lambda^{2})^{-1/2},
\nonumber
\end{align}
and moreover,
\begin{align}
\overline{\mathcal{H}}_{\rho}(\mu_{0})&:=
R_{1}+R_{1}\varepsilon_{1}\max\left\{L_{\delta}+\frac{1}{2}  \gamma^{2},\frac{3}{4}  \right\}\Vert(x,v)\Vert_{L^{2}(\mu_{0})}^{2}
\nonumber
\\
&\qquad
+R_{1}\varepsilon_{1}\left(L_{\delta}+\frac{1}{2} \gamma^{2}\right)\frac{b_{\delta}+d}{m_{\delta}}
+R_{1}\varepsilon_{1}\frac{3}{4} d
+2R_{1}\varepsilon_{1}\left( f(0)+\frac{\Vert\nabla f(0)\Vert^{2}}{2L_{\delta}}\right), \label{ineq-H-rho-upper}
\end{align}
where $\Vert(x,v)\Vert_{L^{2}(\mu_{0})}^{2}:=\int_{\mathbb{R}^{2d}}\Vert(x,v)\Vert^{2}\mu_{0}(dx,dv)$, 
and finally, $C_{0}$ is defined as:
\begin{equation}
C_{0}:=\hat{\gamma}\cdot\left(\left(L_{\delta}^{2}C_{x}^{d}+\Vert\nabla f(0)\Vert^{2}\right)\frac{1}{\gamma}
+\sqrt{\left(L_{\delta}^{2}C_{x}^{d}+\Vert\nabla f(0)\Vert^{2}\right)\frac{1}{\gamma}}\right)^{1/2},
\end{equation}
where $\hat{\gamma}$ is defined in \eqref{def-hat-gamma} and $C_{x}^{d}$ is defined in \eqref{def-Cxv-d}. 
Thus, it is easy to see that
$C_{0}=\mathcal{O}(C_{1})$,
where $C_{1}$ is given in \eqref{def-C-1}
and we can choose $\sigma^{2}=\mathcal{O}(\eta)$
with 
\begin{equation*}
\eta=\tilde{\Theta}\left(\frac{\varepsilon^{50}\mu_{\ast}}{d^{3}\left(\log(1/\mu_{\ast})\right)^{2}}\right),
\end{equation*}
and the batch-size $b=\Omega(\sigma^{-2})$ such that
\begin{equation*}
b=\tilde{\Theta}\left(\frac{d^{3}\left(\log(1/\mu_{\ast})\right)^{2}}{\varepsilon^{50}\mu_{\ast}}\right),
\end{equation*}
where $\mu_{\ast}$ is defined in \eqref{def-mu-star}.
Hence, the stochastic gradient computations require
\begin{equation*}
\hat{K}=Kb=\tilde{\mathcal{O}}\left(\frac{d^{7}\left(\log(1/\mu_{\ast})\right)^{5}}{\varepsilon^{132}\mu_{\ast}^{3}}\right).
\end{equation*}
The proof is complete.
\hfill $\Box$

    


\subsection*{Proof of Lemma~\ref{lemma:sum_gix_smooth}}

We first show that $H_i(x):= \max(0,h_{i}(x))^{2}$ is continuously differentiable and convex. Note that the functions $h_i(x)$ and $\max(0,x)$ are both convex in $x$. Since the composition of convex functions is convex, $H_i(x)$ is convex. By the chain rule for convex functions in Section~3.3 of \citet{browien1999convex}, the subdifferential of $H_i(x)$ is given by
\begin{equation*}
    \partial H_i(x) = \begin{cases}
    0 &\text{if } h_i(x) \leq 0,\\
    2h_i(x)\nabla h_i(x) &\text{if } h_i(x) > 0, 
    \end{cases}
\end{equation*}
where in the case $h_{i}(x)=0$, we used the fact that the subdifferential of the convex function $\max(0,x)$ at $x=0$ is given by the interval $[0,1]$; so that by the chain rule, the subdifferential of $\partial H_i(x) = 2\max(0, h_i(x)) [0,1] = 0 $ is single-valued for $h_i(x)= 0$. Since the subdifferential of $H_i(x)$ is single-valued for any $x$ and is also continuous, we conclude that $H_i(x)$ is continuously differentiable.

Let $\mathcal{C}_i$ be the convex set on which $h_i(x) \leq 0$, i.e. $\mathcal{C}_i := \{ x \in \mathbb{R}^d ~:~ h_i(x) \leq 0\}$. Since $h_i(x)$ is continuous, $x \in \mbox{bd}(\mathcal{C}_i)$ if and only if $h_i(x) = 0$ where $\mbox{bd}(\cdot)$ denotes the boundary of a set. Note that the Hessian of $H_i$, denoted by $\text{Hess}_{i}$, is continuous except at the boundary of $\mathcal{C}_i$ and can be computed as
\begin{equation*}
\text{Hess}_{i}(x) = 2\left[ \nabla h_i(x) \cdot (\nabla h_i(x))^{\top} + h_i(x) \nabla^2 h_i(x)\right],
\end{equation*}
if $x\not\in \mathcal{C}_i$. This is the case when $h_i(x)>0$. On the other hand, for $x\in \mbox{int}(\mathcal{C}_i)$, we have $H_i(x)=0$ and $\text{Hess}_{i}(x) = 0$ where $\mbox{int}(\cdot)$ denotes the interior of a set. Therefore, for any $x\in \mathbb{R}^d \setminus \mbox{bd}(\mathcal{C}_i)$,
\begin{equation}
    \|\text{Hess}_{i}(x)\| \leq  2 \left( \| \nabla h_i(x)\|^2  + \max\nolimits_{x\in\mathbb{R}^{d}} |h_i(x)| \left\|\nabla^2 h_i(x)\right\|\right) \leq \ell_i:= 2\left(N_i^2 + P_i\right),
    \label{ineq-Hessian-bound}
\end{equation} 
where $\| \cdot \|$ denotes the matrix 2-norm (largest singular value), and we used the triangular inequality and sub-multiplicativity of the matrix 2-norm in \eqref{ineq-Hessian-bound}. 
So far, we have shown that $H_i(x)$ is $\ell_i$-smooth on the open set that excludes the boundary points of $\mathcal{C}_i$. For establishing smoothness at the boundary points $x \in \mbox{bd}(\mathcal{C}_i)$, our proof relies on a more technical argument as the Hessian of $H_i$ may not even exist for $x \in \mbox{bd}(\mathcal{C}_i)$.\footnote{For example, in dimension one; the unit ball around origin is defined by $m=2$ constraints with $h_1(x)= x-1\leq 0$ and $h_2(x)=-x-1\leq 0$ where $\mbox{bd}(\mathcal{C}_1)=\{1\}$ and $\mbox{bd}(\mathcal{C}_2)=\{-1\}$. In this case, $\text{Hess}_{1}(x) = 0$ for $-1<x<1$ and $\text{Hess}_{1}(x) = 2$ for $x>1$ and the Hessian does not exist at $x=1$.} Our argument will roughly use the fact that boundary points constitute a measure zero set and the gradient of $H_i$ is continuous at the boundary. For this purpose, next, we consider the line $\ell(t) := x + t(y-x)$ that passes through the points $x$ and $y$, parameterized by the scalar $t\in \mathbb{R}$.  
Let 
\begin{equation*}
T:= \left\{ t \in [0,1] ~:~ \ell(t) \in \mbox{bd}(\mathcal{C}_i)\right\}
\end{equation*}
correspond to the set of times $t$ when the line segment between $x$ and $y$ crosses the boundary of the set $\mathcal{C}_i$. If we introduce $z(t) := \nabla H_i(\ell(t))$, then $z(t)$ is continuous, and it is continuously differentiable except when $t\in T$. Since $\mathcal{C}_i$ is closed, $T$ is closed. Recalling that $\mathcal{C}_i$ is convex, roughly speaking, the line segment cannot go strictly out of the set $\mathcal{C}_i$ and then re-enter. We have three different cases: 

\begin{itemize} 
    \item [I.] $T$ is the empty set: This case can arise when the line segment of $x$ and $y$ (including the endpoints) never intersects the set $\mathcal{C}_i$. In this case, $H_i$ is twice continuously differentiable along the line segment. Thus, by Taylor's theorem with a remainder, 
     we have
     \begin{align*}
     \| \nabla H_i(x) - \nabla H_i(y)\| &= \| z(1) - z(0) \|
     =\left\| \int_{t=0}^1 z'(t) dt\right\| 
     \\
     &=  \left\| \int_{t\in[0,1]}\text{Hess}_{i}(\ell(t)) dt \right\| 
     \leq  L_i \|x-y\|,
     \end{align*}
     where we used \eqref{ineq-Hessian-bound}. 
    \item [II.] $T = [t_{1}, t_{2}]$ for some $t_{1} \leq t_{2}$ with the convention that $T$ is a singleton when $t_{1} = t_{2}$. In this case, $z(t)$ may not be differentiable for some points in $[0,1]$; however, we can approximate the interval [0,1] with unions of intervals where $z(t)$ is differentiable. More specifically, for any given $\varepsilon>0$, we consider the closed intervals
    $I_1 = [\varepsilon, t_1 -\varepsilon], I_2 = [t_1 + \varepsilon, t_2 - \varepsilon], I_3 = [t_2+\varepsilon, 1- \varepsilon]$ with the convention that $[a,b]$ denotes the empty set when $a>b$. The union $\bigcup_{i=1,2,3}I_i$ approximates the interval $[0,1]$ when $\varepsilon$ is sufficiently small. The function $z(t)$ is continuously differentiable for every $t \in I_i$ for $i=1,2,3$ if $\varepsilon>0$ is small enough except when $t\in T$. Furthermore, by the continuity of $z(t)$ and the fact that $z(t)=0$ for $t\in T$, we have
      \begin{align*} \nabla H_i(x) - \nabla H_i(y) &= z(0) - z(1) \\
      &= \int_{t\in I_1 \cap T^c}z'(t)dt + \int_{t\in I_2 \cap T^c }z'(t)dt + \int_{t\in I_3\cap T^c}z'(t)dt  + o(\varepsilon)
      \\
      &= \int_{t\in (I_1 \cup I_2 \cup I_c) \cap T^c } \text{Hess}_{i} (x + t(y-x)) dt  + o(\varepsilon), 
    \end{align*}
    where $T^c$ denotes the  complement of the set $T$ and we used the fact that $z(t)$ is continuously differentiable on the set $t\in I_i \cap T^c$ for any $i \in \{1,2,3\}$. Taking the limit as $\varepsilon\to 0$, by a similar argument to Case I, we obtain $\|\nabla H_i(x) - \nabla H_i(y)\| \leq \ell_i \|x-y\| $, 
    where $\ell_{i}:=2(N_{i}^{2}+M_{i}P_{i})$.
    
      \item [III.] $T = \{t_1, t_2\}$ for some $t_1\neq t_2$. This case can be treated similarly to Case II by considering the intervals $I_1, I_2$, and $I_3$.
\end{itemize}

Combining these cases, we can conclude that $H_i(x)$ is $\ell_i$-smooth on $\mathbb{R}^d$, where $\ell_i:=2(N_i^2+M_iP_i)$. Hence $\sum_{i=1}^{m}\max(0,h_{i}(x))^{2} = \sum_{i=1}^m H_i(x)$ is $\ell$-smooth, where $\ell := \sum_{i=1}^m \ell_i = 2 \sum_{i=1}^m \left( N_i^2+M_iP_i \right)$. This completes the proof.
\hfill $\Box$

\subsection*{Proof of Corollary~\ref{prop:lp-norm}}

To use Lemma~\ref{lemma:sum_gix_smooth}, we need to prove that $h(x)$ satisfies its assumptions. Note that the function  $t_i(x) := |x_i|^p$ is twice continuously differentiable in $x=(x_{1},\ldots,x_{d})\in\mathbb{R}^d$ for $p\geq 2$, $i=1,2,\ldots,d$ and the function $t_0:\mathbb{R}_{\geq 0} \to \mathbb{R}_{\geq 0}$ defined as $t_0(z):=z^{1/p}$ is twice continuously differentiable unless $z=0$. Since the sum and composition of twice continuously differentiable functions remain twice continuously differentiable, we conclude that the $p$-norm $\lVert x \rVert_p := t_0\left( \sum_{i=1}^{d} |x_i|^p\right) = \left( \sum_{i=1}^{d} |x_i|^p\right) ^{1/p}$ is twice continuously differentiable unless $x=0$. Therefore, we conclude that $h(x)$ is twice continuously differentiable on the set 
\begin{equation*}
\mathcal{B}:=\left\{x\in \R^d :\,\, h(x) \geq 0\right\}, 
\end{equation*}
which does not include $x=0$. Since the $p$-norm is convex, $h(x)$ is also convex. For the rest, it suffices to prove that on the set $\mathcal{B}$, $h(x)$ has bounded gradients, and the product of $|h(x)|$ and the Hessian is bounded. For any $x\neq 0$, the gradient of $h(x)$ is given by:
\begin{equation*}
\nabla h(x) = \left( \left(|x_i|/\lVert x \rVert_{p}\right)^{p-1}\mbox{sgn}(x_i),\,\,1\leq i\leq d \right),
\end{equation*}
with $\mbox{sgn}(x) :=
    -1$  if $x<0$, $1$ if $x>0$, and $0$ if $x=0$.
According to the definition of $p$-norm $\lVert x \rVert_p = \left( \sum_{i=1}^{d} |x_i|^p\right) ^{1/p}$, we have $|x_i|\leq \lVert x \rVert _p$ for any $i$, such that $|(\nabla h(x))_i| \leq 1$, which implies that $\lVert \nabla h(x) \rVert \leq \sqrt{d}$. Next, we consider the Hessian matrix of $h(x)$. After some computations, the entries $(i,j)$ of the Hessian matrix of $h$ are given by 
\begin{equation}\label{equ:Hessian}
    \left[\nabla^2 h(x)\right]_{i,j} = \begin{cases}
    (p-1)\frac{1}{\lVert x \rVert_p^p}\left( |x_i|^{p-2} \lVert x\rVert_p - \frac{|x_i|^{2p-2}}{\lVert x\rVert_p^{p-1}}\right) &\text{ if } i = j,\\
    -\mbox{sgn}(x_ix_j)(p-1)\frac{|x_i|^{p-1}|x_j|^{p-1}}{\lVert x \rVert_p^{2p-1}} &\text{ if } i\neq j,
    \end{cases}
\end{equation}
provided that $x\neq 0$. Note that the Hessian matrix $\nabla^2 h(x)$ is continuous unless $x=0$.

Since $|x_i|\leq \lVert x \rVert _p$ for any $i$, we obtain the following bounds for the elements of Hessian matrix on the set $\mathcal{B}=\{x\in \R^d :\,\, h(x) \geq 0\}=\{x\in \R^d: \,\, \lVert x\rVert_p \geq R\}$:
\begin{equation*}
    0\leq \left| h(x)\right|\left[\nabla^2 h(x)\right]_{i,i} \leq (p-1)\frac{\lVert x \rVert_p-R}{\lVert x \rVert_p^p}\left( 2\|x\|^{p-1} \right)
    \leq 2(p-1)\frac{\lVert x \rVert_p-R}{\lVert x \rVert_p}
    \leq \frac{2(p-1)}{R},
\end{equation*}
and for $i\neq j$, we have
    \begin{align*}
    |h(x)|\cdot\left|\left[\nabla^2 h(x)\right]_{i,j}\right|
    \leq (p-1)\frac{\lVert x \rVert_p-R}{\lVert x \rVert_p} 
    \leq p-1.
    \end{align*}
Therefore, by applying the Gershgorin circle theorem (see, e.g., \citet{fan1958note}), we obtain 
\begin{equation*}
|h(x)|\nabla^2 h(x) \preceq \left(\frac{2}{R} + (d-1)\right) (p-1) I.
\end{equation*}
Hence, $\max(0,h(x))^2$ is $\ell$-smooth with $\ell=\left(\frac{2}{R} + (d-1)\right) (p-1)$ and the proof is complete.


\subsection*{Proof of Lemma~\ref{lem-S-dissipative}}

The proof is similar to the proof of Lemma~\ref{lem:S} with some minor differences to the potential non-convexity of the set $\mathcal{C}$. 
By the assumption for every $x \in \mathbb{R}^d$ there exists a unique point of $\mathcal{C}$
nearest to $x$. Then the fact that $S(x)=\left(\delta_{\mathcal{C}}(x)\right)^2$
is $\ell$-smooth and continuously differentiable with the gradient $\nabla S(x) = 2 (x - \mathcal{P}_{\mathcal{C}}(x))$ is a direct consequence of \citet[Theorem~4.8]{federer1959curvature}. Note that for $x_1, x_2 \in \mathbb{R}^d$,
\begin{equation*}
 \|\nabla S(x_1) - \nabla S(x_2)\| \leq 2 \|x_1 - x_2\| + 2 \|\mathcal{P}_{\mathcal{C}}(x_1) - \mathcal{P}_{\mathcal{C}}(x_2)
\| \leq 4 \|x_1 - x_2\|,
\end{equation*}
where in the last step we applied \citet[Theorem~4.8, part (8)]{federer1959curvature}. Therefore, $S$ is $\ell$-smooth with $\ell=4$. Also,
$$
\langle x, \nabla S(x) \rangle = \langle x,  2 (x - \mathcal{P}_{\mathcal{C}}(x)) \rangle \geq 2\|x\|^2 - R\|x\| \geq m_S \|x\|^2 - b_S,
$$
for $m_S = 1$, $b_S = R^2/4$. This completes the proof.
\hfill $\Box$


\subsection*{Proof of Lemma~\ref{lem:f:plus:S:dissipative}}

Lemma~\ref{lem-S-dissipative} shows that $S(x)$ is $(m_{S},b_{S})$-dissipative and $\ell$-smooth. 
Then it follows that $f+\frac{1}{\delta}S$ 
is also $L_{\delta}$-smooth, 
where $L_{\delta}:=L+\frac{\ell}{\delta}$.
By $(m_{S},b_{S})$-dissipativity of $S$, we have
\begin{align}
\left\langle x,\nabla f(x)+\frac{1}{\delta}\nabla S(x)\right\rangle
&\geq 
\langle x,\nabla f(x)\rangle
+\frac{m_{S}}{\delta}\Vert x\Vert^{2}-\frac{b_{S}}{\delta}
\nonumber
\\
&\geq
\langle x,\nabla f(x)-\nabla f(0)\rangle
-\Vert x\Vert\cdot\Vert \nabla f(0)\Vert
+\frac{m_{S}}{\delta}\Vert x\Vert^{2}-\frac{b_{S}}{\delta}
\nonumber
\\
&\geq
-L\Vert x\Vert^{2}
-\Vert x\Vert\cdot\Vert\nabla f(0)\Vert
+\frac{m_{S}}{\delta}\Vert x\Vert^{2}-\frac{b_{S}}{\delta}
\nonumber
\\
&\geq
-L\Vert x\Vert^{2}
-\frac{1}{2}\Vert x\Vert^{2}-\frac{1}{2}\Vert\nabla f(0)\Vert^{2}
+\frac{m_{S}}{\delta}\Vert x\Vert^{2}-\frac{b_{S}}{\delta},
\label{m:delta:dissipative}
\end{align}
where we used $L$-smoothness of $f$.
Therefore, $f+\frac{1}{\delta}S$ 
is also $(m_{\delta},b_{\delta})$-dissipative 
with 
$m_{\delta}:=-L-\frac{1}{2}+\frac{m_{S}}{\delta}>0$
and
$b_{\delta}:=\frac{1}{2}\Vert\nabla f(0)\Vert^{2}+\frac{b_{S}}{\delta}$,
provided that $\delta<m_{S}/(L+\frac{1}{2})$.
This completes the proof.
\hfill $\Box$


\subsection*{Proof of Lemma~\ref{lem-dissipative-implies-lower-bound}}

Since $f$ is $L$-smooth, we have
\begin{equation*}
f(x)\geq f(0)-\Vert\nabla f(0)\Vert\cdot\Vert x\Vert-\frac{L}{2}\Vert x\Vert^{2},
\end{equation*}
and since $S$ is $(m_{S},b_{S})$-dissipative and bounded below by $0$, 
by Lemma~2 in \citet{Raginsky}, 
we have 
\begin{equation*}
S(x)\geq\frac{m_{S}}{3}\Vert x\Vert^{2}-\frac{b_{S}}{2}\log 3,
\end{equation*}
for any $x\in\mathbb{R}^{d}$ and thus
\begin{align}
f(x)+\frac{S(x)}{\delta}
&\geq f(0)-\Vert\nabla f(0)\Vert\cdot\Vert x\Vert-\frac{L}{2}\Vert x\Vert^{2}
+\frac{m_{S}}{3\delta}\Vert x\Vert^{2}-\frac{b_{S}}{2\delta}\log 3
\nonumber
\\
&\geq f(0)-\frac{1}{2}\Vert\nabla f(0)\Vert^{2}-\frac{1}{2}\Vert x\Vert^{2}-\frac{L}{2}\Vert x\Vert^{2}
+\frac{m_{S}}{3\delta}\Vert x\Vert^{2}-\frac{b_{S}}{2\delta}\log 3
\geq -M,
\end{align}
where
$M:=-f(0)+\frac{1}{2}\Vert\nabla f(0)\Vert^{2}+\frac{b_{S}}{2\delta}\log 3$,
provided that $\delta\leq\frac{2m_{S}}{3(1+L)}$.
This completes the proof.
\hfill $\Box$


\subsection*{Proof of Lemma~\ref{condition:thm:final}}

If Assumption~\ref{assump:f:2} and Assumption~\ref{assump:C} hold, 
according to Lemma~\ref{lem-dissipative-implies-lower-bound}, function $f+\frac{1}{\delta}S$ is uniformly lower bounded, i.e. $f+\frac{1}{\delta}S \geq -M$ for  an explicit non-negative scalar $M$ defined in \eqref{M:eqn}, which leads to the result that $f+\frac{1}{\delta}S+M$ is non-negative. Then according to Lemma~\ref{lem:f:plus:S:dissipative}, function $f+\frac{1}{\delta}S$ is $L_\delta$-smooth and $(m_\delta,b_\delta)$-dissipative, where $L_\delta,m_\delta,b_\delta$ is defined in \eqref{m:b:delta}. By Lemma~2 in \citet{Raginsky}, 
we have 
\begin{equation*}
f(x)+\frac{S(x)}{\delta}+M\geq\frac{m_{\delta}}{3}\Vert x\Vert^{2}-\frac{b_{\delta}}{2}\log 3,
\end{equation*}
for any $x\in\mathbb{R}^{d}$.
Hence $e^{-f}$ is integrable over $\mathcal{C}$, and moreover,
\begin{equation}
\int_{\mathbb{R}^{d}}e^{\frac{m_{\delta}}{6}\Vert x\Vert^{2}}e^{-f(x)-\frac{S(x)}{\delta}}dx
\leq
e^{\frac{b_{\delta}}{2}\log 3+M}\int_{\mathbb{R}^{d}}e^{-\frac{m_{\delta}}{6}\Vert x\Vert^{2}}dx<\infty.
\end{equation}
So that the assumptions in Theorem~\ref{thm:final} are satisfied with $\hat{\alpha}=\frac{m_{\delta}}{6}$ and $\hat{x}=0$. 
This completes the proof.
\hfill $\Box$


\subsection*{Proof of Lemma~\ref{condition:thm:final:2}}

Since it follows from Lemma~\ref{lem:S}
that $S(x)$ is convex and $\ell$-smooth, 
it follows that under Assumption~\ref{assump:f:1}, $f+\frac{1}{\delta}S$ 
is also $\mu$-strongly convex and $L_{\delta}$-smooth, 
where $L_{\delta}:=L+\frac{\ell}{\delta}$. Moreover, we notice that 
since $f$ is $\mu$-strongly convex, 
$f(x)\geq f(x_{\ast})+\frac{\mu}{2}\Vert x-x_{\ast}\Vert^{2}$, 
where $x_{\ast}$ is the unique minimizer of $f$. 
Hence, $e^{-f}$ is integrable over $\mathcal{C}$ and moreover 
\begin{equation}\label{exp:int:1}
\int_{\mathbb{R}^{d}}e^{\frac{\mu}{4}\Vert x-x_{\ast}\Vert^{2}}e^{-\frac{S(x)}{\delta}-f(x)}dx
\leq\int_{\mathbb{R}^{d}}e^{\frac{\mu}{4}\Vert x-x_{\ast}\Vert^{2}}e^{-f(x)}dx
\leq 
e^{-f(x_{\ast})}\int_{\mathbb{R}^{d}}e^{-\frac{\mu}{4}\Vert x-x_{\ast}\Vert^{2}}dx<\infty,
\end{equation}
so that the assumptions in Theorem~\ref{thm:final} are satisfied with $\hat{\alpha}=\frac{\mu}{4}$ and $\hat{x}=x_{\ast}$. 
This completes the proof.
\hfill $\Box$


\subsection*{Proof of Lemma~\ref{lemm:strongly convex}}

We denote $x_{\mathcal{C}^\alpha}$ and $y_{\mathcal{C}^\alpha}$ as the projections of $x$ and $y$ onto $\mathcal{C}^{\alpha}$. Since we can compute that 
$S^\alpha(x)=\|x-x_{\mathcal{C}^\alpha}\|^2$ and $\nabla S(x) = 2(x-x_{\mathcal{C}^\alpha})$, we have:
\begin{equation*}
\nabla S^\alpha(x)- \nabla S^\alpha(y) = 2\left(x-x_{\mathcal{C}^\alpha}\right)- 2\left(y-y_{\mathcal{C}^\alpha}\right) = 2(x-y) - 2\left(x_{\mathcal{C}^\alpha} - y_{\mathcal{C}^\alpha}\right).
\end{equation*}
It follows that:
\begin{align}
(\nabla S^\alpha(x) -\nabla S^\alpha(y))^{\top} (x-y) &= 2\|x-y\|^2 - 2\left(x_{\mathcal{C}^\alpha} - y_{\mathcal{C}^\alpha}\right)^{\top}(x-y)\nonumber
\\
&\geq 2\|x-y\|^2 - 2\left\|x_{\mathcal{C}^\alpha} - y_{\mathcal{C}^\alpha}\right\|\|x-y\|.\label{combining:1}
\end{align}
By the assumptions, $\mathcal{C}^\alpha:=\{x: h^\alpha(x) \leq 0\}$, where $h^\alpha(x)$ is a continuous $(\alpha+\beta)$-strongly convex function. By the convexity of $h^\alpha$, it is Lipschitz on  compact sets \citep{cvx-locally-lip} and therefore there exists a positive constant $B$ such that $\| y\|\leq B$ for any $y \in \partial h^{\alpha}(x)$ and $x\in \mathcal{C}^\alpha$. 
According to Corollary~2 in \citet{vial1982strong}, the set $\mathcal{C}^\alpha$ is strongly convex with radius $B/(\alpha+\beta)$ in the sense of Definition~1.1 in \citet{balashov2012lipschitz}.\footnote{A nonempty subset $\mathcal{C} \subset \mathbb{R}^d$ is called strongly convex of radius $R>0$ if it can be represented as the intersection of closed balls of radius $R>0$, i.e., there exists a subset $X\subset \mathbb{R}^d$ such that $\mathcal{C}=\bigcap_{x\in X}B_R(x)$, where $B_R(x)$ is a closed ball with radius $R$ centered with $x$, see Def.~1.1 in \citet{balashov2012lipschitz}.} 
Then by applying Corollary~2.1 in \citet{balashov2012lipschitz}, for any $x,y\in \mathbb{R}^d\backslash U(\mathcal{C}^\alpha,\rho)$, we have:
\begin{equation*}
\left\|x_{\mathcal{C}^\alpha} - y_{\mathcal{C}^\alpha}\right\| \leq \frac{B}{B+(\alpha+\beta) \rho}\|x-y\|.
\end{equation*}

By combining these two inequalities, we have:
\begin{equation*}
(\nabla S^\alpha(x) -\nabla S^\alpha(y))^{\top} (x-y) \geq \frac{2(\alpha+\beta) \rho}{B+(\alpha +\beta)\rho} \|x-y\|^2.
\end{equation*}
By Theorem~2.1.10 in \citet{nesterov2013introductory}, we conclude that the penalty function $S^\alpha(x)$ is strongly convex with constant $\frac{2(\alpha+\beta) \rho}{B+(\alpha+\beta) \rho}$ outside the $\rho$-neighborhood of the set $\mathcal{C}^{\alpha}$. The proof is complete.
\hfill $\Box$


\subsection*{Proof of Corollary~\ref{cor:strongly:convex}}

By Lemma~\ref{lemm:strongly convex}, the penalty function $S^\alpha(x)$ is strongly convex with constant $\frac{2(\alpha+\beta) \rho}{B+(\alpha+\beta) \rho}$ on the set $\mathbb{R}^{d}\backslash U(\mathcal{C}^\alpha,\rho)$, where $U(\mathcal{C}^\alpha, \rho)$ is the open $\rho$-neighborhood of $\mathcal{C}^\alpha$ i.e. 
\begin{equation*}
U(\mathcal{C}^\alpha, \rho):=\{x: \text{dist}(x,\mathcal{C}^\alpha)<\rho\}.
\end{equation*}
Since $\mathcal{C}^\alpha$ is contained in an Euclidean ball centered at $0$ of radius $R$, 
it follows that $S^\alpha$ is strongly convex with constant $\frac{2(\alpha+\beta) \rho}{B+(\alpha+\beta) \rho}$
outside a Euclidean ball with radius $R+\rho$ and moreover, 
\begin{equation}
S^\alpha(x)\geq S^\alpha(y)+\langle\nabla S^\alpha(x),y-x\rangle+\frac{1}{2}\frac{2(\alpha+\beta) \rho}{B+(\alpha+\beta) \rho}\Vert x-y\Vert^{2},
\end{equation}
for any $x,y$ outside an Euclidean ball with radius $R+\rho$.
On the other hand, by Assumption~\ref{assump:f:2}, it follows 
that for any $x,y$:
$f(x)\geq f(y)+\langle\nabla f(x),y-x\rangle-\frac{L}{2}\Vert x-y\Vert^{2}$,
which implies:
\begin{align*}
f(x)+\frac{S^\alpha(x)}{\delta}
&\geq f(y)+\frac{S^\alpha(y)}{\delta}
\\
&\qquad
+\left\langle\nabla f(x)+\frac{S^\alpha(x)}{\delta},y-x\right\rangle+\frac{1}{2}\left(\frac{2(\alpha+\beta) \rho}{\delta(B+(\alpha+\beta) \rho)}-L\right)\Vert x-y\Vert^{2},
\end{align*}
for any $x,y$ outside an Euclidean ball with radius $R+\rho$.
This completes the proof.
\hfill $\Box$



\subsection*{Proof of Lemma~\ref{lem:close-piecewise-quadratic}}
We start with defining
\begin{equation*}
U(x):=f(x)+\frac{S^{\alpha}(x)}{\delta} + u(x),
\end{equation*}
where
\begin{equation}
u(x):=
\begin{cases}
 \frac{m + L}{2}\|x\|^2 &\mbox{for} \quad \Vert x\Vert<R+\rho,\\
 -\frac{\mu_{\delta}}{4}\|x\|^2 +a_\delta\|x\| + b_\delta    &\mbox{for} \quad R+\rho \leq \Vert x\Vert \leq (R+\rho) \left( 1 + \frac{2(m + L)}{\mu_{\delta}} \right) ,\\  
 c_\delta &\mbox{for} \quad \Vert x\Vert > (R+\rho) \left( 1 + \frac{2(m + L)}{\mu_{\delta}} \right),
\end{cases}
\end{equation}
with 
\begin{align*}
&a_\delta :=  (m + L + \mu_{\delta}/2)(R+\rho),
\\
&b_\delta := -\frac{1}{2}(R+\rho)^2 (m + L + \mu_{\delta}/2),
\\
&c_\delta :=(R+\rho)^2 \left( m+L + \frac{2(m+L)^2}{\mu_{\delta}} \right).
\end{align*}

In the first region, when $\|x\| < R+ \rho$,
we observe that the function $u(x)$ is a piecewise-defined quadratic that is clearly $(m+L)$-strongly convex. Since $S^\alpha(x)$ is convex and $f$ is $L$-smooth, this implies that $U$ is $m$-strongly convex in the first region when $\|x\| < R+ \rho$.

In the second region, when $R+\rho \leq\Vert x\Vert\leq (R+\rho) \left( 1 + \frac{2(m + L)}{\mu_{\delta}} \right)$, $u$ is a quadratic that is $\mu_{\delta}/2$-strongly concave (or equivalently $-u(x)$ is $\mu_{\delta}/2$-strongly convex) and $f +S/\delta$ is strongly convex with constant $\mu_{\delta}$, consequently $U$ is strongly convex with constant $\mu_{\delta}/2$. 

In the third region, outside the Euclidean ball with radius $(R+\rho) \left( 1 + \frac{2(m + L)}{\mu_{\delta}} \right)$, we observe that $u(x)\equiv c_\delta$ is a constant. Therefore $U = f+S^{\alpha}/\delta+u$ is $\mu_{\delta}$-strongly convex. 

Moreover, it is straightforward to check that the piecewise function $u$ has continuous derivatives and is of class $C^1$ and therefore $U=f+\frac{S^\alpha}{\delta}+u$ is a $C^{1}$ function. 
Finally, it is easy to check that $ \sup_{x\in\mathbb{R}^d} \|u(x)\| = c_\delta$.
Therefore, 
\begin{equation*}
\sup_{x\in\mathbb{R}^{d}}\left(U(x)-\left(f(x)+\frac{S^\alpha(x)}{\delta}\right)\right)-\inf_{x\in\mathbb{R}^{d}}\left(U(x)-\left(f(x)+\frac{S^\alpha(x)}{\delta}\right)\right)\leq 2 c_\delta,
\end{equation*}
and the result follows. The proof is complete.
\hfill $\Box$

\end{document}